\documentclass[10pt,twocolumn,letterpaper]{article}

\usepackage{cvpr}              % To produce the CAMERA-READY version
\usepackage[accsupp]{axessibility} % Improves PDF readability for those with visual impairments.

\usepackage{bm}
\usepackage{multirow}
\usepackage{arydshln}
\newcommand{\fst}[1]{\textbf{#1}}
\newcommand{\snd}[1]{\emph{\underline{#1}}}
\newcommand{\method}{NDE\xspace}

\usepackage[dvipsnames]{xcolor}

\definecolor{cvprblue}{rgb}{0.21,0.49,0.74}
\usepackage[pagebackref,breaklinks,colorlinks,citecolor=cvprblue]{hyperref}

\DeclareGraphicsExtensions{.pdf,.jpg,.jpg}

\def\paperID{6660} % *** Enter the Paper ID here
\def\confName{CVPR}
\def\confYear{2024}

\title{Neural Directional Encoding\\
for Efficient and Accurate View-Dependent Appearance Modeling}

\author{
Liwen Wu$^{1}$\thanks{This work was partially done during an internship at Adobe Research.} \quad
Sai Bi$^2$\quad
Zexiang Xu$^2$\quad
Fujun Luan$^2$\quad
Kai Zhang$^2$\\
Iliyan Georgiev$^2$\quad
Kalyan Sunkavalli$^2$\quad
Ravi Ramamoorthi$^1$\\
$^1$UC San Diego \quad $^2$Adobe Research
}

\begin{document}

\maketitle

\begin{abstract}
Novel-view synthesis of specular objects like shiny metals or glossy paints remains a significant challenge.
Not only the glossy appearance but also global illumination effects, including reflections of other objects in the environment, are critical components to faithfully reproduce a scene.
In this paper, we present Neural Directional Encoding (\method), a view-dependent appearance encoding of neural radiance fields (NeRF) for rendering specular objects.
NDE transfers the concept of feature-grid-based spatial encoding to the angular domain, significantly improving the ability to model high-frequency angular signals.
In contrast to previous methods that use encoding functions with only angular input, we additionally cone-trace spatial features to obtain a spatially varying directional encoding, which addresses the challenging interreflection effects.
Extensive experiments on both synthetic and real datasets show that a NeRF model with NDE (1)~outperforms the state of the art on view synthesis of specular objects, and (2)~works with small networks to allow fast (real-time) inference.
The project webpage and source code are available at: \url{https://lwwu2.github.io/nde/}.

\end{abstract}

\begin{figure}[t]
    \centering
    \setlength\tabcolsep{1pt}
    \resizebox{0.99\linewidth}{!}{
        \begin{tabular}{cccc}
        % Labels for the two big images
        \multicolumn{2}{c}{\textbf{NDE (ours)}} &
        \multicolumn{2}{c}{Ground truth}\\
        % The two big images
        \multicolumn{2}{@{\hskip -4mm}c@{\hskip -4mm}}{\includegraphics[trim={90 0 90 0},clip,width=0.5275\linewidth]{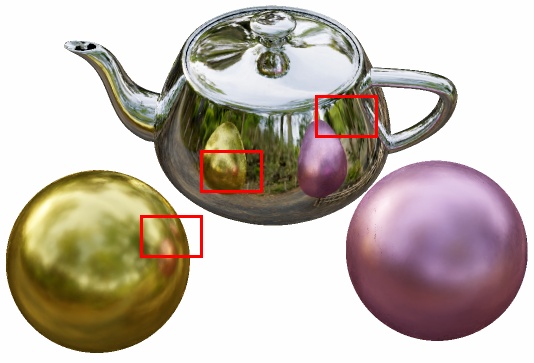}} &
        \multicolumn{2}{@{\hskip -4mm}c@{\hskip -4mm}}{\includegraphics[trim={90 0 90 0},clip,width=0.5275\linewidth]{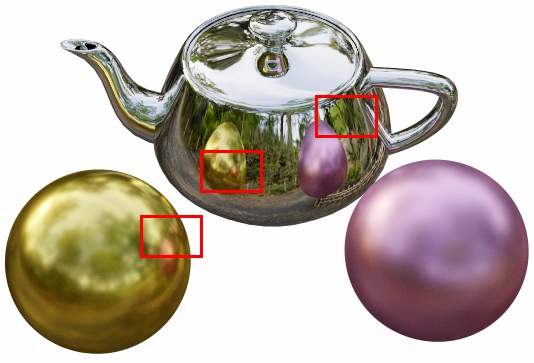}} \\[-1.5mm]
        \hdashline
        \\[-3.5mm]
        % \multicolumn{2}{r}{\textbf{FPS: 75}} & \\%[-.9mm]
        % The crops with labels underneath
        \includegraphics[width=0.26\linewidth]{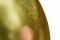}&
        \includegraphics[width=0.26\linewidth]{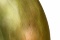}&
        \includegraphics[width=0.26\linewidth]{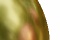}&
        \includegraphics[width=0.26\linewidth]{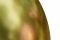}\\[-.9mm]
        \includegraphics[width=0.26\linewidth]{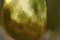}&
        \includegraphics[width=0.26\linewidth]{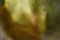}&
        \includegraphics[width=0.26\linewidth]{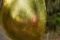}&
        \includegraphics[width=0.26\linewidth]{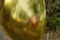}\\[-.9mm]
        \includegraphics[width=0.26\linewidth]{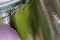}&
        \includegraphics[width=0.26\linewidth]{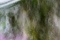}&
        \includegraphics[width=0.26\linewidth]{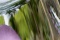}&
        \includegraphics[width=0.26\linewidth]{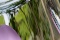}\\
         ENVIDR~\cite{Liang2023ENVIDRID} & Ref-NeRF~\cite{verbin2022ref} & \textbf{NDE (ours)} & Ground truth\\
        0.52 FPS & 0.02 FPS & \textbf{75 FPS} & \\
        \end{tabular}
    }
    \caption{
        \textbf{Ours vs.\ analytical encoding.}
        Methods like Ref-NeRF~\cite{verbin2022ref} use an analytical function to encode viewing directions in large MLPs, failing to model complex reflections (column 1-2 of the insets).
        Instead, we encode view-dependent effects into feature grids with better interreflection parameterization, successfully reconstructing the details on the teapot and even multi-bounce reflections of the pink ball (3rd column of the insets) with little computational overhead (75 FPS on an NVIDIA 3090 GPU).
    }
    \label{fig:teaster}
\end{figure}

\section{Introduction}
\label{sec:introduction}

% Some of the most beautiful appearances in our visual world are from specular objects like glossy paints, plastics, shiny metals or silken cloth.  
% Capturing these visual effects from real photographs for view synthesis requires properly handling both the spatial geometry and view-dependent appearance.
% While recent neural radiance field (NeRF)~\cite{mildenhall2020nerf} methods have made impressive progress on efficient spatial representation and encoding using learnable feature grids~\cite{liu2020neural,wu2022diver,sun2022direct,chen2022tensorf,muller2022instant,chan2022efficient},
% much less attention has been devoted to high-frequency angular-domain appearance modeling.
% Glossy objects not only show specular highlights, but also global illumination effects, reflecting other objects in the environment,
% such that a good directional encoding for these effects is as important as spatial encoding.
% We present a feature-grid-like \emph{neural directional encoding} that can efficiently model the specular effects with higher quality than current state of the art.

Some of the most compelling appearances in our visual world arise from specular objects like metals, plastics, glossy paints, or silken cloth.  
Faithfully reproducing these effects from photographs for novel-view synthesis requires capturing both geometry and view-dependent appearance.
Recent neural radiance field (NeRF)~\cite{mildenhall2020nerf} methods have made impressive progress on efficient geometry representation and encoding using learnable spatial feature grids ~\cite{liu2020neural,wu2022diver,sun2022direct,chen2022tensorf,muller2022instant,chan2022efficient}.
However, modeling high-frequency view-dependent appearance has achieved much less attention.
Efficient encoding of directional information is just as important, for modeling effects such as specular highlights and glossy interreflections.
In this paper, we present a feature-grid-like \emph{neural directional encoding} (\method) that can accurately model the appearance of shiny objects.

View-dependent colors in NeRFs (\eg~\cite{verbin2022ref}) are commonly obtained by decoding spatial features and encoded direction. This approach necessitates a large multi-layer perceptron (MLP) and exhibits slow convergence with analytical directional encoding functions.
To that end, we bring feature-grid-based encoding to the directional domain, representing reflections from distant sources via learnable feature vectors stored on a global environment map (\cref{subsec:direct-encoding}).
Features localize signal learning, reducing the MLP size required to model high-frequency far-field reflections.

Besides far-field reflections, spatially varying near-field interreflections are also key effects in rendering glossy objects.
These effects cannot be accurately modeled by NeRF's spatio-angular parameterization whose directional encoding does not depend on the position.
In contrast, we propose a novel spatio-spatial parameterization by \emph{cone-tracing a spatial feature grid} (\cref{subsec:indirect-encoding}) to encode near-field reflections. The cone tracing accumulates spatial encodings along the queried direction and position, thus it is spatially varying.
%In contrast, we propose a novel spatio-spatial parameterization by \emph{cone-tracing a spatial feature grid} (\cref{subsec:indirect-encoding}) to agument the directional encoding for interreflections,
% which is spatially varying and found less likely to overfit.
%These effects cannot be accurately modeled by NeRF’s spatio-angular parameterization that leads to incorrect extrapolations under novel views.
%In contrast, we propose a novel spatio-spatial parameterization by \emph{cone-tracing a spatial feature grid} (\cref{subsec:indirect-encoding}) to augment the directional encoding for interreflections and implicitly regularize the view dependent signal.
While prior works consider only single-bounce or diffuse interreflections~\cite{Liang2023ENVIDRID}, our representation is able to model general multi-bounce reflection effects.

Overall, our neural directional encoding (NDE) achieves both high-quality modeling of view-dependent effects and fast evaluation.
Figure~\ref{fig:teaster} demonstrates NDE incorporated into NeRF, showing (1)~accurate rendering of specular objects---a difficult challenge for the state of the art (\cref{subsec:view-synthesis}), and
(2)~high inference speed that can be pushed to real-time without obvious quality loss (\cref{subsec:real-time}).

\section{Related work}
\label{sec:related-work}

Novel-view synthesis aims to render a 3D scene from unseen views given a set of image captures with camera poses.
Neural radiance fields (NeRF)~\cite{mildenhall2020nerf} has recently emerged as a promising solution to this task, utilizing an implicit scene representation and volume rendering to synthesize photorealistic images.
Follow-up works achieve state-of-the-art results in this area, for unbounded scenes~\cite{zhang2020nerf++,barron2022mip}, in-the-wild captures~\cite{martin2021nerf}, and sparse- or single-view reconstruction~\cite{chen2021mvsnerf,wang2021ibrnet,lin2023vision,gu2023nerfdiff,trevithick2021grf,trevithick2023}.
While the original NeRF method~\cite{mildenhall2020nerf} is computationally inefficient, it can be visualized in real-time by baking the reconstruction into voxel- \cite{yu2021plenoctrees,garbin2021fastnerf,hedman2021baking,reiser2021kilonerf} or feature-grid-based representations (discussed below).
The volumetric representation has been extended to work with signed distance fields (SDF)~\cite{wang2021neus,yariv2021volume} for better geometry acquisition,
and the volume-rendering concept has also been applied to other 3D-related tasks such as object generation~\cite{chan2021pi,chan2022efficient,liu2023one,poole2022dreamfusion,lin2023magic3d}.

\paragraph{Feature-grid-based NeRF.}
\vspace{-2.5ex}
NeRF's positional encoding~\cite{mildenhall2020nerf} is a key component for the underlying multi-layer perceptron (MLP) network to learn high-frequency spatial and directional signals.
However, the MLP size needs to be large, which leads to slow training and inference.
Instead, methods like NSVF~\cite{liu2020neural} and DVGO~\cite{sun2022direct} interpolate a 3D volume of learnable feature vectors to encode the spatial signal, showing faster training and inference with even better spatial detail.
Addressing the sparsity in typical scene geometry, later works avoid maintaining a large dense 3D grid via volume-compression techniques such as hash grids~\cite{muller2022instant} and tensor factorization~\cite{chen2022tensorf, chan2022efficient, fridovich2023k}.
These methods are compact and scale up the feature grid to large scenes~\cite{muller2022instant, barron2023zipnerf} and even work with SDF-based models~\cite{yariv2023bakedsdf,li2023neuralangelo}.
The essence of feature-grid encoding is to interpolate feature vectors attached to geometry primitives,
and similar ideas have also been applied to irregular 3D grids~\cite{ rosu2023permutosdf, kulhanek2023tetra}, point clouds~\cite{xu2022point, kerbl20233d, keselman2023flexible,zhang2023nerflets}, and meshes~\cite{chen2023mobilenerf}.
Operations like mip-mapping are trivial on feature grids, enabling efficient anti-aliasing and range query of NeRF models~\cite{wu2022diver,hu2023tri,barron2023zipnerf}---something we also leverage in this paper to encode rough reflection.

\paragraph{Rendering specular objects.}
\vspace{-2.5ex}
Apart from geometry, view-dependent effects like reflections from rough surfaces are a crucial component in photorealistic novel-view synthesis. Reflections are conventionally modeled by fitting local light-field functions~\cite{flynn2019deepview,mildenhall2019local,kalantari2016learning}.
A 4D light field presents more degrees of freedom than the constraints from input images, which necessitates additional regularization to avoid overfitting.
Inverse-rendering approaches introduce such a constraint by solving for parametric BRDFs and lighting,
then using forward rendering to reconstruct the light field.
Spherical-basis lighting~\cite{zhang2021physg} or split-sum approximation~\cite{Munkberg_2022_CVPR, liu2023nero} are usually used to tamper the Monte Carlo variance of specular-reflection derivatives~\cite{belhe2023importance}.
ENVIDR~\cite{Liang2023ENVIDRID} and NMF~\cite{mai2023neural} further explicitly consider global-illumination effects by ray-tracing one or few bounces of indirect lighting.
On the other hand, Ref-NeRF~\cite{verbin2022ref} uses an integrated directional encoding (IDE) to directly improve NeRF's view-dependent effects.
IDE encodes the reflected direction rather than viewing direction to let the network learn an environment-map-like function and is pre-filtered to account for rough reflection effects.
Our neural directional encoding, similar to IDE, can model general view-dependent appearance without assuming simplified lighting or reflections but with smaller computation cost.
\begin{figure*}[t]
    \centering
    \includegraphics[width=\linewidth]{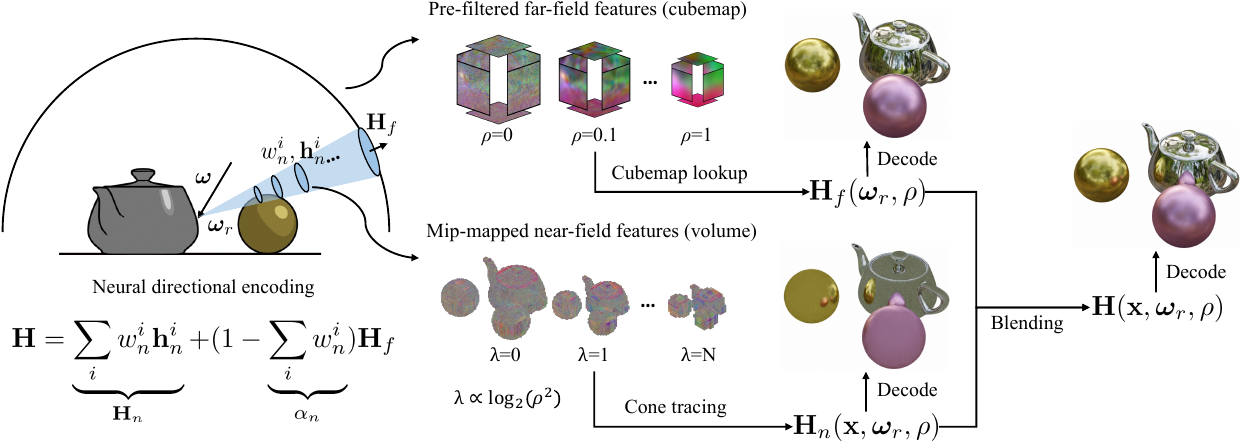}
    \caption{
        \textbf{Pipeline of our neural directional encoding (NDE).}
        We encode far-field reflections into a cubemap and near-field interreflections into a volume. Both representations store learnable feature vectors to encode direction and are mip-mapped to account for rough reflections. Given a reflected ray, the features are combined by tracing a cone of size proportional to the surface roughness to aggregate spatial features with cubemap features blended as the background. The result is fed into an MLP to output the specular color (\cref{eq:color-mlp}).
        %Two distinct representations store learnable feature vectors to encode direction information and are mip-mapped to model rough reflections. We encode far-field reflections into a global cubemap and spatially varying near-field reflections into a volume. Given a reflected ray, we trace a cone of size proportional to the surface roughness to aggregate spatial features which are blended with background cubemap features. The result is fed into an MLP to output the specular color (\cref{eq:color-mlp}).
    }
    \label{fig:pipeline}
\end{figure*}
\section{Preliminaries}
\label{sec:background}
We assume opaque objects with diffuse and specular components and demonstrate our directional encoding using a surface-based model that represents a scene using a signed distance field (SDF) $s(\mathbf{x})$ and a color field $\mathbf{c}(\mathbf{x},\bm{\omega})$ (dependent on the viewing direction $\bm{\omega}$).
The SDF is converted to NeRF's density field $\sigma$ 
following VolSDF~\cite{yariv2021volume}
with a learnable parameter $\beta$ controlling the boundary smoothness:
\begin{equation}
    \sigma(\mathbf{x}) = \begin{cases}
    \frac{1}{2\beta} \exp\left(\frac{s(\mathbf{x})}{\beta}\right) & \text{if } s(\mathbf{x}) \leq 0,\\
    \frac{1}{\beta} \left(1-\frac{1}{2}\exp\left(-\frac{s(\mathbf{x})}{\beta}\right)\right) & \text{otherwise}.
    \end{cases}
\label{eq:volsdf}
\end{equation}%
The color $\mathbf{C}(\mathbf{x},\bm{\omega})$ of a ray with origin $\mathbf{x}$ and direction $\bm{\omega}$
can thus be volume-rendered~\cite{max1995optical}:
\begin{gather}
    \mathbf{C}(\mathbf{x},\bm{\omega})\!=\!\sum_i
    w(\sigma(\mathbf{x}_i))\mathbf{c}(\mathbf{x}_i,\bm{\omega}),\ \text{where}\\
    w(\sigma(\mathbf{x}_i))=\left(1-e^{-\sigma(\mathbf{x}_i)\delta_i}\right)
    \prod_{j<i}e^{-\sigma(\mathbf{x}_j)\delta_j},
    \label{eq:volume-rendering}
\end{gather}%
with $\delta_i\!=\!\|\mathbf{x}_i\!-\!\mathbf{x}_{i-1}\|_2$ and $\mathbf{x}_i$ denoting the $i^\text{th}$ sample point along the ray.
Like Ref-NeRF~\cite{verbin2022ref},
we decompose the color $\mathbf{c}$ into a diffuse color $\mathbf{c}_d$, specular tint $\mathbf{k}_s$, and specular color $\mathbf{c}_s$ queried in reflected direction $\bm{\omega}_r$ with surface normal $\mathbf{n}$ given by the SDF gradient:
%derived from the SDF gradient:
\begin{equation}
\begin{gathered}
    \mathbf{c}(\mathbf{x},\bm{\omega})=\mathbf{c}_d(\mathbf{x})
        +\mathbf{k}_s(\mathbf{x})\mathbf{c}_s(\mathbf{x},\bm{\omega}_r), \quad\text{where}\\
        \bm{\omega}_r = \text{reflect}(\bm{\omega},\mathbf{n}),
        \quad
        \mathbf{n}=\text{normalize}(\nabla_\mathbf{x}s(\mathbf{x})).
\label{eq:color}
\end{gathered}
\end{equation}%
Here, the specular color $\mathbf{c}_s$ is decoded from an MLP that conditions on 
spatial feature $\mathbf{f}(\mathbf{x})$, 
directional encoding $\mathbf{H}$ controlled by surface roughness $\rho$, and the cosine term $\mathbf{n}\cdot\bm{\omega}$:
\begin{equation}
    \mathbf{c}_s(\mathbf{x},\bm{\omega}_r) = \text{MLP}(
          \mathbf{f}(\mathbf{x}),
          \mathbf{H}(\mathbf{x},\bm{\omega}_r,\rho(\mathbf{x})),
          \mathbf{n}\cdot\bm{\omega}).
    \label{eq:color-mlp}
\end{equation}%
$\mathbf{c}_d,\mathbf{k}_s,\mathbf{f},\rho$ come from a spatial MLP (\cref{subsec:optimization}).

\paragraph{Discussion on directional encoding.}
Previous works~\cite{verbin2022ref,mildenhall2020nerf} use an analytical function for 
$\mathbf{H}$ dependent only on $\bm{\omega}_r$ (and optionally $\rho$), which has several limitations:
(1)~the encoding function is fixed (not learnable), 
and (2)~the spatial context only comes from $\mathbf{f}(\mathbf{x})$.
Both require the decoder MLP to be large to fit the spatio-angular details of the specular color,
which can be expensive and slow.
%(1)~the encoding is fixed, such that only MLP parameters control the angular capacity; 
%and (2)~there is no spatial context.
%(1)~the encoding is fixed, thus the angular capacity solely depends on that of the decoding MLP;
%and (2)~there is a lack of spatial context for the encoding. 
%As a result, a large network is required to fit specular color details, which can be expensive and slow.
%Note that our encoding $\mathbf{H}$ depends not just on angular direction but also on spatial location,
%which is important for handling interreflection effects as discussed in~\cref{subsec:indirect-encoding}.
\section{Neural directional encoding}
\label{sec:method}
To minimize the MLP complexity,
we use a learnable neural directional encoding that also depends on the spatial location.
Specifically, our NDE encodes different types of reflection by different representations,
which include a cubemap feature grid $\mathbf{h}_f$ for far-field reflections and a spatial volume $\mathbf{h}_n$ that models near-field interreflections.
As shown in \cref{fig:pipeline}, we compute $\mathbf{H}$ by first cone-tracing $\mathbf{h}_n$ accumulated along the reflected ray, yielding near-field feature $\mathbf{H}_n$ (\cref{subsec:indirect-encoding}),
and blending the far-field feature $\mathbf{H}_f$ queried from $\mathbf{h}_f$ in the same direction (\cref{subsec:direct-encoding}):
\begin{equation}
    \mathbf{H}(\mathbf{x},\bm{\omega}_r,\rho) =
    \mathbf{H}_n(\mathbf{x},\bm{\omega}_r,\rho)
    +(1-\alpha_n)\mathbf{H}_f(\bm{\omega}_r,\rho),
\label{eq:nde}
\end{equation}%
where $\alpha_n$ is the cone-traced opacity~\cite{li2022nerfacc},
and both features are mip-mapped with $\rho$ deciding the mip level.

\subsection{Far-field features}
\label{subsec:direct-encoding}
Feature-grid-based representations~\cite{chen2022tensorf,muller2022instant,liu2020neural,wu2022diver,sun2022direct} speed-up spatial signal learning by storing feature vectors in voxels for local signal control.
Similarly, we place feature vectors $\mathbf{h}_f$ at every pixel of a global cubemap to encode ideal specular reflections.
The cubemap is pre-filtered to model reflections under rough surfaces in the split-sum~\cite{Karis:2013} style, where the $k^\text{th}$ level mip-map $\mathbf{h}_f^k$ is created by convolving the downsampled $\mathbf{h}_f$ using a GGX kernel~\cite{walter2007microfacet} $D$ with canonical roughness $\rho_k$ evenly spaced in $[0,1]$:
\begin{equation}
        \mathbf{h}_f^k=\text{convolution}(\text{downsample}(\mathbf{h}_f,k), D(\rho_k)).
    \label{eq:direct-feature}
\end{equation}%
Given the surface roughness,
we perform a cubemap lookup in the reflected direction and interpolate between mip levels to get the far-field feature:
\begin{equation}
    \mathbf{H}_f(\bm{\omega}_r,\rho)=\text{lerp}\left(\mathbf{h}^k_f(\bm{\omega}_r),\mathbf{h}^{k+1}_f(\bm{\omega}_r),\frac{\rho\!-\!\rho_k}{\rho_{k+1}\!-\!\rho_k}\right),
\label{eq:direct_feature_lerp}
\end{equation}%
\mbox{where $\text{lerp}(\cdot)$ denotes linear interpolation and $\rho\in[\rho_k,\rho_{k+1}]$}.

\begin{figure}[t]
    \centering
    \setlength\tabcolsep{1.0pt}
    \resizebox{0.99\linewidth}{!}{
        \begin{tabular}{cccc}
            \includegraphics[width=0.3\linewidth]{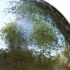}&
            \includegraphics[width=0.3\linewidth]{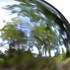}&
            \includegraphics[width=0.3\linewidth]{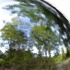}&
            \includegraphics[width=0.3\linewidth]{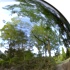}\\
            IDE small & IDE large & $\mathbf{H}_f$ \textbf{small (ours)} & Ground truth
        \end{tabular}
    }
    \caption{
        \textbf{Our cubemap-based feature encoding} requires only a small MLP (2 layers, 64 width) to model details in mirror reflections (3rd image) comparable with IDE~\cite{verbin2022ref} (2nd image; 8 layers, 256 width MLP) that fails when the MLP is small (1st image).
    }
    \label{fig:direct}
\end{figure}

The cubemap-based encoding allows signals in different directions to be optimized independently by tuning the feature vectors.
This is easier to optimize than globally solving the MLP parameters,
making it more suitable to model high-frequency details in the angular domain (\cref{fig:direct}).
%\liwen{
The coarse level feature is a consistently filtered version of the fine level,
which is empirically found to be better constrained than using independent feature vectors at each mip level~\cite{kuznetsov2021neumip,zeltner2023real}.
%}

%\sai{I don't think we can claim convergence faster since we we don't validate this and we still require a long training time.}

\begin{figure}[t]
    \centering
    \setlength\tabcolsep{5.0pt}
    \resizebox{0.99\linewidth}{!}{
    \begin{tabular}{ccc}
        \includegraphics[width=0.35\linewidth]{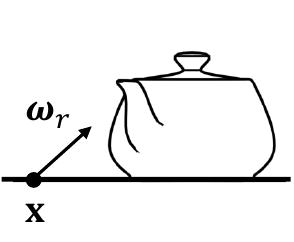}&
        \includegraphics[width=0.35\linewidth]{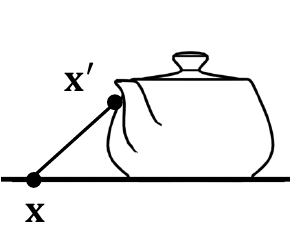}&
        \includegraphics[width=0.35\linewidth]{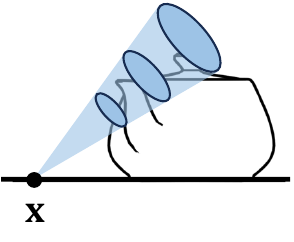}\\[-1mm]
        Spatio-angular&
        Spatio-spatial&
        Cone-traced\\[1mm]
    \end{tabular}
    }
    \caption{
        \textbf{Spatio-spatial encoding} (middle) is equivalent to the common spatio-angular encoding (left) of mirror reflections, but it captures the variation of $\mathbf{x}'$ across different $\mathbf{x}$.
        The idea can be extended to model rough reflections by cone tracing mip-mapped spatial features covered by the reflection cone (right).
        %\kai{can we make volume rendering of features along indirect rays more explicit in the figure or caption?}
        %\iliyan{I would suggest the opposite: the cone tracing is extension, and I'd argue is optional, for filtering}
    }
    \label{fig:spatio-spatial}
\end{figure}
\subsection{Near-field features}
\label{subsec:indirect-encoding}
%\iliyan{sentences are too long, consider breaking them up}
%\kai{One general writing tip: consider giving the symbols an intuitive name, and spell the symbols' names at least once at the beginning of each section to facilitate reading}
%\kai{Consider adding a sentence at the beginning: we integrate features along indirect rays using volume rendering to use as near-field feature}
Parameterizing the specular color by a spatial and angular feature is sufficient for distant reflections,
but lacks expressivity for near-field interreflections:
different points query the same $\mathbf{h}_f$,
so spatially varying components can end up being averaged out during optimization.
Our insight is that the spatio-angular reflection can also be parameterized as a spatio-spatial function of current and next bounce location (\cref{fig:spatio-spatial}).
Therefore, an MLP can decode the second bounce spatial feature with $\mathbf{f}(\mathbf{x})$ in \cref{eq:color-mlp} to get mirror reflections.

For rough reflections, we aggregate the averaged second bounce feature under the reflection lobe by cone tracing~\cite{crassin2011interactive} (\cref{fig:spatio-spatial}, right),
%\iliyan{unclear what this density is, also why it's different from $\sigma$}
which volume renders the mip-mapped spatial features $\mathbf{h}_n$ using the mip-mapped density $\sigma_n$ along the reflected ray $\mathbf{x}\!+\!\bm{\omega}_rt$ with mip level $\lambda_i\!=\!\log_2(2r_i)$ at sample point $\mathbf{x}^\prime_i$ decided by the cone’s footprint $r_i\!=\!\sqrt{3}\rho^2\|\mathbf{x}-\mathbf{x}^\prime_i\|_2$:
\begin{equation}
\begin{gathered}
    \label{eq:indirect}
    \mathbf{H}_n(\mathbf{x},\bm{\omega}_r,\rho)
    =\sum_iw_n^i\mathbf{h}_n^i, \quad\text{where}\\
    w^i_n=w(\sigma_n(\mathbf{x}'_i,\lambda_i)),\quad
    \mathbf{h}^i_n=\mathbf{h}_n(\mathbf{x}'_i,\lambda_i).
\end{gathered}
\end{equation}%
The cone’s footprint is selected to cover the GGX lobe at $\mathbf{x}$ (see supplemental document).
Note that we do not use the SDF-converted $\sigma$ in \cref{eq:volsdf} as it cannot be mip-mapped; 
instead, we optimize a separate $\sigma_n$ to match $\sigma$ (\cref{subsec:optimization}) jointly with the indirect feature 
$\mathbf{h}_n$.
Both are decoded from a tri-plane~\cite{chan2022efficient} $\mathbf{T}_n$,
whose each 2D plane is mip-mapped similar to Tri-MipRF~\cite{hu2023tri}:
\begin{equation}
    \!\!\!\sigma_n(\mathbf{x}'_i,\lambda_i),\mathbf{h}_n(\mathbf{x}'_i,\lambda_i)
    \!=\!\text{MLP}(\text{mipmap}(\mathbf{T}_n(\mathbf{x}'_i),\lambda_i)).\!
\end{equation}%
%\fujun{A bit confusing to me on the multi-bounce part. Are you saying the feature encodes radiance field in it, so it's naturally supporting multi-bounce illumination (whereas second-bounce colors cannot achieve that)? Alternatively, one can use NeRF to approximate indirect lighting (like in InvRender~\cite{zhang2022modeling} and TensorIR~\cite{jin2023tensoir}).}

\begin{figure}[t]
    \centering
    \setlength\tabcolsep{5.0pt}
    \resizebox{0.99\linewidth}{!}{
    \begin{tabular}{ccc}
    \includegraphics[width=0.35\linewidth]{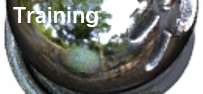}&
    \includegraphics[width=0.35\linewidth]{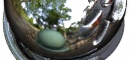}&
    \includegraphics[width=0.35\linewidth]{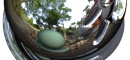}\\
    \includegraphics[width=0.35\linewidth]{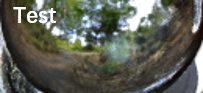}&
    \includegraphics[width=0.35\linewidth]{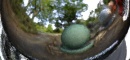}&
    \includegraphics[width=0.35\linewidth]{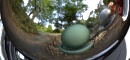}\\
    \textbf{Ours without} $\mathbf{H}_n$ & \textbf{Ours with} $\mathbf{H}_n$ & Ground truth
    \end{tabular}
    }
    \caption{
    \textbf{Our cone-traced near-field features} successfully reconstruct the reflected spheres (2nd column) under novel views, which are overfitted by the angular-only encoding (1st column).
    }
    \label{fig:indirect}
\end{figure}

%~\fujun{Since $\mathbf{x}'_i$ varies across different $\mathbf{x}$?}, ~\kai{Maybe it's more straightforward to say that the indirect rays are spatially varying, hence volume-rendered near-field features are spatially varying too?}
%Since $\mathbf{x}'_i$ for each $\mathbf{x}$ are different,
%the spatio-spatial parameterization provides a spatially varying encoding,
The indirect rays are spatially varying,
hence the cone-traced near-field features are spatially varying too.
This has advantages over the angular-only feature for learning interreflections and is empirically less likely to overfit (\cref{fig:indirect}).
This is because the same $\mathbf{h}_n$ is traced from different rays in training, such that the underlying representation is well-constrained.
$\mathbf{H}_n$ and $\mathbf{H}_f$ are similar to the foreground and background colors in regular volume rendering,
so $\mathbf{H}_f$ can be naturally composited with $\mathbf{H}_n$ using the opacity $\alpha_n\!=\!1-\prod_i e^{-\sigma_n(\mathbf{x}'_i,\lambda_i)\delta_i}\!=\!\sum_i w_n^i$ as in \cref{eq:nde}.

% \begin{figure}[t]
%     \centering
%     \setlength\tabcolsep{1.0pt}
%     \includegraphics[width=1.0\linewidth]{images/architecture}
%     \caption{
%     \textbf{Network architectures.} N$\times$M denotes an M layer MLP of width N.
%     }
%     \label{fig:architectures}
% \end{figure}

\begin{figure}[t]
    \centering
    \setlength\tabcolsep{1.0pt}
    \includegraphics[width=1.0\linewidth]{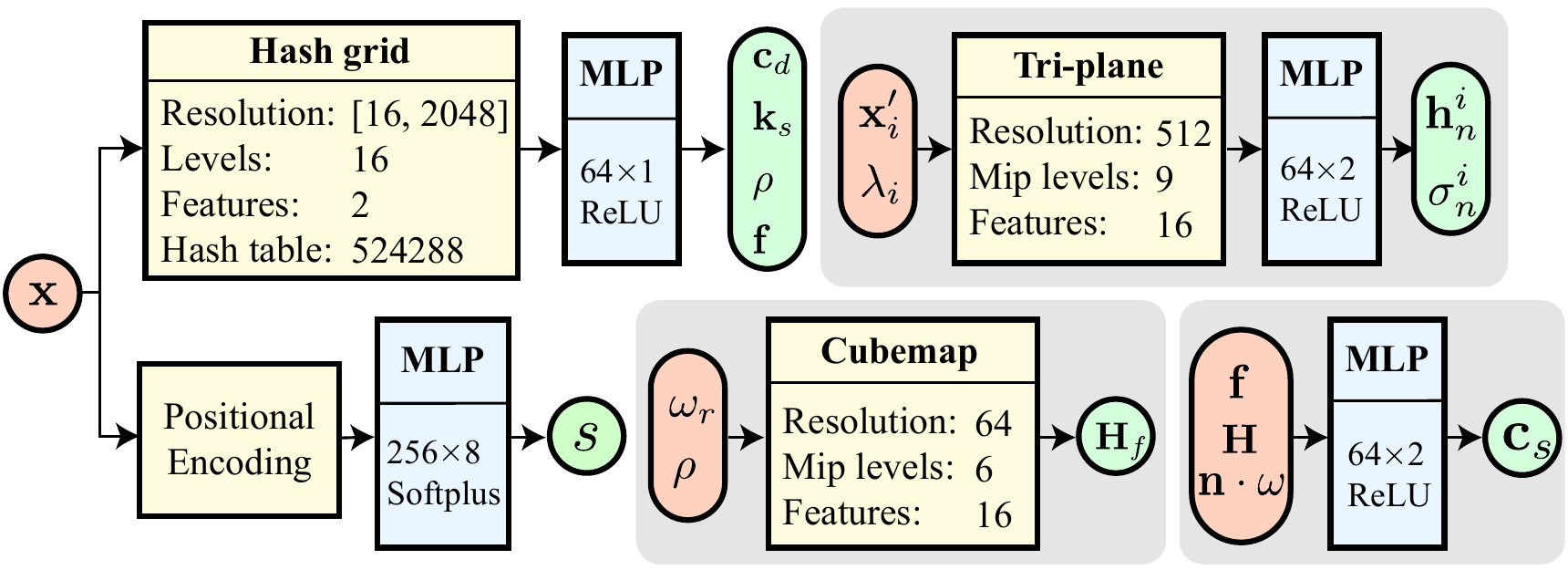}
    \caption{
    \textbf{Network architectures.} $N\!\times\!M$ denotes an $M$-layer MLP of width $N$.
    }
    \label{fig:architectures}
\end{figure}

\subsection{Optimization}
\label{subsec:optimization}
Figure~\ref{fig:architectures} shows our network architectures.
Stable geometry optimization is essential for modeling specular objects,
so we use the positional-encoded MLP from VolSDF~\cite{yariv2021volume} to output the SDF.
To reduce computation cost,
a hash grid is used to encode other spatial features ($\mathbf{c}_d,\mathbf{k}_s,\rho,\mathbf{f}$),
and all other MLPs are tiny.
The representation is optimized through the Charbonnier loss~\cite{barron2022mip} between ground truth pixel color $\mathbf{C}_\text{gt}$ and our rendering $\mathbf{C}$ in tone-mapped space:
\begin{equation}
    L = \sum_{\mathbf{x},\bm{\omega}} 
    \sqrt{\left\|
    \Gamma(\mathbf{C}(\mathbf{x},\bm{\omega}))-\mathbf{C}_\text{gt}(\mathbf{x},\bm{\omega})
    \right\|_2^2+0.001},
\label{eq:loss-photo}
\end{equation}%
where $\Gamma$ is the tone-mapping function~\cite{Munkberg_2022_CVPR}.
%\iliyan{tone mapping is not applied to the ground-truth color?}\liwen{yes}

\paragraph{Occupancy-grid sampling.}
\vspace{-2.5ex}
\cref{eq:volume-rendering,eq:indirect} are accelerated by an occupancy-grid estimator~\cite{li2022nerfacc} to get rid of computations in empty space.
This is especially important for the efficient near-field feature evaluation, 
since we trace a reflected ray for each primary ray sample.
%since for every primary ray sample we also trace a reflected ray.
The primal ray rendering uses a fixed ray marching step of 0.005.
Following~\cite{crassin2011interactive}, we choose the cone tracing step proportional to its footprint: $\max{(0.5r_i,0.005)}$, and query a mip-mapped occupancy grid for the correct occupancy information.

\paragraph{Regularization.}
\vspace{-2.5ex}
Given the primary samples $\mathbf{x}_i$, 
Eikonal loss~\cite{yariv2021volume} $L_\text{eik}$ is applied to regularize the SDF,
and we implicitly regularize $\sigma_n$ to match $\sigma$ by encouraging the rendering using $\sigma_n$ at mip level 0 to be close to the ground truth:
\begin{equation}
\begin{gathered}
    L_\sigma\!=\!\sum_{\mathbf{x},\bm{\omega}} \|
        \mathbf{C}_\sigma(\mathbf{x},\bm{\omega})
        -\mathbf{C}_\text{gt}(\mathbf{x},\bm{\omega})
    \|_2^2, \quad \text{where}\\
\mathbf{C}_\sigma(\mathbf{x},\bm{\omega})=\sum_iw(\sigma_n(\mathbf{x}_i,0))\mathring{\mathbf{c}}(\mathbf{x}_i,\bm{\omega}),
\end{gathered}
    \label{eq:loss-reg}
\end{equation}%
$\mathring{\square}$ denotes stop-gradient to prevent $\sigma_n$ affecting appearance.
The total loss is $L+0.1L_\text{eik}+0.01L_\sigma$.

\paragraph{Implementation details.}
\vspace{-2.5ex}
We implement our code using PyTorch~\cite{paszke2019pytorch}, NerfAcc~\cite{li2022nerfacc}, and CUDA.
The optimization takes 400k steps using the Adam optimizer~\cite{kingma2014adam} with 0.0005 learning rate and dynamic batch size~\cite{muller2022instant} targeting for 32k primary point samples.
We use the scheduler from BakedSDF~\cite{hedman2021baking} to anneal $\beta$ in \cref{eq:volsdf} for more stable convergence.
Because the SDF uses a positional-encoded MLP,
each scene still requires 10$\sim$18 hours to train on an NVIDIA 3090 GPU with 15GB GPU memory usage.
\section{Experiments}
\label{sec:experiments}
We evaluate our method on view synthesis of specular objects using synthetic and real scenes.
The synthetic scenes include the Shinny Blender dataset~\cite{verbin2022ref} and the Materials scene from the NeRF Synthetic dataset~\cite{mildenhall2020nerf},
all rendered without background;
the real scenes come from NeRO~\cite{liu2023nero} which contain backgrounds and reflections of the capturer in the images.
The rendering quality is compared in terms of PSNR, SSIM~\cite{wang2004image}, LPIPS~\cite{zhang2018unreasonable},
and the inference speed in FPS is recorded on an NVIDIA 3090 GPU.

\paragraph{Background and capturer.}
\vspace{-2.5ex}
For real scenes, we use a separate Instant-NGP~\cite{muller2022instant} with coordinate contraction~\cite{barron2022mip} to render backgrounds.
Similarly to NeRO~\cite{liu2023nero}, the reflection of the capturer is encoded by blending a capturer plane feature $\mathbf{h}_c$ of opacity $\alpha_c$ between $\mathbf{H}_f$ and $\mathbf{H}_n$:
\begin{equation}
\begin{gathered}
    \mathbf{H}=\mathbf{H}_n
    +(1-\alpha_n)(\alpha_c\mathbf{h}_c+(1-\alpha_c)\mathbf{H}_f), \; \text{where}\\
    \alpha_c,\mathbf{h}_c=\text{MLP}(
    \text{mipmap}(\mathbf{T}_c(\mathbf{u}),\lambda_c))
\end{gathered}
\label{eq:camera-feature}
\end{equation}%
are decoded from a mip-mapped 2D feature grid $\mathbf{T}_c$;
$\mathbf{u},\lambda_c$ are the ray-plane intersection coordinate and the mip-level derived from the intersection footprint.
Jointly optimizing foreground and background networks can be unstable,
so we apply stabilization loss from NeRO~\cite{liu2023nero} and modify the specular color computation for the first 200k steps:
$\mathbf{h}_f,\mathbf{h}_n,\mathbf{h}_c$ are sampled and decoded into colors first,
then the colors are blended to get $\mathbf{c}_s$.
Compared to blending the feature and decoding, 
we find the decoding-then-blending strategy provides better geometry optimization.

\begin{table}[t]
    \centering
    \setlength\tabcolsep{2 pt}
    \resizebox{0.99\linewidth}{!}{
    \begin{tabular}{l c c c c c c c c}
    \toprule
    \textbf{Method} & \textbf{Mat.} & \textbf{Teapot} & \textbf{Toaster} & \textbf{Car} & \textbf{Ball} & \textbf{Coffee} & \textbf{Helmet} & \textbf{Mean} \\
    \midrule
    \multicolumn{9}{c}{\textbf{PSNR} $\uparrow$}\\
    \midrule
NeRO &24.85 & 40.29 & \snd{27.31} & 26.98 & 31.50 & 33.76 & 29.59 & 30.61 \\
ENVIDR &29.51 & 46.14 & 26.63 & 29.88 & 41.03 & \snd{34.45} & \snd{36.98} & 34.95 \\
Ref-NeRF &\fst{35.41} & \snd{47.90} & 25.70 & \fst{30.82} & \fst{47.46} & 34.21 & 29.68 & \snd{35.88} \\
\textbf{NDE (ours)} &\snd{31.53} & \fst{49.12} & \fst{30.32} & \snd{30.39} & \snd{44.66} & \fst{36.57} & \fst{37.77} & \fst{37.19} \\
    \midrule
    \multicolumn{9}{c}{\textbf{SSIM} $\uparrow$}\\
    \midrule
NeRO &0.878 & 0.993 & 0.891 & 0.926 & 0.953 & 0.960 & 0.953 & 0.936 \\
ENVIDR &0.971 & \fst{0.999} & \snd{0.955} & \fst{0.972} & \fst{0.997} & \fst{0.984} & \fst{0.993} & \fst{0.982} \\
Ref-NeRF &\fst{0.983} & 0.998 & 0.922 & 0.955 & \snd{0.995} & 0.974 & 0.958 & \snd{0.969} \\
\textbf{NDE (ours)} &\snd{0.972} & \fst{0.999} & \fst{0.968} & \snd{0.968} & \snd{0.995} & \snd{0.979} & \snd{0.990} & \fst{0.982} \\
    \midrule
    \multicolumn{9}{c}{\textbf{LPIPS} $\downarrow$}\\
    \midrule
NeRO &0.138 & 0.017 & 0.162 & 0.064 & 0.179 & 0.099 & 0.102 & 0.109 \\
ENVIDR &0.026 & \snd{0.003} & 0.097 & \snd{0.031} & \fst{0.020} & \snd{0.044} & \snd{0.022} & \snd{0.035} \\
Ref-NeRF &\snd{0.022} & 0.004 & \snd{0.095} & 0.041 & 0.059 & 0.078 & 0.075 & 0.053 \\
\textbf{NDE (ours)} &\fst{0.017} & \fst{0.002} & \fst{0.039} & \fst{0.024} & \snd{0.022} & \fst{0.033} & \fst{0.014} & \fst{0.022} \\
\bottomrule
    \end{tabular}
    }
        \caption{\textbf{Quantitative comparison on synthetic scenes} showing our encoding (NDE) is either the \fst{best} or \snd{second best} compared to other methods for view synthesis of specular objects.
    }
    \label{tab:synthetic}
\end{table}

\begin{figure*}[t]
    \centering
    \setlength\tabcolsep{0.5pt}
    \resizebox{1.0\linewidth}{!}{
    \begin{tabular}{c cccc cccc}
& ENVIDR~\cite{Liang2023ENVIDRID}&
Ref-NeRF~\cite{verbin2022ref}&
\textbf{NDE (ours)}&
Ground truth &
ENVIDR&
Ref-NeRF&
\textbf{NDE (ours)}&
GT\\
 & \includegraphics[trim={230 110 150 270},clip,width=0.2\linewidth]{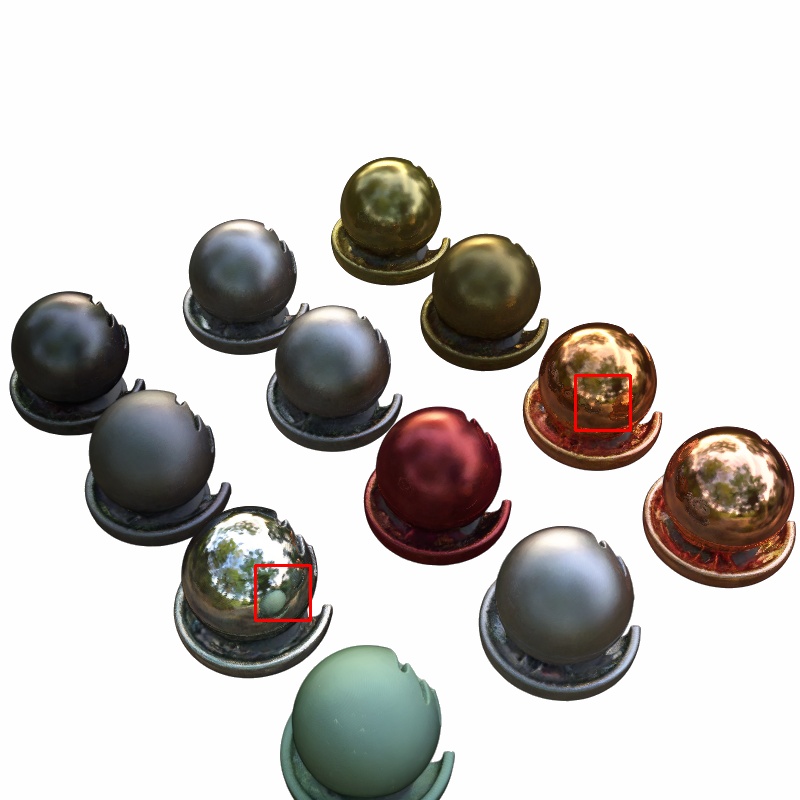} &
\includegraphics[trim={230 110 150 270},clip,width=0.2\linewidth]{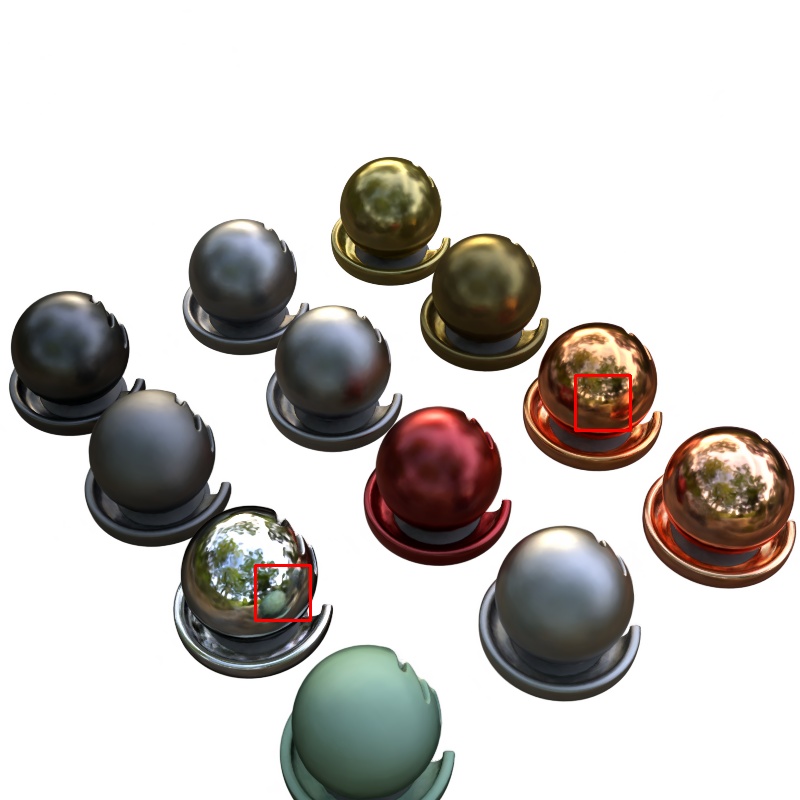} &
\includegraphics[trim={230 110 150 270},clip,width=0.2\linewidth]{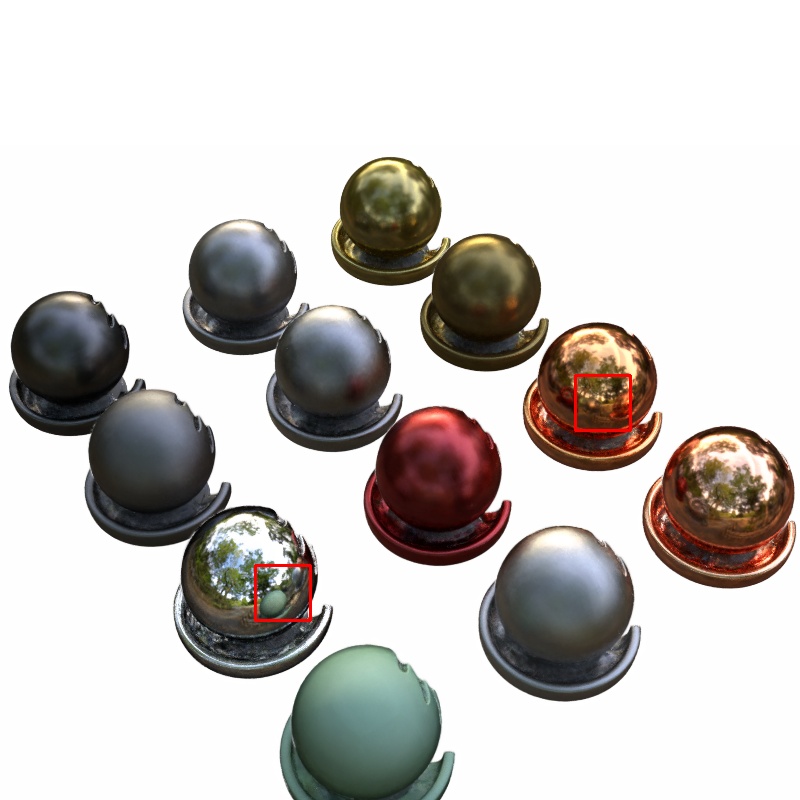} &
\includegraphics[trim={230 110 150 270},clip,width=0.2\linewidth]{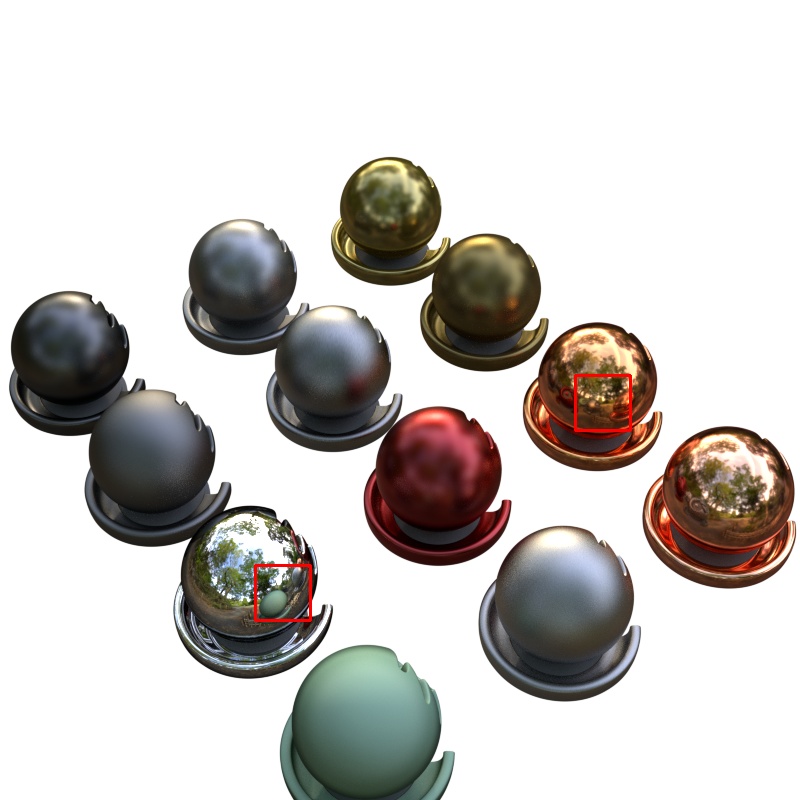} &
\includegraphics[width=0.1\linewidth]{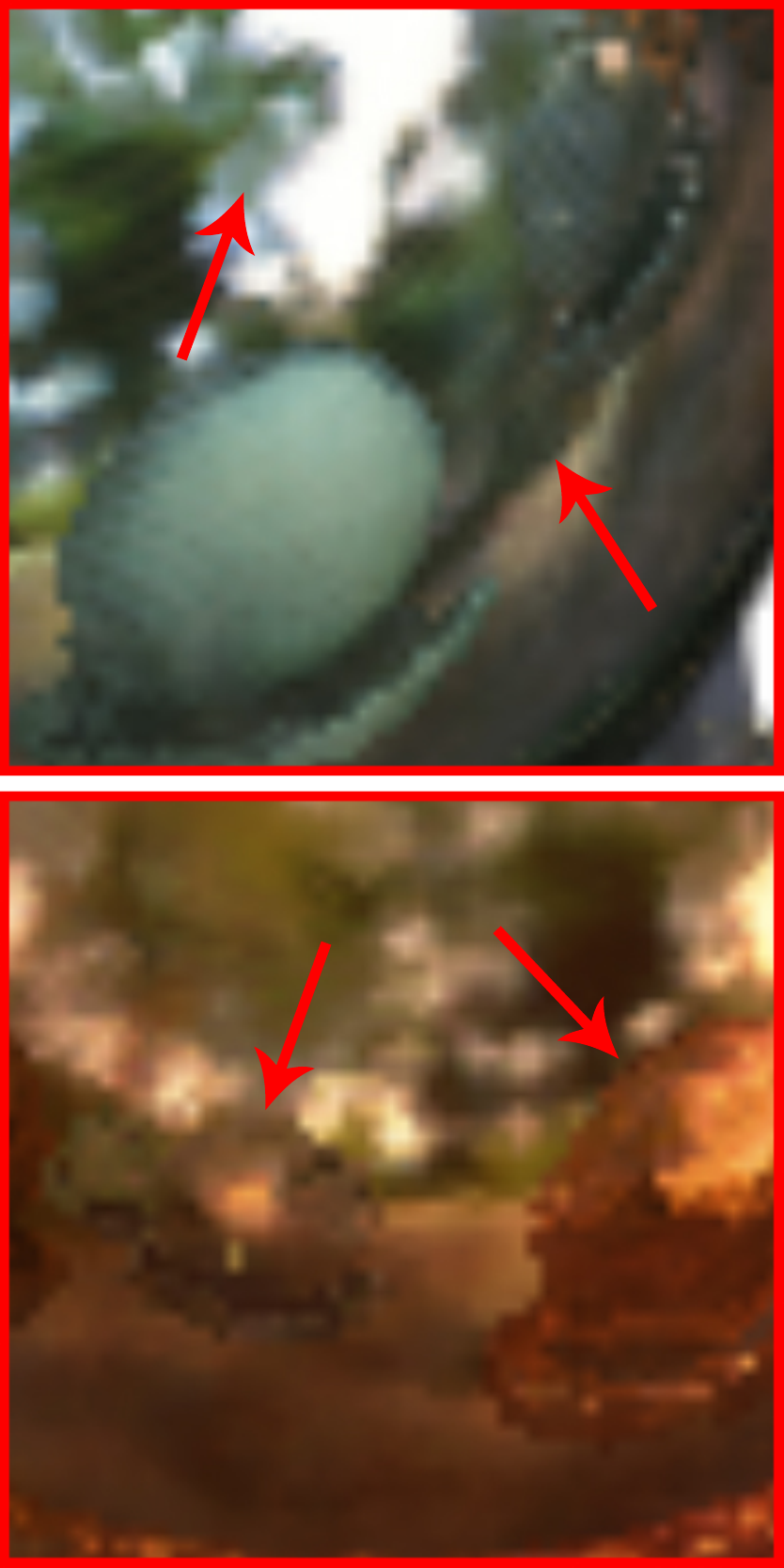} &
\includegraphics[width=0.1\linewidth]{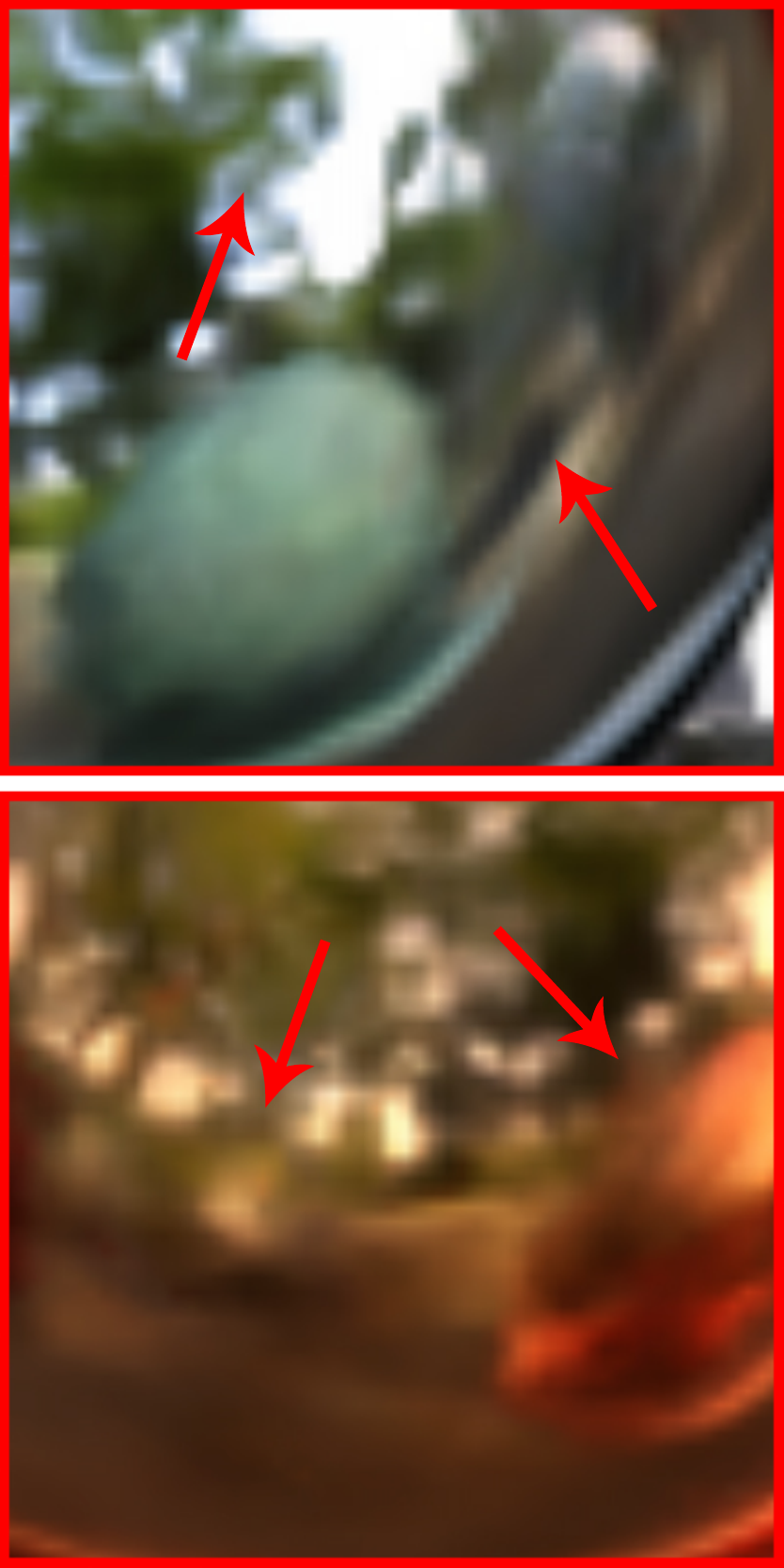} &
\includegraphics[width=0.1\linewidth]{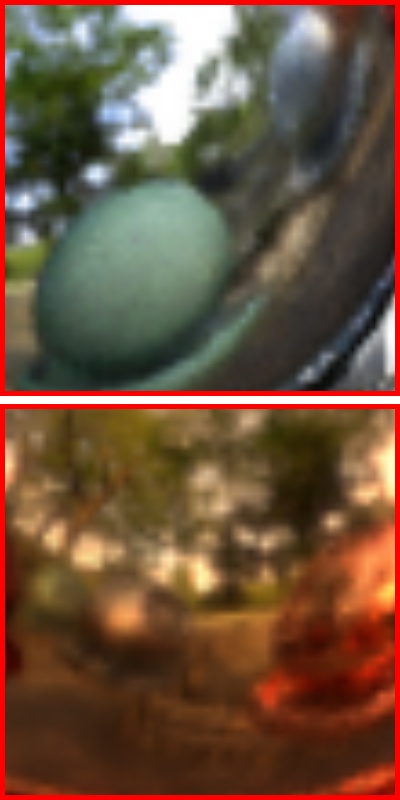} &
\includegraphics[width=0.1\linewidth]{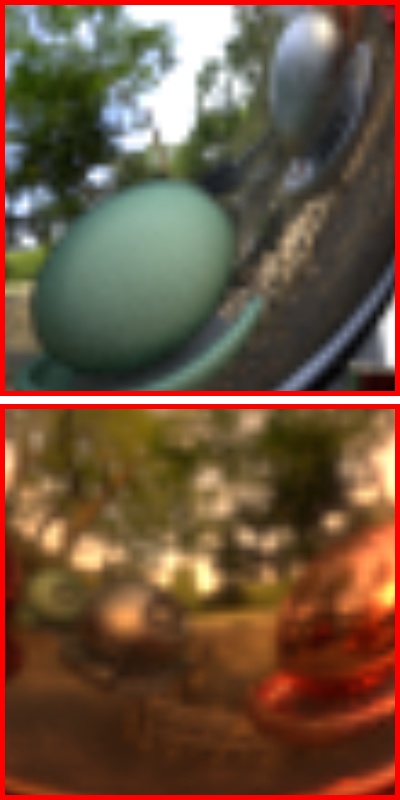} 
\\
\raisebox{1.75\height}{\rotatebox[origin=c]{90}{Rendering}}&
\includegraphics[trim={150 200 150 100},clip,width=0.2\linewidth]{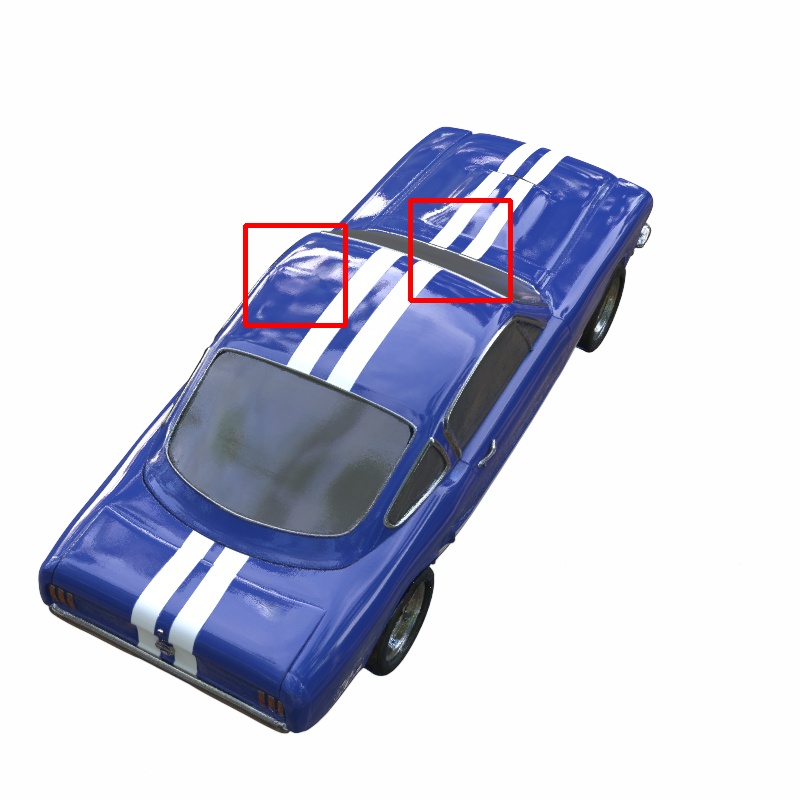} &
\includegraphics[trim={150 200 150 100},clip,width=0.2\linewidth]{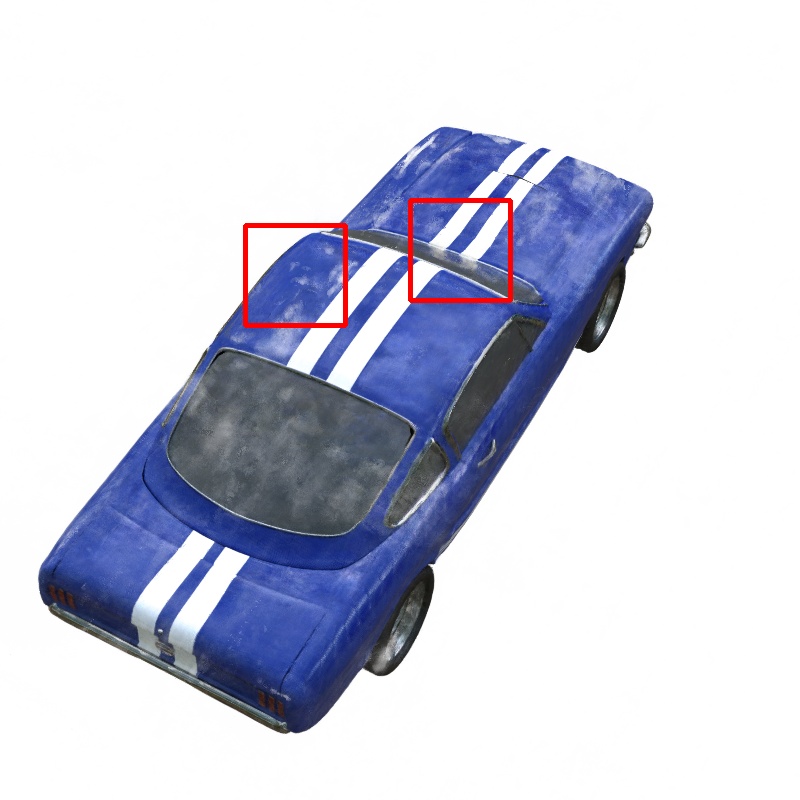} &
\includegraphics[trim={150 200 150 100},clip,width=0.2\linewidth]{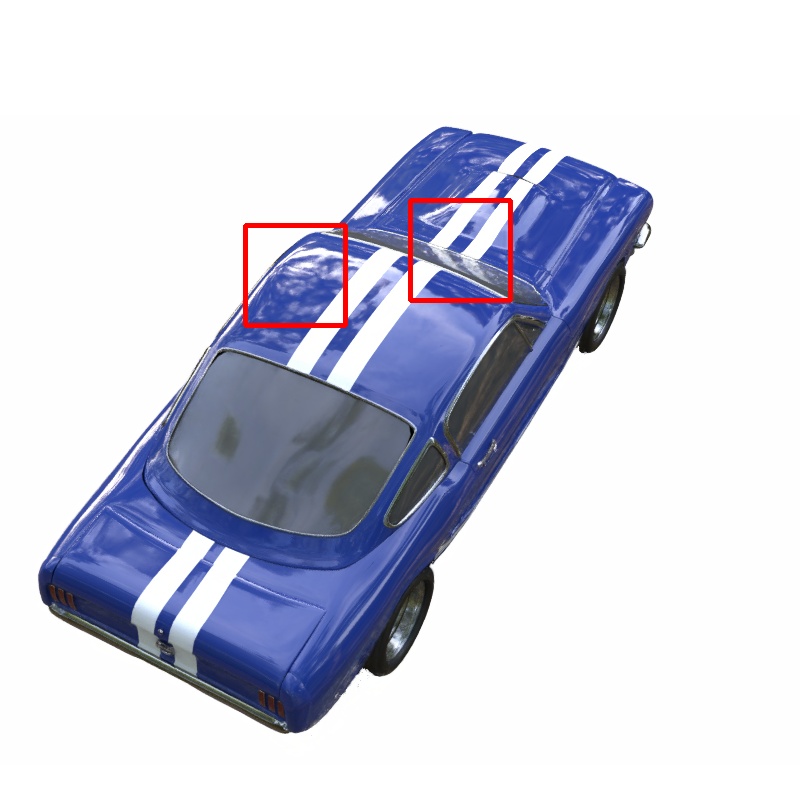} &
\includegraphics[trim={150 200 150 100},clip,width=0.2\linewidth]{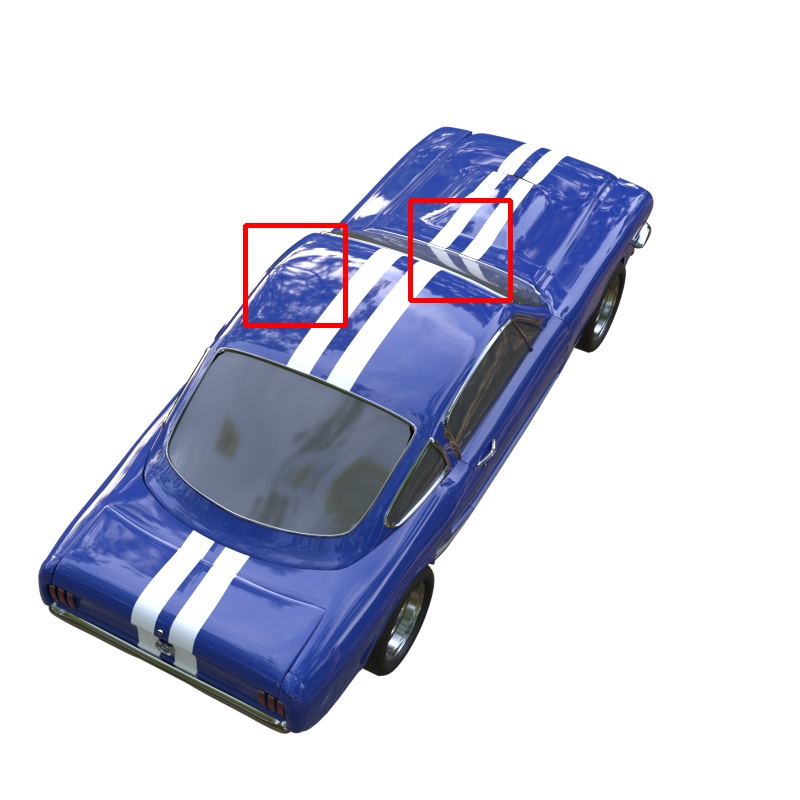} &
\includegraphics[width=0.1\linewidth]{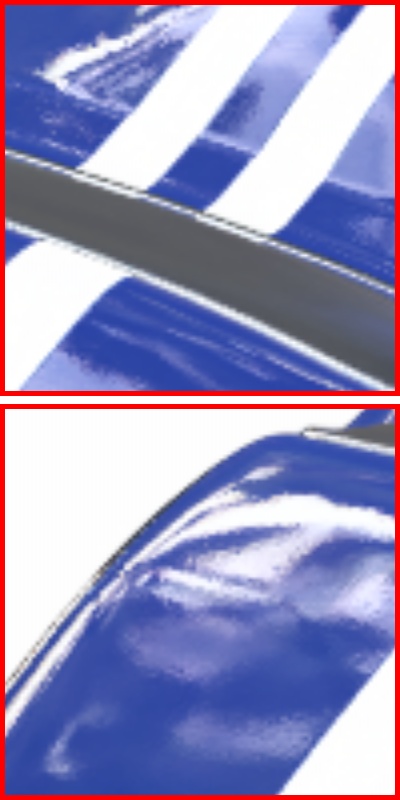} &
\includegraphics[width=0.1\linewidth]{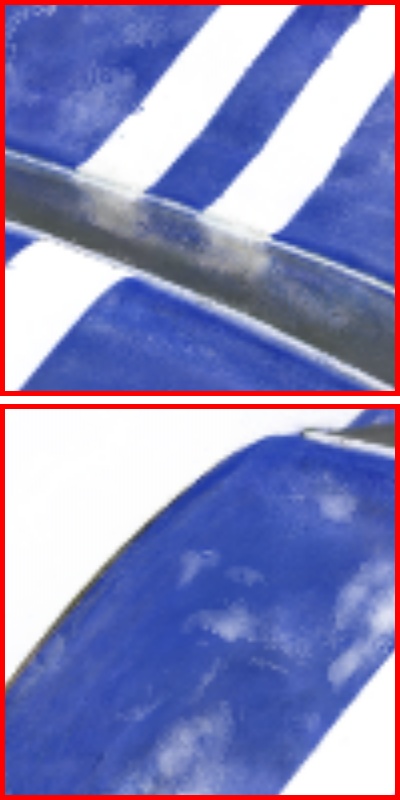} &
\includegraphics[width=0.1\linewidth]{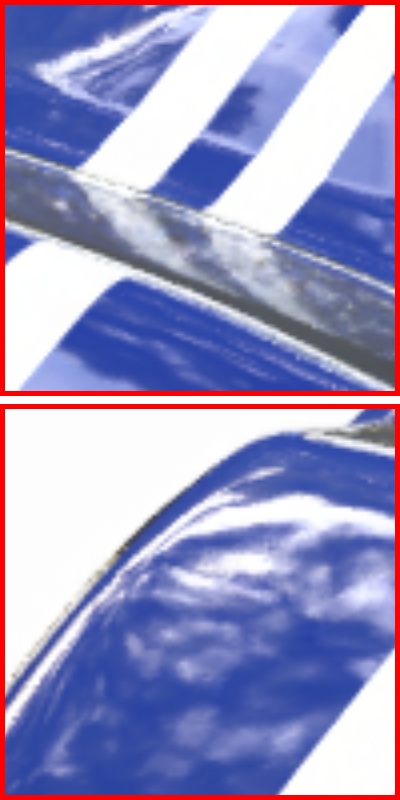} &
\includegraphics[width=0.1\linewidth]{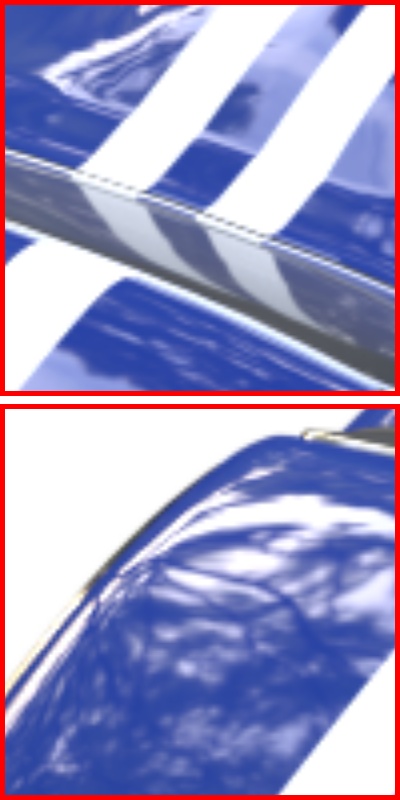} 
\\
&
\includegraphics[trim={180 200 120 100},clip,width=0.2\linewidth]{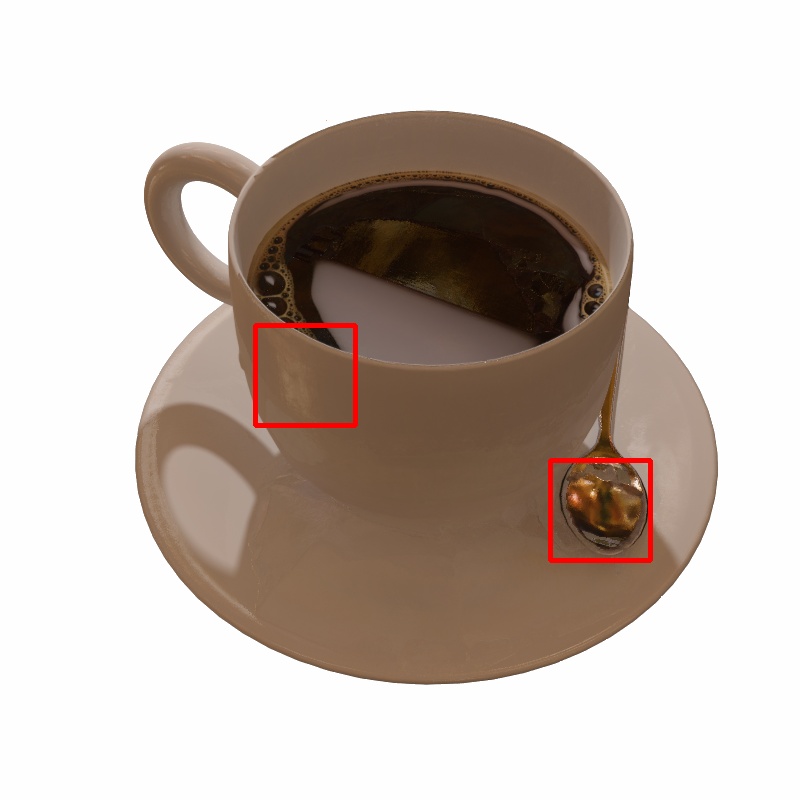} &
\includegraphics[trim={180 200 120 100},clip,width=0.2\linewidth]{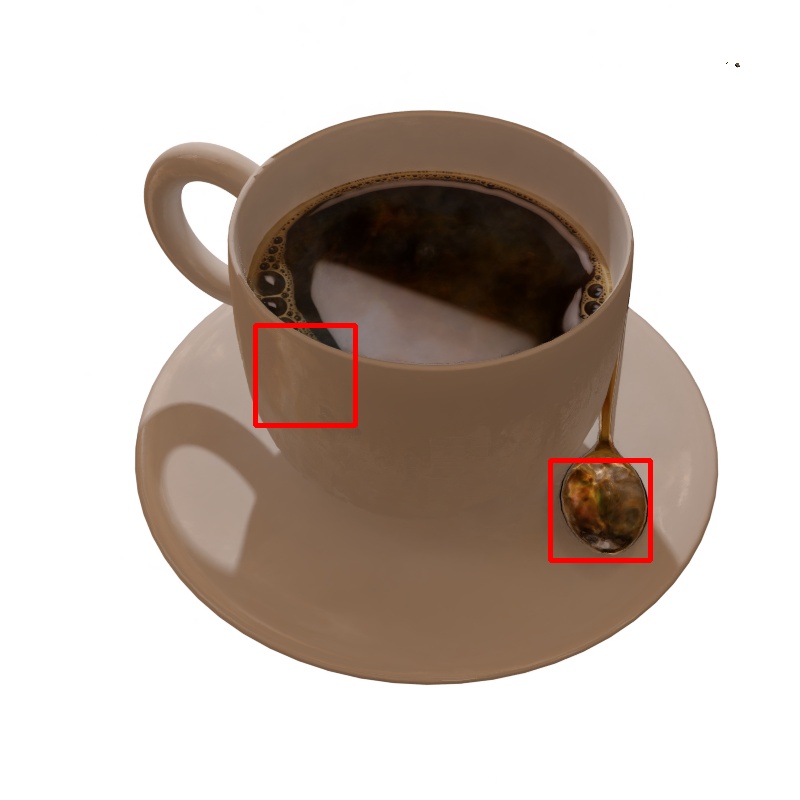} &
\includegraphics[trim={180 200 120 100},clip,width=0.2\linewidth]{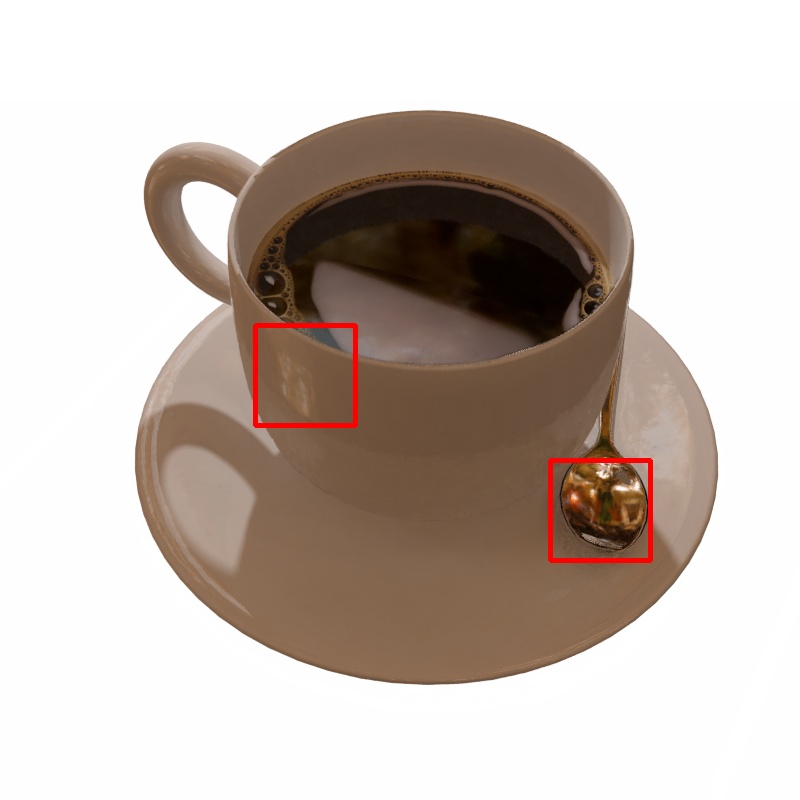} &
\includegraphics[trim={180 200 120 100},clip,width=0.2\linewidth]{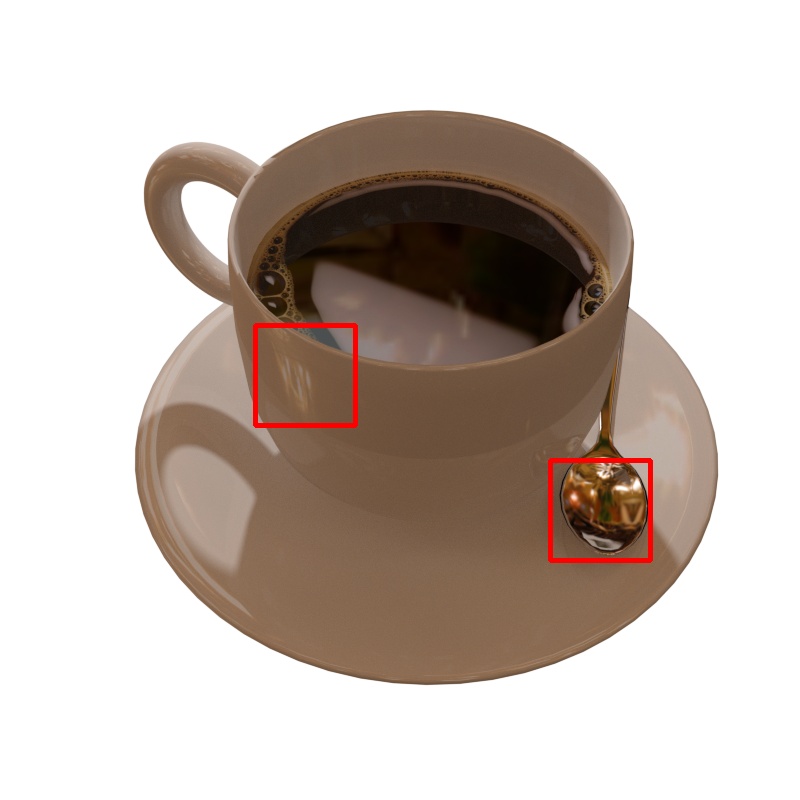} &
\includegraphics[width=0.1\linewidth]{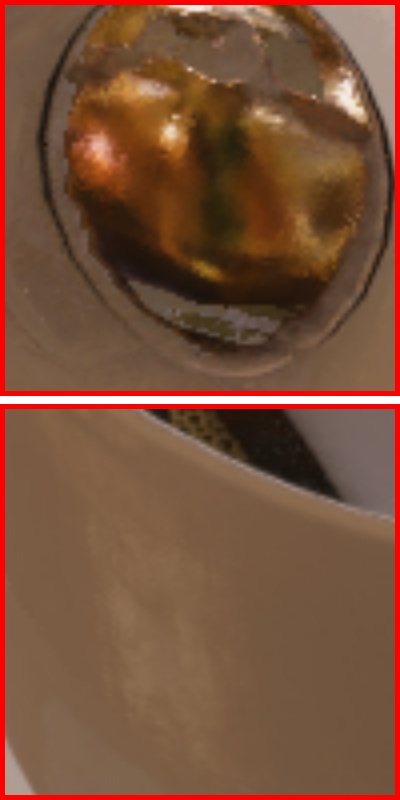} &
\includegraphics[width=0.1\linewidth]{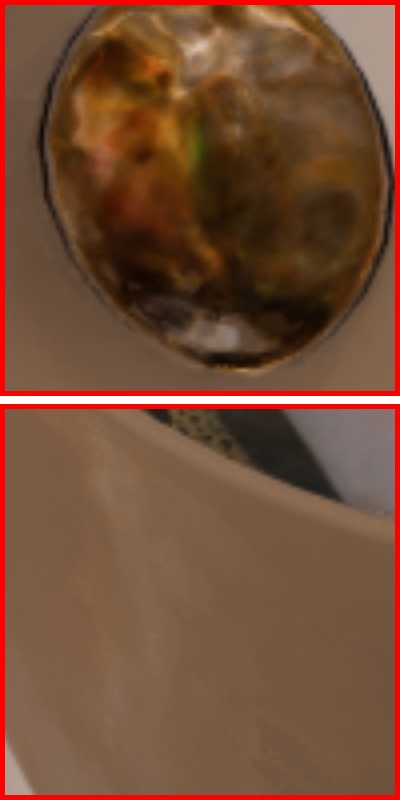} &
\includegraphics[width=0.1\linewidth]{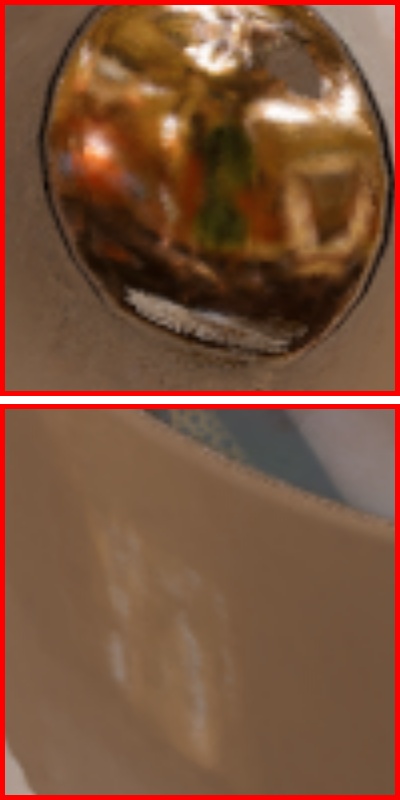} &
\includegraphics[width=0.1\linewidth]{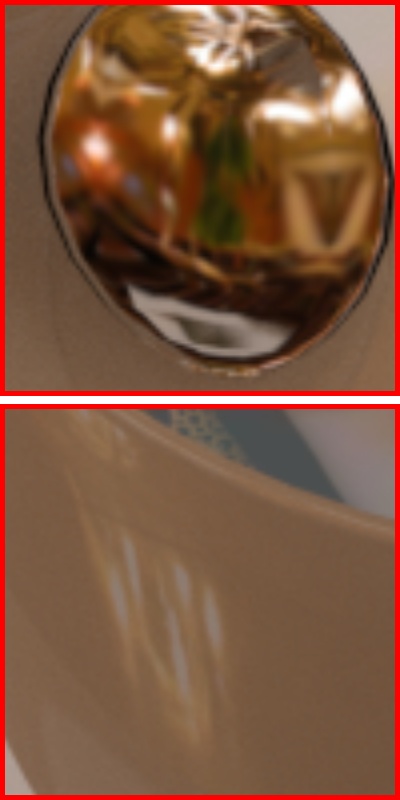} \\
\hline\\[-3mm]
\raisebox{1.1\height}{\rotatebox[origin=c]{90}{Rendering/Normal}} &
\includegraphics[trim={50 30 50 70},clip,width=0.2\linewidth]{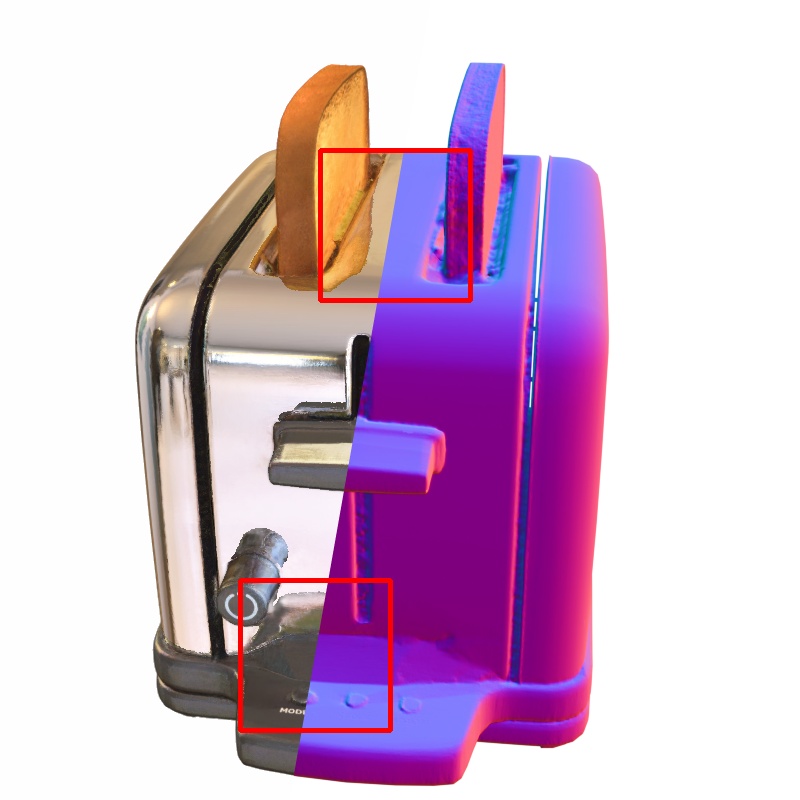} &
\includegraphics[trim={50 30 50 70},clip,width=0.2\linewidth]{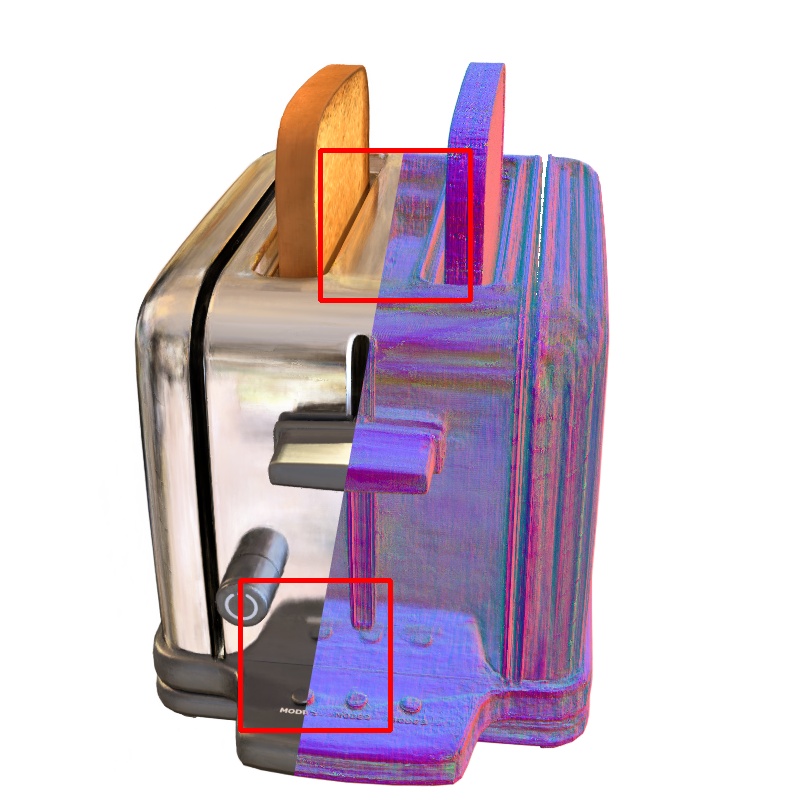} &
\includegraphics[trim={50 30 50 70},clip,width=0.2\linewidth]{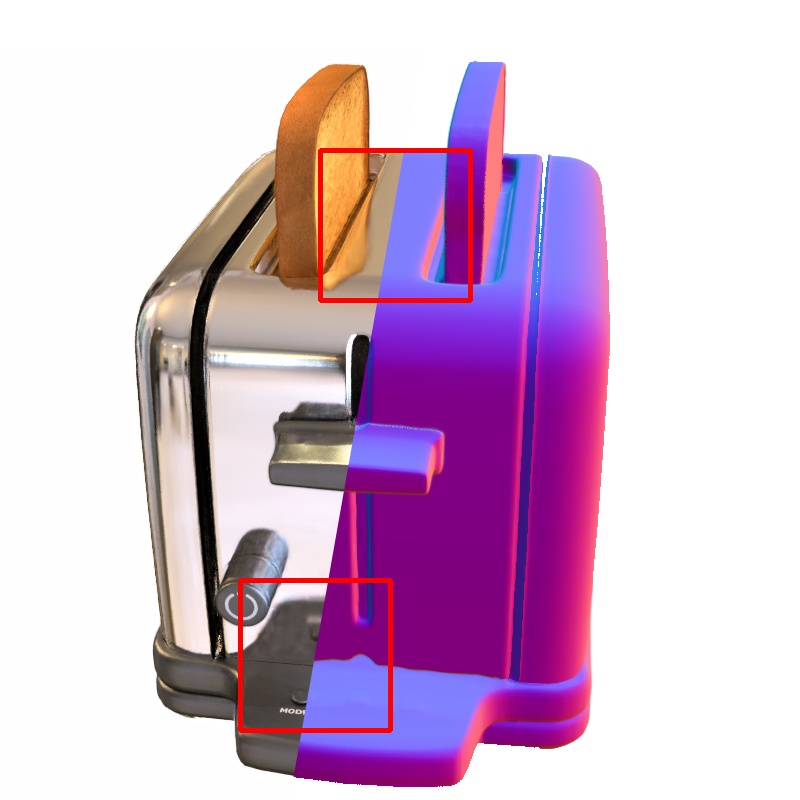} &
\includegraphics[trim={50 30 50 70},clip,width=0.2\linewidth]{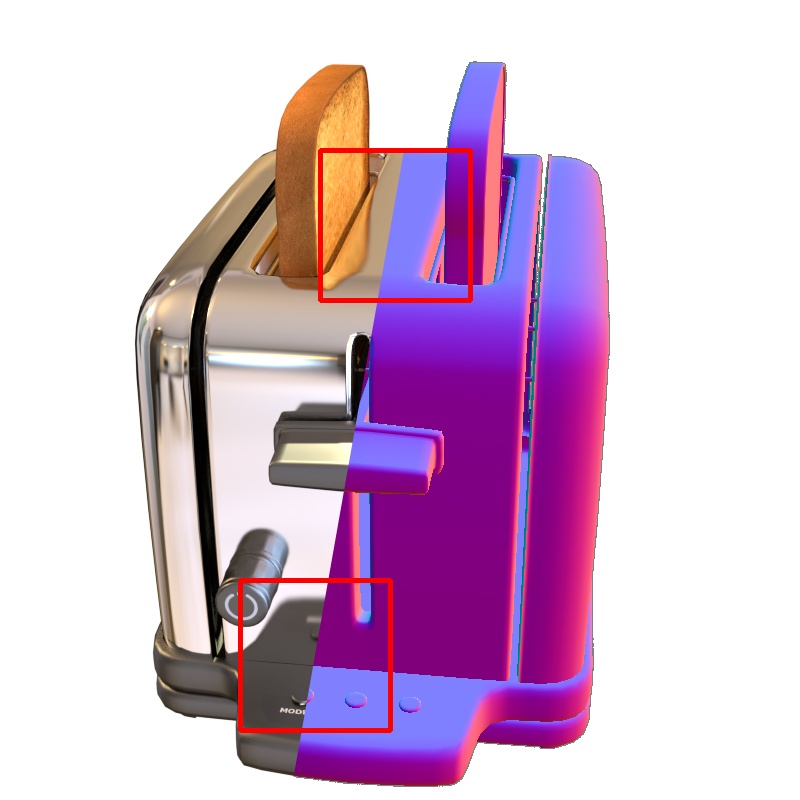} &
\includegraphics[width=0.1\linewidth]{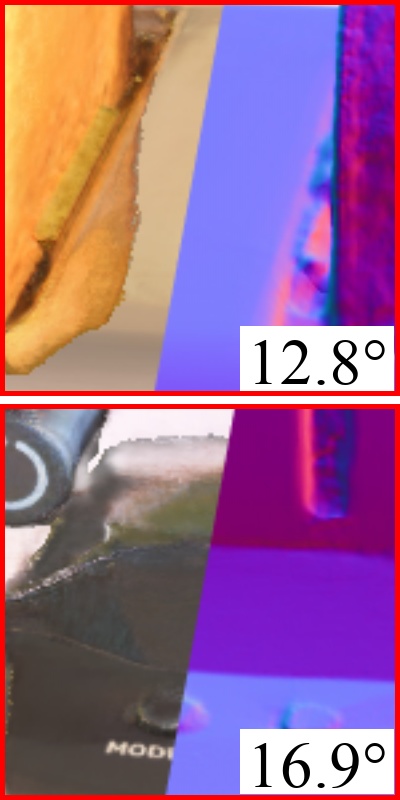} &
\includegraphics[width=0.1\linewidth]{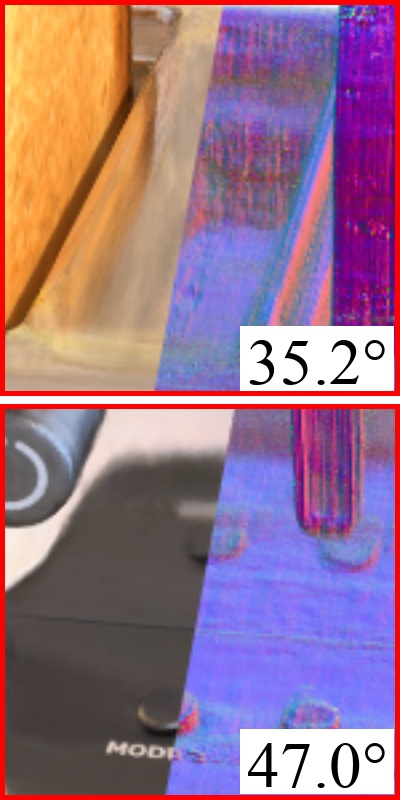} &
\includegraphics[width=0.1\linewidth]{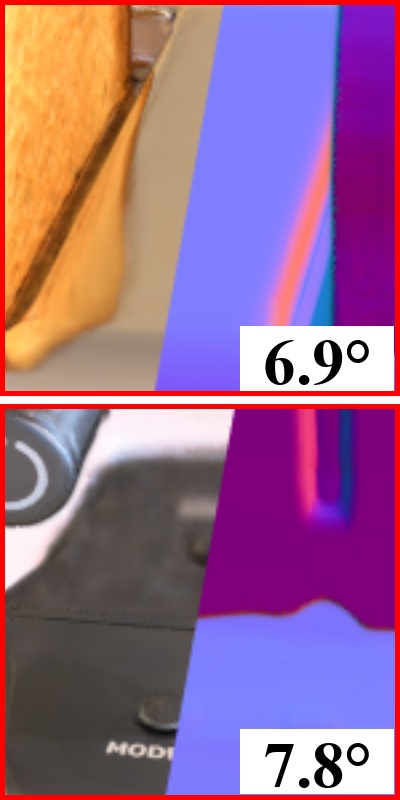} &
\includegraphics[width=0.1\linewidth]{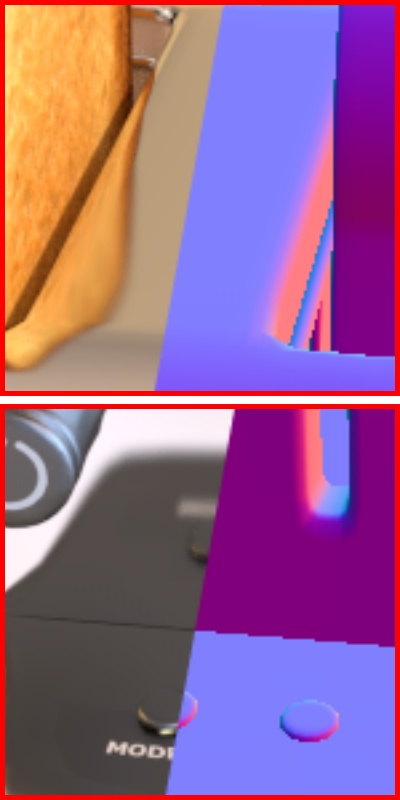}\\

    \end{tabular}
    }
    \caption{\textbf{Qualitative results for synthetic scenes} show our NDE successfully models the fine details of reflections from both environment lights (mirror sphere and car top) and other objects (glossy interreflections on spheres; zoom in to see the difference).
    Ref-NeRF tends to use wrong geometry to fake interreflections (2nd column on bottom).
    In contrast, our encoding has sufficient capacity to model interreflections, which enables more accurate normals (3rd column on bottom).
    Mean angular error of the normal is shown in the insets.
    }
    \label{fig:synthetic}
\end{figure*}
\subsection{View synthesis}
\label{subsec:view-synthesis}

%\paragraph{Baselines.}
%\vspace{-2.5ex}
We compare against NeRO~\cite{liu2023nero}, 
ENVIDR~\cite{Liang2023ENVIDRID}, and Ref-NeRF~\cite{verbin2022ref} on synthetic scenes.
All methods except for Ref-NeRF use SDFs,
and we evaluate NeRO after the BRDF estimation as it shows better performance.
Ideally, both backgrounds and reflections from the capturer should be removed when evaluating renderings of specular objects, which is difficult for the real scenes.
%Therefore, we only show qualitative results of real scenes compared to NeRO with PSNR from manually annotated image patches.
Therefore, we only qualitatively compare real scenes against NeRO with PSNR computed on the foreground zoom-ins without the capturer.
%\iliyan{Not very clear: You mean that we compare real scenes only against NeRO and with PSNR computed on the zoom-ins?}

\begin{figure*}[t]
    \centering
    \setlength\tabcolsep{1.0pt}
    \resizebox{1.0\linewidth}{!}{
    \begin{tabular}{ccc}
        %NeRO~\cite{liu2023nero} &NDE &Ground truth&
        NeRO~\cite{liu2023nero} & \textbf{NDE (ours)} &Ground truth\\

        \includegraphics[width=0.34\linewidth]{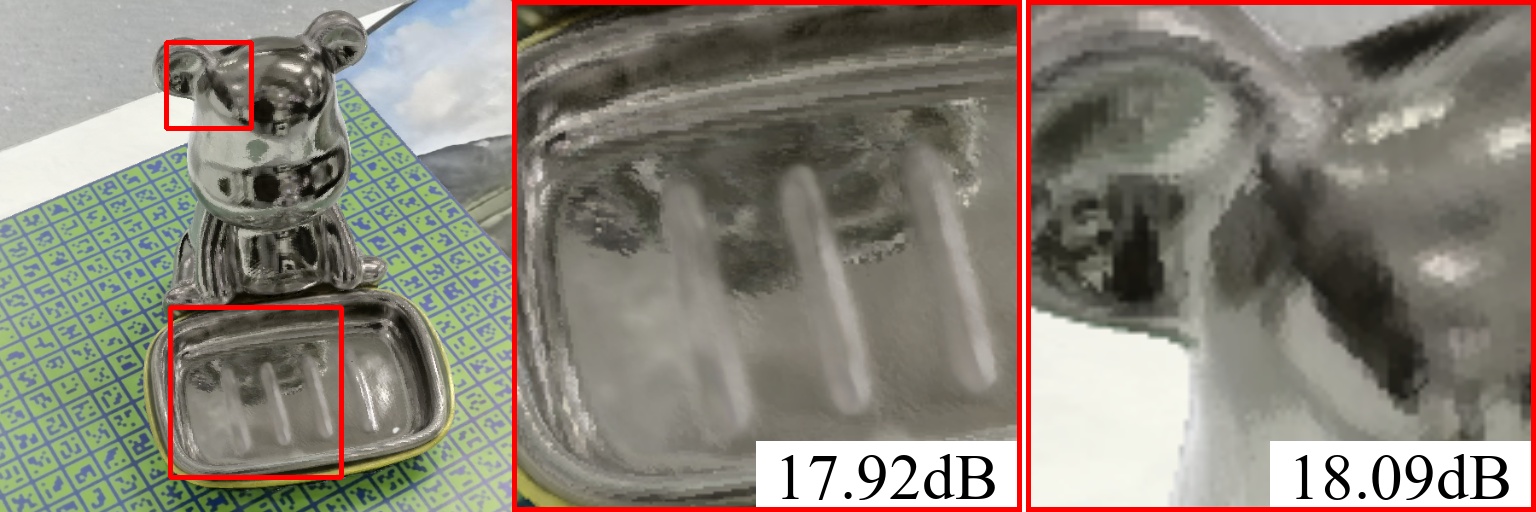}&
        \includegraphics[width=0.34\linewidth]{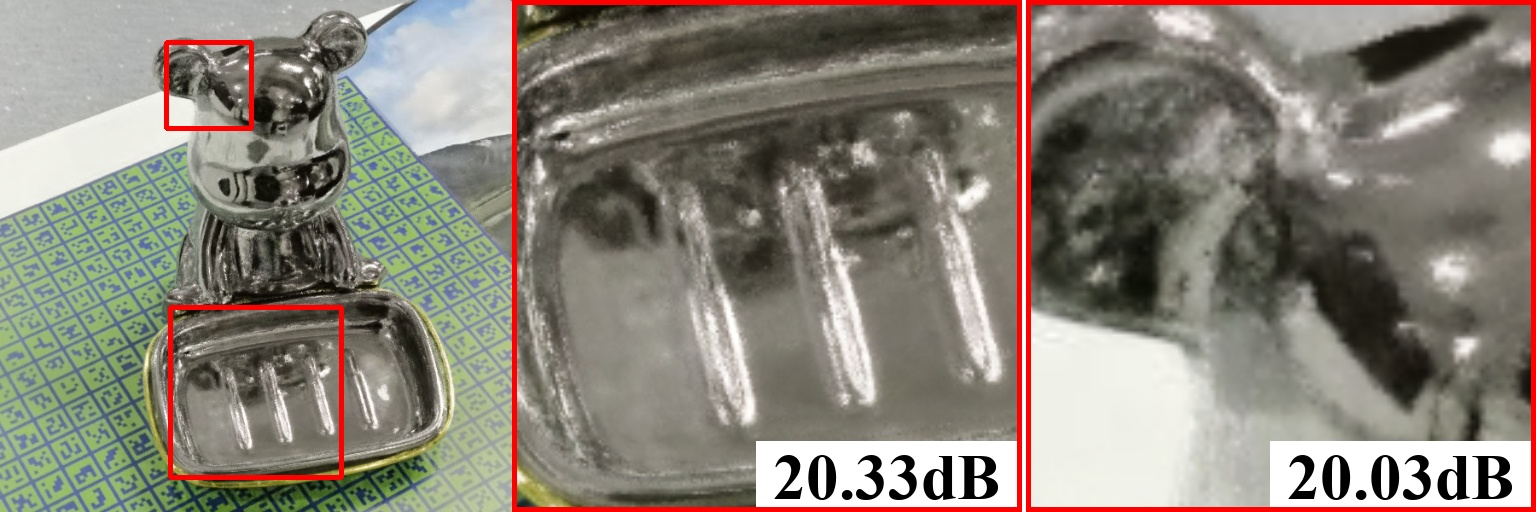}&
        \includegraphics[width=0.34\linewidth]{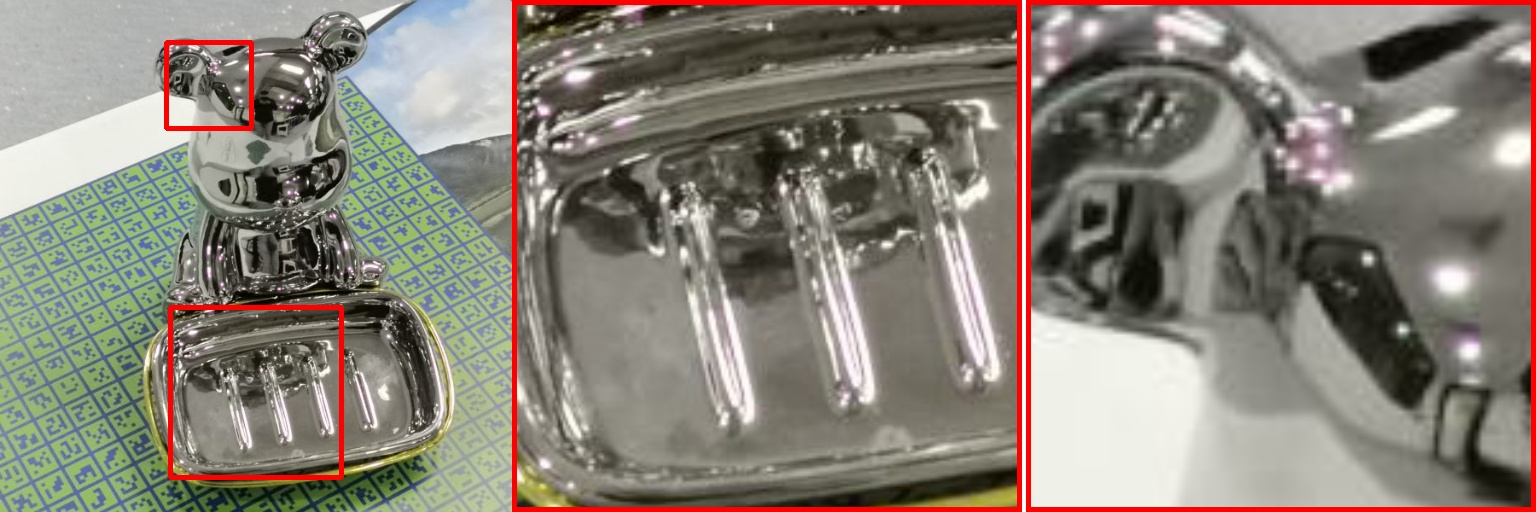}\\

        \includegraphics[width=0.34\linewidth]{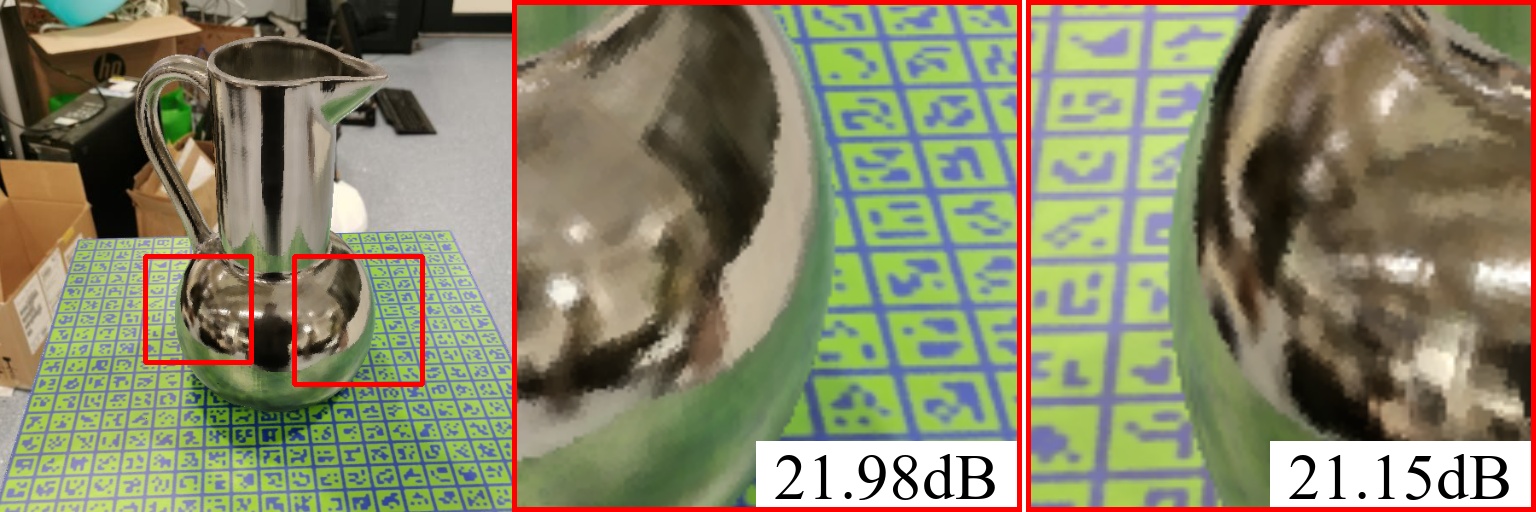}&
        \includegraphics[width=0.34\linewidth]{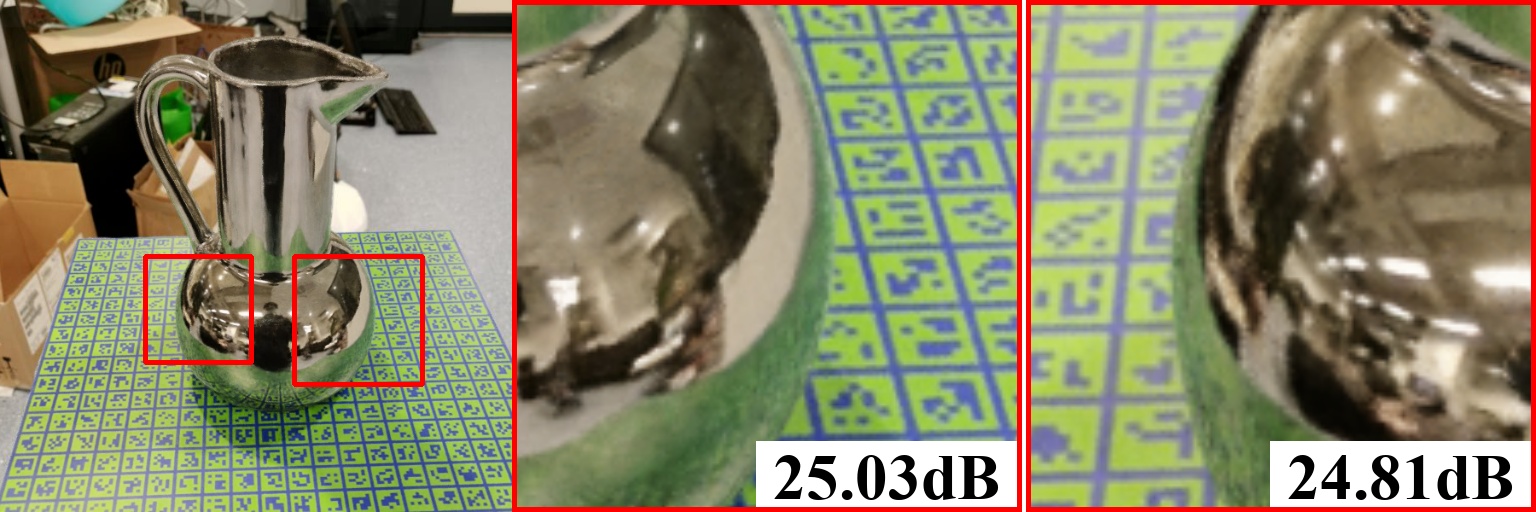}&
        \includegraphics[width=0.34\linewidth]{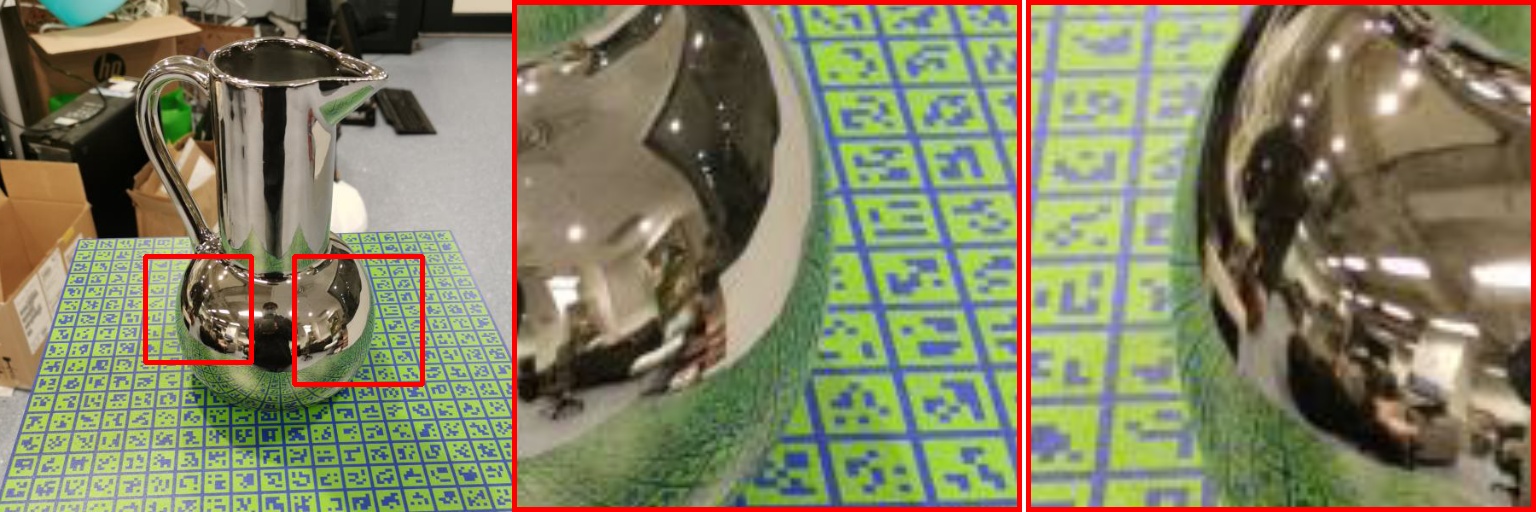}\\

        \includegraphics[width=0.34\linewidth]{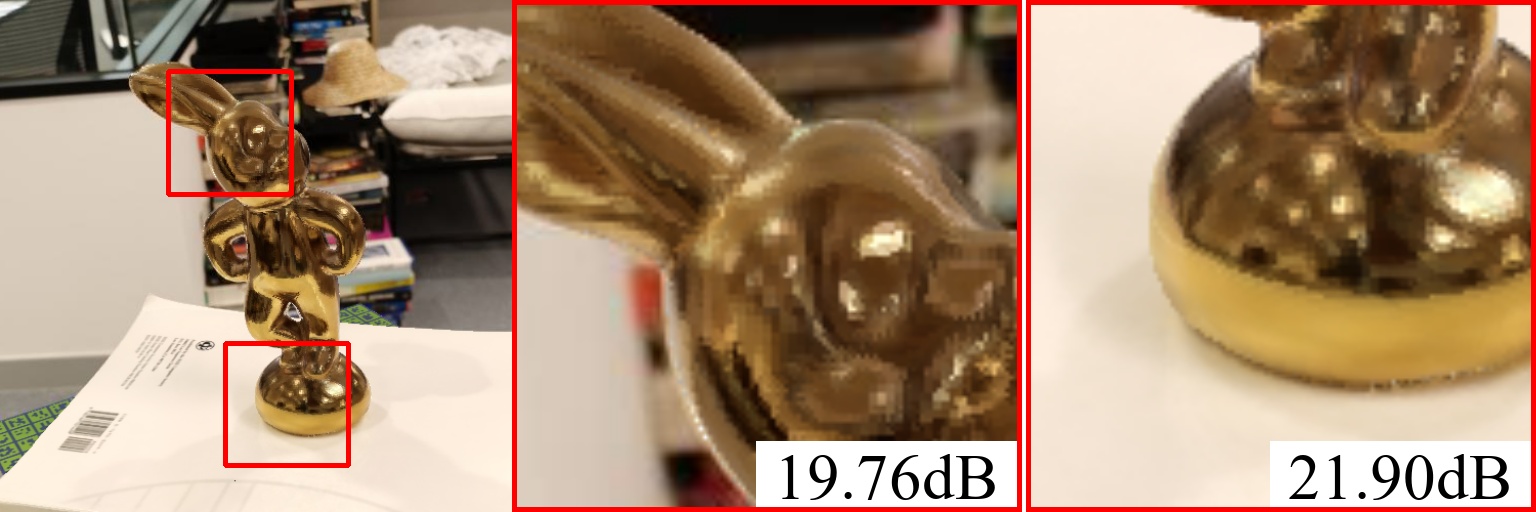}&
        \includegraphics[width=0.34\linewidth]{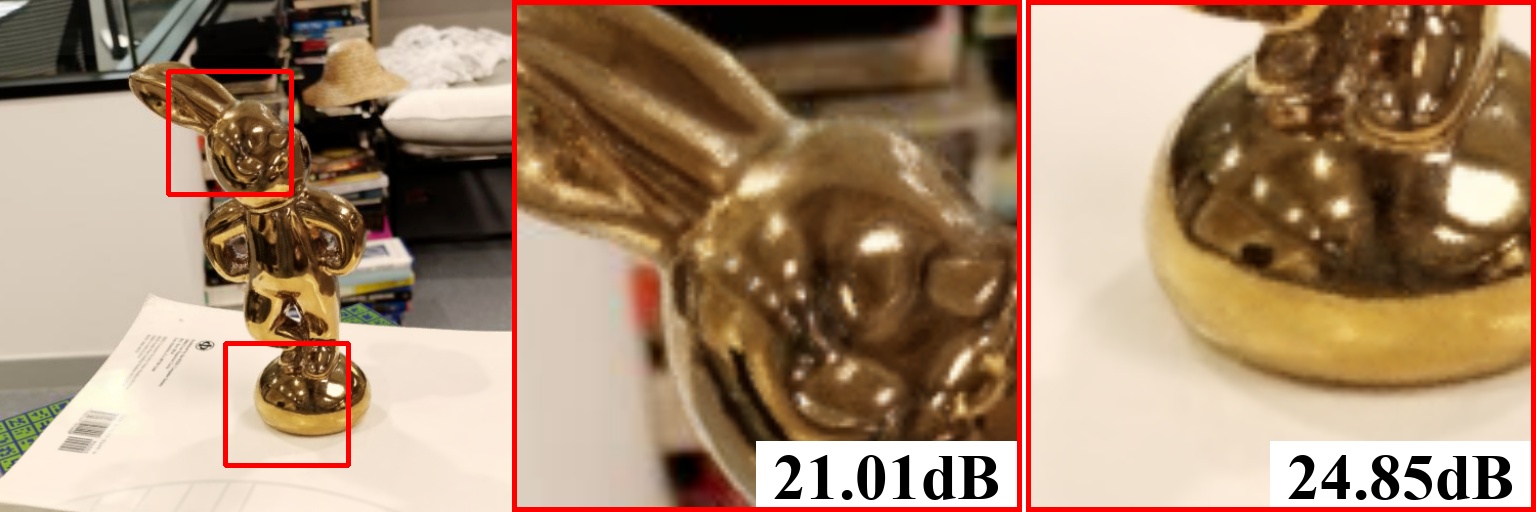}&
        \includegraphics[width=0.34\linewidth]{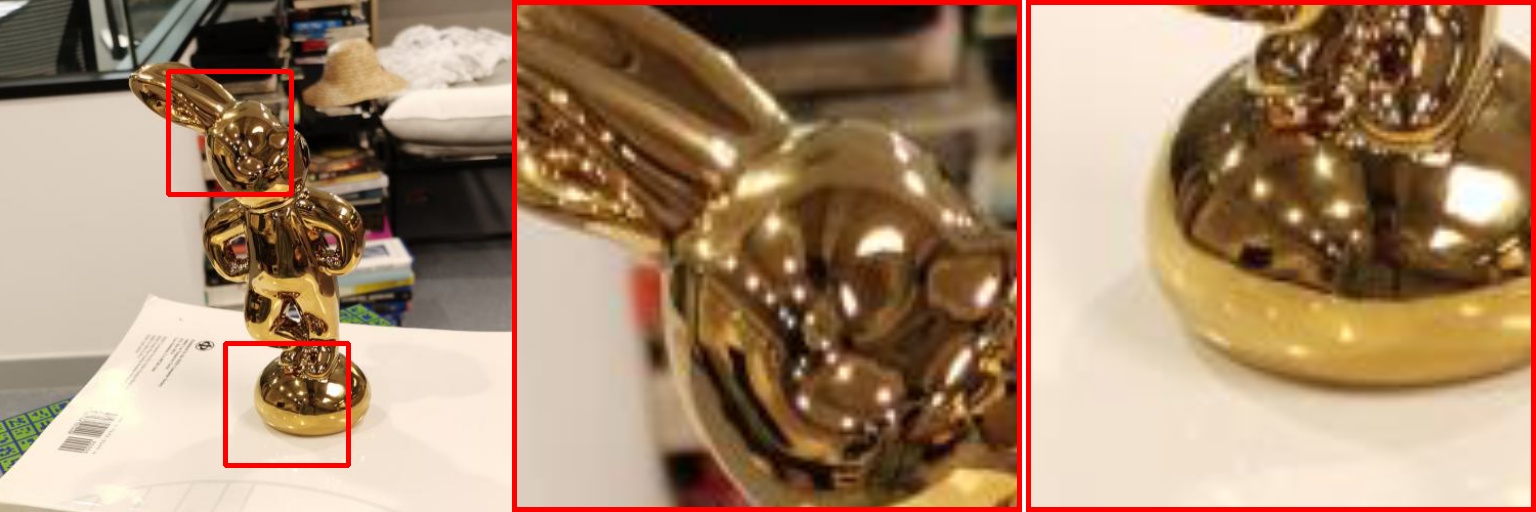}\\
    \end{tabular}
    }
    \caption{
        \textbf{Qualitative comparison on real scenes.} Our NDE gives better reconstruction of the interreflections (the bear's plate and bottom of the vase) and detailed highlights from the environment. 
        Numbers in the insets are image PSNR values.
    }
    \label{fig:real}
\end{figure*}
\paragraph{Results.}
\vspace{-2.5ex}
Overall, our method gives the best rendering quality on synthetic scenes with quantitative results either better or comparable with the baselines (\cref{tab:synthetic}).
This is because our NDE gives the most detailed modeling of both far-field reflections and interreflections,
which also helps improve the geometry reconstruction (\cref{fig:synthetic} bottom).
While ENVIDR's SSIM is slightly better than ours in several scenes,
we not only achieve much better PSNRs (surpassing 2dB), but also higher LPIPS scores.
The PSNR on the Materials (Mat.) scene is worse than Ref-NeRF's because the SDF is inefficient at modeling the concave geometry of the sphere base.
However, our directional MLP is much smaller (\cref{subsec:real-time}),
and we still achieve perceptually better appearance as shown in the insets of \cref{fig:synthetic}.
The qualitative comparison in \cref{fig:real} shows that NDE extends well to real scenes,
producing clearer specular reflections of the complex real-world environments compared to NeRO.

\begin{figure}[t]
    \centering
    \setlength\tabcolsep{0.0pt}
    \resizebox{0.95\linewidth}{!}{
        \begin{tabular}{ccc}
        %\multicolumn{3}{c}{\large\textbf{Reflection separation}}\\[1mm]
        \includegraphics[width=0.3667\linewidth]{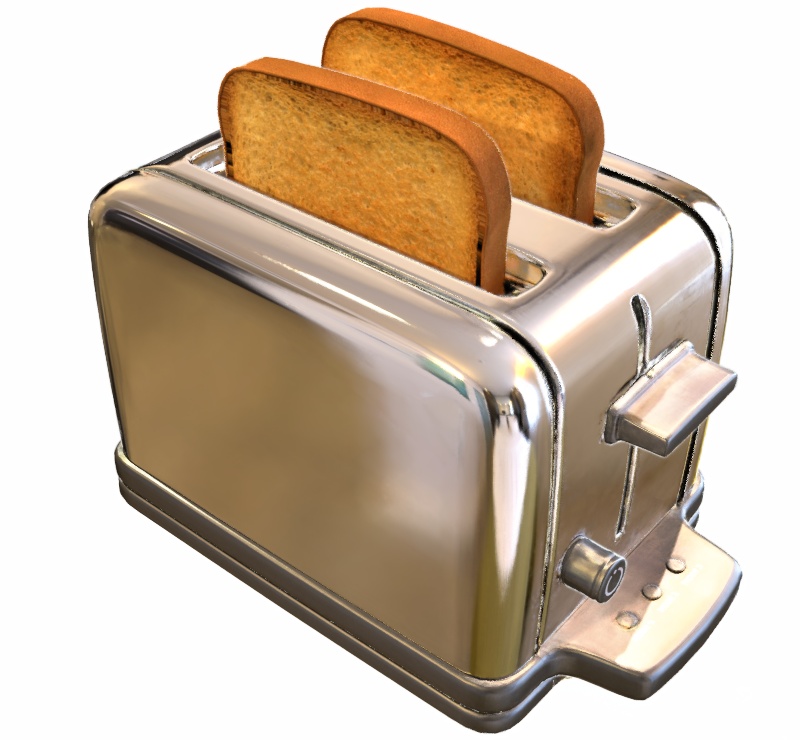}&
        \includegraphics[width=0.3667\linewidth]{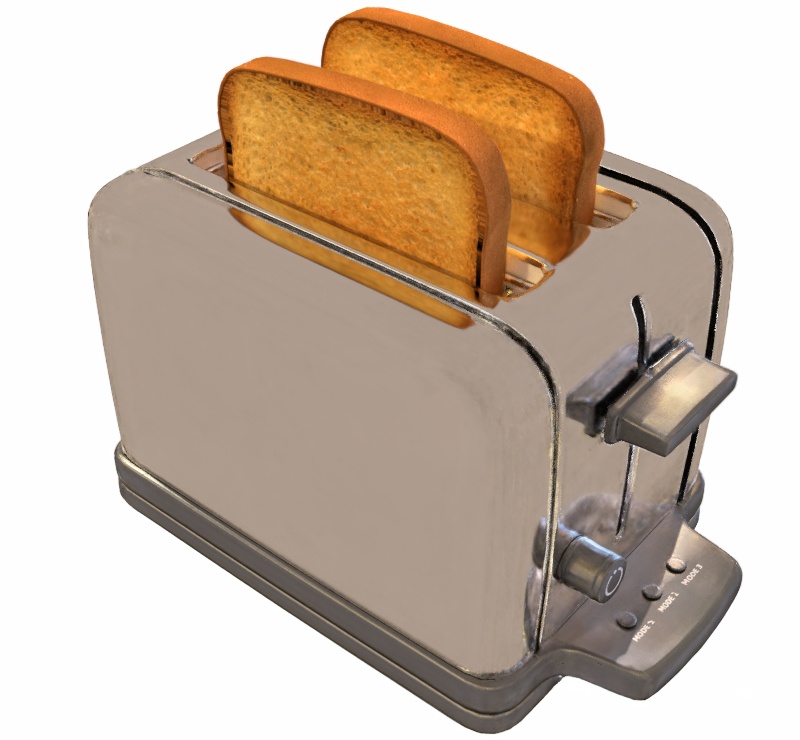}&
        \includegraphics[width=0.3667\linewidth]{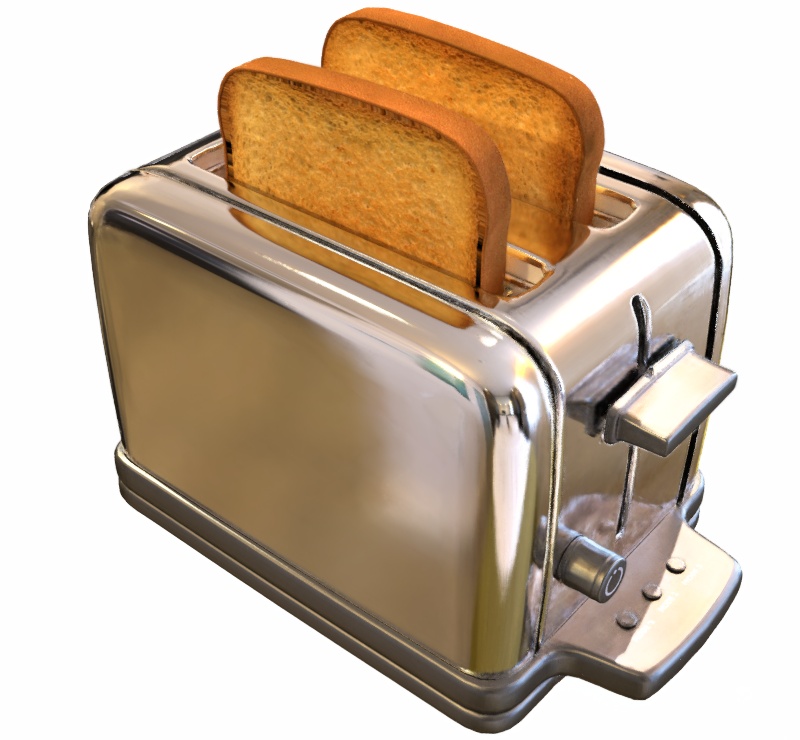}\\
        Far-field reflections & Near-field reflections & Combined\\
        \includegraphics[width=0.3667\linewidth]{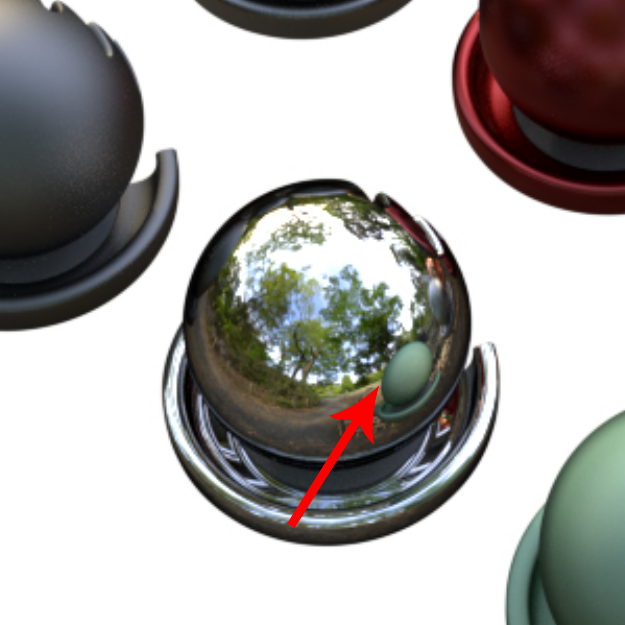}&
        \multicolumn{2}{c}{\includegraphics[width=0.7334\linewidth]{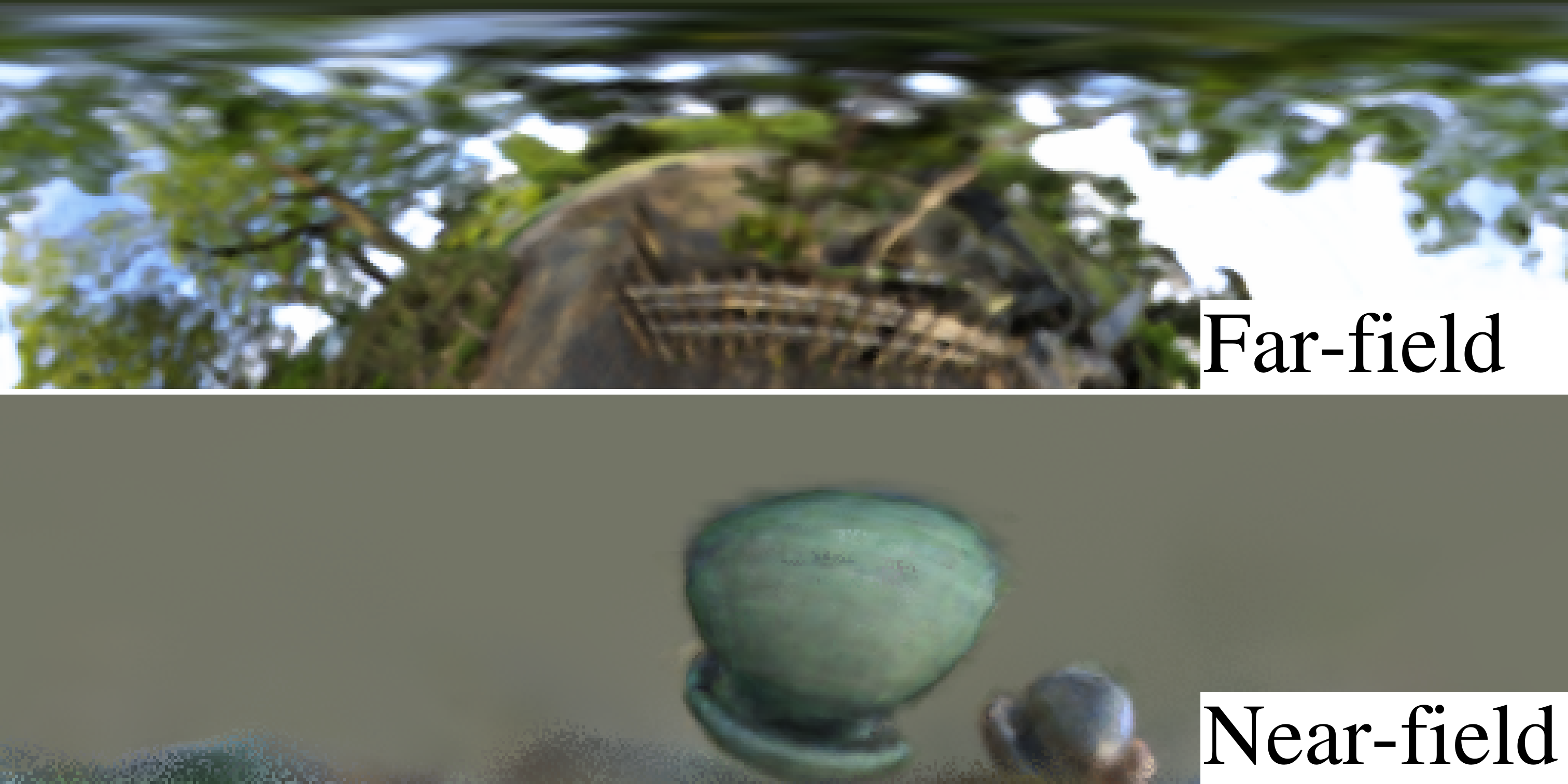}}
        \end{tabular}
    }
    \caption{
        \textbf{Reflection separation.}
        We can visualize different reflection effects by feeding corresponding features into the network.
    }
    \label{fig:reflection-field}
\end{figure}

\begin{figure}[t]
    \centering
    \setlength\tabcolsep{0.0pt}
    \resizebox{0.95\linewidth}{!}{
        \begin{tabular}{c}
             %\textbf{\large Object editing}\\[1mm]
             \includegraphics[width=1.1001\linewidth]{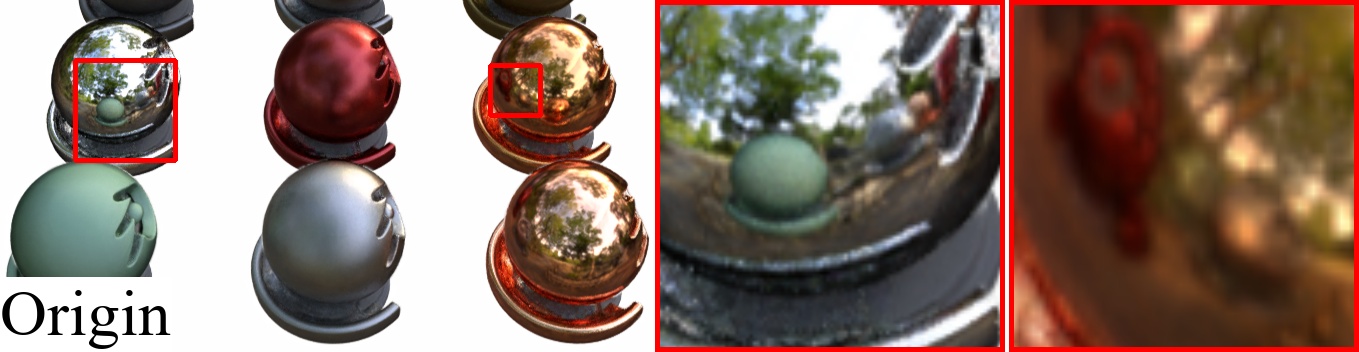}\\[-.5ex]
             \includegraphics[width=1.1001\linewidth]{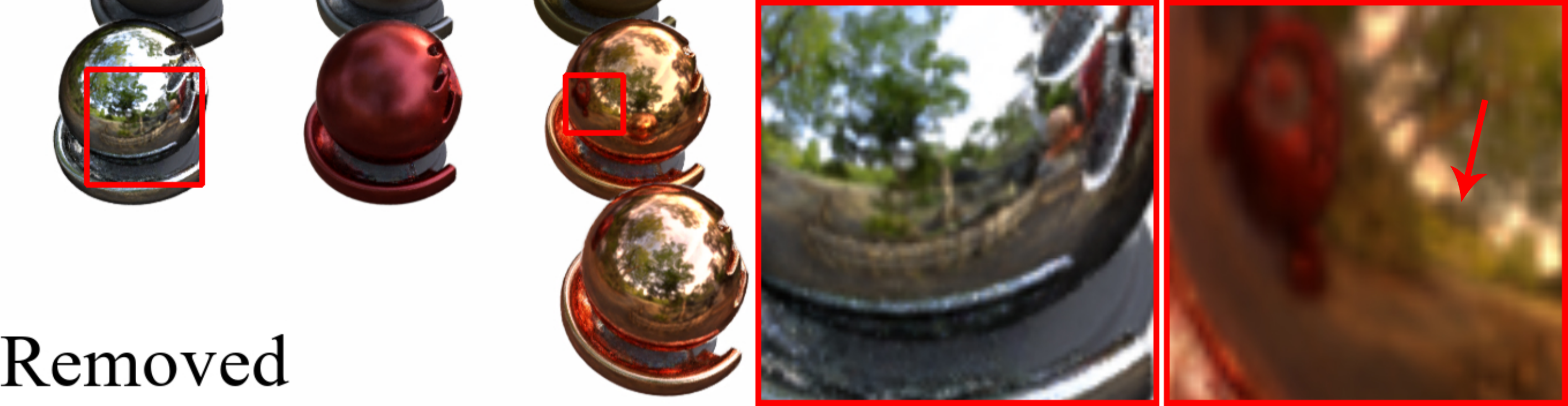}
        \end{tabular}
    }
    \caption{
        \textbf{Editability of our encoding.}
        Reflections from the deleted spheres can be removed by deleting the volume of their indirect features (bottom).
    }
    \label{fig:editability}
\end{figure}

\paragraph{Editability.}
\vspace{-2.5ex}
The near- and far-field features provide a natural separation of different reflections,
allowing us to render these effects separately by excluding $\mathbf{H}_f$ or $\mathbf{H}_n$ during inference (\cref{fig:reflection-field}).
Because interreflections are spatially encoded in the near-field feature grid,
an object and its first-bounce reflections can be removed by masking out both $\sigma$ and $\sigma_n$ from the corresponding regions (\cref{fig:editability}).
This does not work for multi-bounce reflections which are not encoded on the deleted object.

\subsection{Performance comparison}
\label{subsec:real-time}
We compare the evaluation frames per second (FPS) on an $800\!\times\!800$ resolution of the color network and its MLP size (\#Params.) with all baselines in \cref{subsec:view-synthesis} on synthetic scenes.
The color MLPs include the decoder of $\sigma_n,\mathbf{h}_n,\mathbf{c}_s$ for our model (\cref{fig:architectures}),
lighting MLPs for NeRO~\cite{liu2023nero} and ENVIDR~\cite{Liang2023ENVIDRID},
and the directional MLP for Ref-NeRF~\cite{verbin2022ref}.
The spatial-network evaluation is excluded to eliminate the difference caused by different geometry representations, network architectures, and sampling strategies.
For each method, we choose the rendering batch size that maximizes its performance.

\paragraph{Results.}
\vspace{-2.5ex}
As shown in the top half of \cref{tab:inference}, 
our NDE takes a fraction of a second to evaluate, because it requires substantially smaller MLPs to infer color without hurting the rendering.
In contrast, other baselines need large MLPs to maintain rendering quality,
which prevents them to be visualized in real-time.

\begin{figure}
    \centering
    \setlength\tabcolsep{0.6pt}
    \resizebox{0.99\linewidth}{!}{
    \begin{tabular}{cccl}
    \includegraphics[width=0.3667\linewidth]{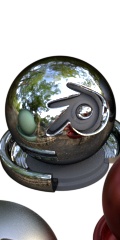}&
    \includegraphics[width=0.3667\linewidth]{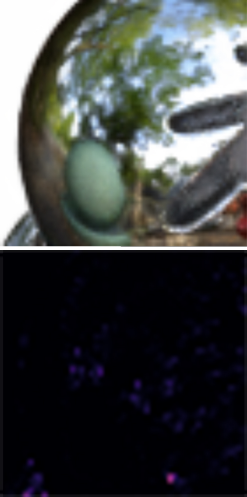}&
    \includegraphics[width=0.3667\linewidth]{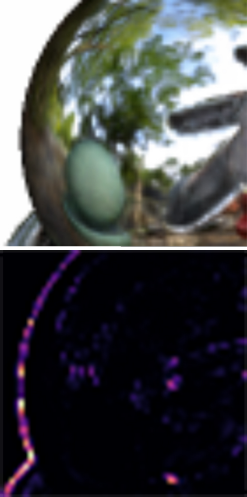}&
    \includegraphics[width=0.04\linewidth]{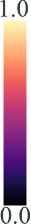}\\
    Ground truth & \textbf{Our offline model} & \textbf{Our real-time model} & 
    \end{tabular}
    }
    \caption{
        \textbf{Error near object boundaries in our real-time model} is caused by the marching-cube extraction of a triangle mesh and its subsequent rasterization (squared error maps at the bottom). This error does not lead to significant qualitative differences (top).
    }
    \label{fig:inference}
\end{figure}

\paragraph{Real-time application.}
\vspace{-2.5ex}
It is possible to create a real-time version of our model by converting the SDF into a mesh through marching cubes~\cite{lorensen1987marching} and baking $\mathbf{c}_d,\mathbf{k}_s,\rho,\mathbf{f}$ into mesh vertices.
The pixel color then can be computed using the rasterized vertex attributes and $\mathbf{c}_s$ decoded from the NDE,
which takes only a single cubemap lookup and cone tracing for each pixel.
As a result, this process requires about the same budget as evaluating a real-time NeRF model~\cite{wu2022diver,sun2022direct,muller2022instant}.
We implement our real-time model (NDE-RT) in WebGL and report the full rendering frame rate (not just color evaluation) at the bottom of \cref{tab:inference} 
%\liwen{
with a real-time baseline 3DGS~\cite{kerbl20233d}.
3DGS is faster as it uses spherical harmonics for color without network evaluation,
which leads to poor specular appearance reconstruction.
%}
Instead, our NDE-RT shows rendering quality comparable to other baselines while achieving frame rates above 60.
The loss in PSNR is mainly due to error around object edges which is cause by the marching-cube mesh extraction and subsequent rasterization (\cref{fig:inference}).
This error does not significantly affect the visual quality and can be resolved by fine-tuning the mesh~\cite{chen2023mobilenerf,Munkberg_2022_CVPR}.

% \begin{figure}
%     \centering
%     \setlength\tabcolsep{1.0pt}
%     \resizebox{0.99\linewidth}{!}{
%     \begin{tabular}{ccc}
%     \includegraphics[width=0.3667\linewidth]{images/inference_gt}&
%     \includegraphics[width=0.3667\linewidth]{images/inference_nde}&
%     \includegraphics[width=0.3667\linewidth]{images/inference_nde(rt)}\\
%     Ground truth & Offline & Real-time
%     \end{tabular}
%     }
%     \caption{
%     \textbf{Aliasing near the object boundary} is the major error source of our real-time model (2nd row). It does not create significant differences in the qualitative results (1st row).
%     }
%     \label{fig:inference}
% \end{figure}

\begin{table}[t]
    \centering
    \setlength\tabcolsep{3 pt}
    \resizebox{1\linewidth}{!}{
    \begin{tabular}{l cc ccc}
    \toprule
    \textbf{Method} & \textbf{FPS}$\uparrow$ & \#\textbf{Params}$\downarrow$ & \textbf{PSNR}$\uparrow$ & \textbf{SSIM}$\uparrow$ & \textbf{LPIPS}$\downarrow$ \\ 
    \midrule
    NeRO     & 0.11 & 454k & 30.61 & 0.936 & 0.109\\
    ENVIDR   & 0.55 & \snd{206k} & 34.95 & \fst{0.982} & 0.035\\
    Ref-NeRF & 0.08 & 521k & \snd{35.88} & 0.969 & 0.053\\
    \textbf{NDE (ours)}      & 3.03 & \fst{75k} & \fst{37.19} & \fst{0.982} & \fst{0.022}\\
    \midrule
    3DGS & \fst{235} & - & 30.30 & 0.949 & 0.076\\
    \textbf{NDE-RT (ours)}   &  \snd{66}  &  \fst{75k} & 35.48 & \snd{0.976} & \snd{0.027}\\
     
    \bottomrule
    \end{tabular}
    }
    \caption{
        \textbf{Performance comparison.} 
        Our NDE achieves high rendering quality, and its use of small MLPs enables fast color evaluation and real-time rendering.
        We report only the evaluation time and parameter counts of color MLPs except for 3DGS (no color MLPs) and our NDE-RT, for which we report the total rendering time.
        All metrics are averaged over the synthetic scenes in \cref{tab:synthetic}. 
    }
    \label{tab:inference}
\end{table}

\subsection{Ablation study}
\label{subsec:ablation-study}

\begin{figure}[t]
    \centering
    \setlength\tabcolsep{1.0pt}
    \resizebox{\linewidth}{!}{
        \begin{tabular}{cc}
           \includegraphics[width=0.55\linewidth]{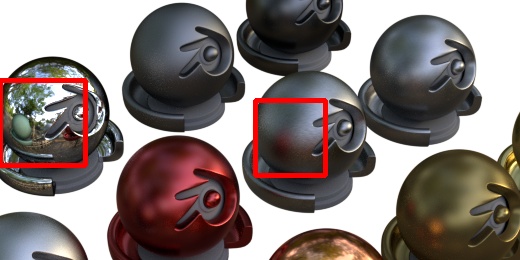}&
            \includegraphics[width=0.55\linewidth]{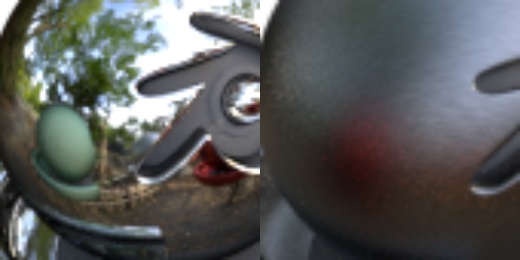}\\[-1mm]
            &Ground truth\\[1.5mm]
            \includegraphics[width=0.55\linewidth]{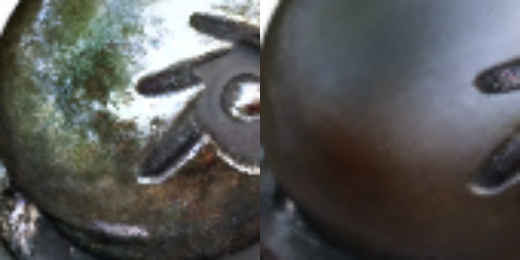}&
            \includegraphics[width=0.55\linewidth]{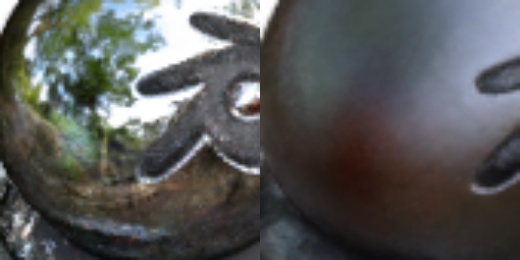}\\[-1mm]
            Analytical $\mathbf{H}_f$ & Cubemap $\mathbf{H}_f$\\[1.5mm]
            \includegraphics[width=0.55\linewidth]{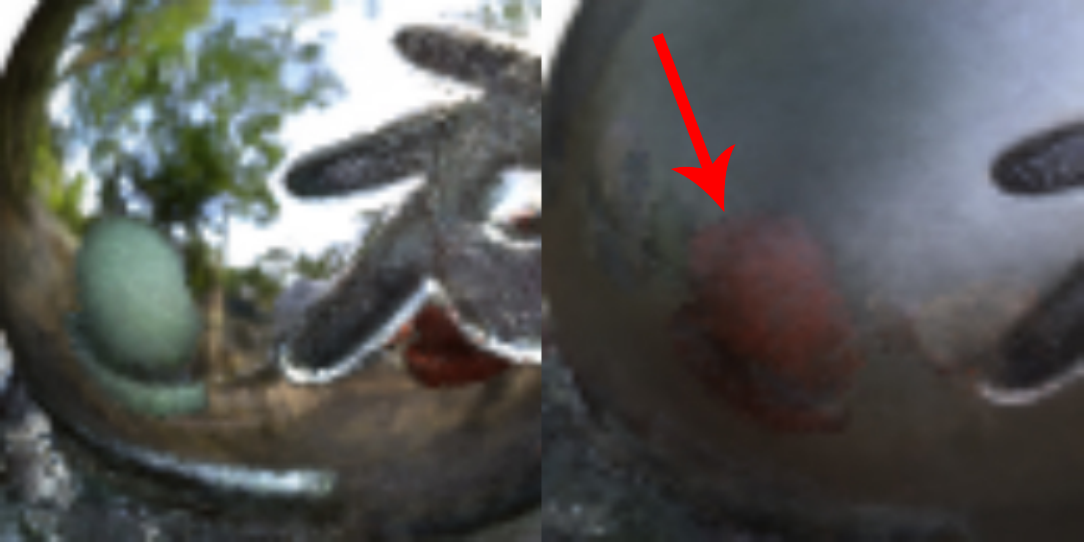}&
            \includegraphics[width=0.55\linewidth]{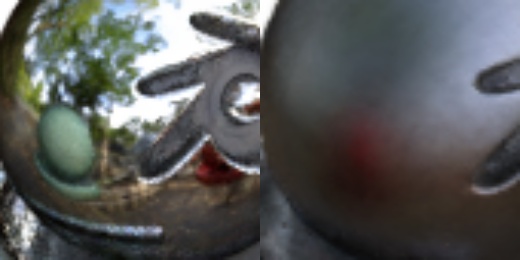}\\[-1mm]
            Volume-rendered $\mathbf{H}_n$ & Cone-traced $\mathbf{H}_n$
        \end{tabular}
    }
    \vspace*{-1mm}
    \caption{
        \textbf{Qualitative ablation of NDE components.}
        Details from the environment light fail to be reconstructed with an analytical encoding (mirror sphere on 2nd row).
        It is also necessary to use the cone-traced near-field feature, otherwise rough surfaces are rendered incorrectly (grey sphere on 3rd row).
    }
    \label{fig:ablation-encoding}
\end{figure}

\paragraph{Different directional encodings.}
%\vspace{-2.5ex}
In \cref{fig:ablation-encoding} we compare different directional encodings on the Materials scene.
IDE~\cite{verbin2022ref} (analytical) with our tiny MLP yields blurry reflections.
Interreflections cannot be reconstructed using only the far-field feature,
and if we volume-render rather than cone-trace the near-field feature, mirror interreflections can be recovered but reflections on rough surfaces look too sharp.
It is therefore necessary to use both the cubemap-based far-field feature and the cone-traced near-field feature to get the best specular appearance (\cref{tab:ablation-encoding}).

\begin{table}[t]
    \centering
    \setlength\tabcolsep{3 pt}
    \resizebox{0.99\linewidth}{!}{
    \begin{tabular}{ll c c c}
    \toprule
    \textbf{Far-field feature} & \textbf{Near-field feature} & \textbf{PSNR}$\uparrow$ & \textbf{SSIM}$\uparrow$ & \textbf{LPIPS}$\downarrow$\\
    \midrule
    Analytical & - & 28.54 & 0.944 & 0.029\\
    Cubemap & - & \snd{30.27} & 0.\snd{962} & \snd{0.022}\\
    Cubemap & Volume-rendered & 29.31 & 0.951 & 0.034\\
    Cubemap & Cone-traced & \fst{31.53} & \fst{0.972} & \fst{0.017}\\
    \bottomrule
    \end{tabular}
    }
    \caption{
    \textbf{Ablation on directional encodings} shows each component of NDE is needed for the best rendering quality. The comparison is made on the Materials scene.
    }
    \label{tab:ablation-encoding}
\end{table}

\paragraph{Network architecture.}
\vspace{-2.5ex}
Table~\ref{tab:ablation-architecture} shows the performance trade-off between different network architectures of our model on synthetic scenes.
Using a smaller MLP width for the decoder of $\sigma_n,\mathbf{h}_n,\mathbf{c}_s$ has only a slight negative impact on the rendering quality but significantly improves real-time performance.
The rendering quality reduction of the real-time model is mainly caused by the error near object edges as discussed in \cref{subsec:real-time}.

\begin{table}[t]
    \centering
    \resizebox{0.99\linewidth}{!}{
    \begin{tabular}{l c  ccc c}
    \toprule
    \textbf{Model}  & \textbf{MLP width} & \textbf{PSNR}$\uparrow$ & \textbf{SSIM}$\uparrow$ & \textbf{LPIPS}$\downarrow$ & \textbf{FPS}$\uparrow$\\
     \midrule
     \multirow{3}{*}{Our offline} & 64 & \fst{37.19} & \fst{0.982} & \fst{0.022} & $<$1\\
      & 32 & \snd{36.69} & \snd{0.979} & \snd{0.026} & $<$1\\
      & 16 & 36.23 & 0.977 & 0.028 & $<$1\\
      \midrule
     \multirow{3}{*}{Our real-time} & 64 & 35.48 & 0.976 & 0.027 & 66\\
      & 32 & 33.97 & 0.971 & 0.034 & \snd{211}\\
      & 16 & 33.71 & 0.969 & 0.036 & \fst{331}\\
     %\midrule
     %\multirow{3}{*}{\begin{tabular}{@{\hskip 0mm}l}
     %     Real-time \\
     %     w/o edge
     %\end{tabular}} 
     % & 64 & 37.46 & 0.979 & 0.022 &\\
     % & 32 & 35.88 & 0.972 & 0.028 &\\
     % & 16 & 35.53 & 0.973 & 0.029 &\\
    \bottomrule
    \end{tabular}
    }
        \caption{\textbf{Ablation on our network architecture.} Using a smaller MLP width introduces a minor loss in rendering fidelity but a noticeable real-time performance boost.
    }
    \label{tab:ablation-architecture}
\end{table}
\begin{table}[t]
    \centering
    \setlength\tabcolsep{2 pt}
    \resizebox{0.99\linewidth}{!}{
    \begin{tabular}{l c c c c c c c c}
    \toprule
         &  \textbf{Mat.} & \textbf{Teapot} & \textbf{Toaster} & \textbf{Car} & \textbf{Ball} & \textbf{Coffee} & \textbf{Helmet} & \textbf{Mean} \\
    \midrule
    \multicolumn{9}{c}{\textbf{PSNR} $\uparrow$}\\
    \midrule

    Hash grid & 30.89 & 49.00 & 29.46 & 30.16 & 43.48 & 34.98 & 37.67 & 36.52\\
    Tri-plane & \fst{31.53} & \fst{49.12} & \fst{30.32} & \fst{30.39} & \fst{44.66} & \fst{36.57} & \fst{37.77} & \fst{37.19}\\
    \midrule
    \multicolumn{9}{c}{\textbf{SSIM} $\uparrow$}\\
    \midrule

    Hash grid & 0.968 & \fst{0.999} & 0.953 & 0.967 & 0.990 & 0.974 & \fst{0.990} & 0.977\\
    Tri-plane & \fst{0.972} & \fst{0.999} & \fst{0.968} & \fst{0.968} & \fst{0.995} & \fst{0.979} & \fst{0.990} & \fst{0.982}\\
    
    \midrule
    \multicolumn{9}{c}{\textbf{LPIPS} $\downarrow$}\\
    \midrule

    Hash grid & 0.019 & \fst{0.002} & 0.058 & 0.025 & 0.031 & 0.043 & \fst{0.014} & 0.027\\
    Tri-plane & \fst{0.017} & \fst{0.002} & \fst{0.039} & \fst{0.024} & \fst{0.022} & \fst{0.033} & \fst{0.014} & \fst{0.022}\\
    
    \bottomrule
    \end{tabular}
    }
    \caption{\textbf{Ablation on mip-mapping strategies} suggests that the mip-mapped tri-plane represents averaged near-field features and density better than the mip-mapped hash grid.
    }
    \label{tab:ablation-mip}
\end{table}
\paragraph{Spatial mip-mapping strategies.}
\vspace{-2.5ex}
Besides mip-mapped tri-plane~\cite{chan2022efficient,hu2023tri},
our architecture can also work with a mip-mapped hash grid~\cite{muller2022instant} for the near-field feature encoding.
Similar to \cite{barron2023zipnerf,li2023neuralangelo}, the hash-grid mip-mapping is implemented by gradually masking out fine-resolution features as the mip level increases.
This results in limited model capacity for rough surfaces where most of the features are masked out,
such that a mip-mapped hash grid produces slightly worse rendering than the tri-plane encoding (\cref{tab:ablation-mip}).

\begin{figure}[t]
    \centering
    \setlength\tabcolsep{1.0pt}
    \resizebox{0.99\linewidth}{!}{
        \begin{tabular}{cccc}
            \includegraphics[width=0.3\linewidth]{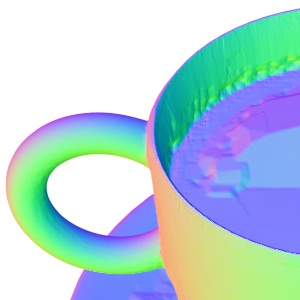}&
            \includegraphics[width=0.3\linewidth]{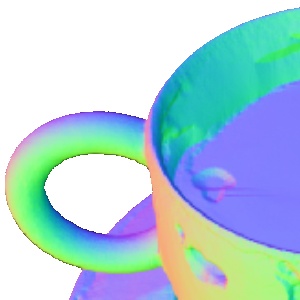}&
            \includegraphics[width=0.3\linewidth]{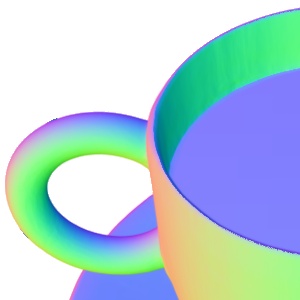}&
            \includegraphics[width=0.3\linewidth]{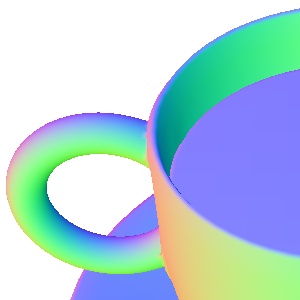}\\
            ENVIDR~\cite{Liang2023ENVIDRID} & \textbf{NDE (hash grid)} & \textbf{NDE (MLP)} & Ground truth
        \end{tabular}
    }
    \caption{
        \textbf{Unstable geometry optimization of specular objects} prevents us from encoding the SDF using a hash grid~\cite{muller2022instant} as it gives incorrect surface normals (middle left).
        This is also the case for other hash-grid-based methods (left).
    }
    \label{fig:limitation}
\end{figure}
\paragraph{Limitations.}
\vspace{-2.5ex}
Like previous works~\cite{verbin2022ref,Liang2023ENVIDRID,liu2023nero},
NDE is sensitive to the quality of the surface normal. 
This prevents us from using more efficient geometry representations such as a hash grid,
which tends to produce corrupted geometry (\cref{fig:limitation}).  
As a result, we use positional-encoded MLPs to model the SDF, which leads to long training times 
%\liwen{
and is difficult for modeling transparent objects. Meanwhile, the editibility of our method is limited.
%}
% As a result, we can only use positional encoded MLPs to model the SDF 
% and our method does not help improve the geometry reconstruction of specular objects.
% As a result, we can only use positional encoded MLPs to model the SDF 
% and normal (\cref{fig:limitation}), which can be inefficient.
\section{Conclusion}
\label{sec:conclusion}
We have adapted feature-based NeRF encodings to the directional domain and introduced a novel spatio-spatial parameterization of view-dependent appearance.
These improvements allow for efficient modeling of complex reflections for novel-view synthesis and could benefit other applications that model spatially varying directional signals, such as neural materials~\cite{kuznetsov2021neumip,zeltner2023real,gflttb2022mipnet} and radiance caching~\cite{muller2021real}.

\vspace{-2.5ex}
\paragraph{Acknowledgements.} 
%This work was funded in part by ONR grant N00014-23-1-2526, NSF grant 2110409, and computational resources from the National Research Platform through NSF grants 2100237 and 2120019. We also gratefully acknowledge gifts from Adobe, Google, Qualcomm, Rembrand, and a Sony Research Award, as well as support from the Ronald L. Graham Chair and the UC San Diego Center for Visual Computing.
This work was supported in part by NSF grants 2110409, 2100237, 2120019, ONR grant N00014-23-1-2526, gifts from Adobe, Google, Qualcomm, Rembrand, a Sony Research Award, as well as the Ronald L. Graham Chair and the UC San Diego Center for Visual Computing.
Additionally, we thank Jingshen Zhu for insightful discussions.

\small
\bibliographystyle{ieeenat_fullname}
\bibliography{main}

% WARNING: do not forget to delete the supplementary pages from your submission 
\begin{appendix}
\section{Additional Implementation Details}

\subsection{Cone tracing footprint}
In Sec.~4.2, we choose the cone to cover the (cosine weighted) GGX distribution~\cite{walter2007microfacet} centered in the reflected direction $\bm{\omega}_r$.
Assuming $\bm{\omega}_r\!=\!(0,0,1)$,
the distribution $D$ with roughness $\rho$ in spherical coordinates $(\theta,\phi)$ can be written as:
\begin{equation}
    D(\theta,\phi)=\frac{\alpha^2\max(\cos\theta,0)}{\pi(\cos^2\theta(\alpha^2-1)+1)^2},
    \ \alpha = \rho^2.
\end{equation}
If we want the cone to cover a certain fraction $T$ of the distribution,
the polar angle $\theta$ should satisfy:
\begin{equation}
\begin{split}
    T&=\int_0^{2\pi}\!\int_0^\theta
    D(\theta',\phi)\sin\theta' d\theta' d\phi\\
    &=\frac{1-\cos^2\theta}{1+\cos^2\theta(\alpha^2-1)}\\
    \Rightarrow &\cos\theta=\sqrt{\frac{1-T}{T(\alpha^2-1)+1}},
\end{split}
\end{equation}
which gives the base cone radius $r_0$:
\begin{equation}
    r_0=\cot\theta=\frac{\sqrt{1-\cos^2\theta}}{\cos\theta}=\sqrt{\frac{T}{1-T}}\rho^2.
\end{equation}
We found $T\!=\!75\%$ in practice gives good results, which suggests $r_0=\sqrt{3}\rho^2$.
Therefore, the footprint at $\mathbf{x}'_i$ from $\mathbf{x}$ is $r_i\!=\!\sqrt{3}\rho^2\|\mathbf{x}-\mathbf{x}^\prime_i\|_2$.

\subsection{Real-time application}
We use a two-pass deferred shading in our real-time model.
The first pass rasterizes the world-space position $\mathbf{x}$, normal $\mathbf{n}$, diffuse color $\mathbf{c}_d$, specular tint $\mathbf{k}_s$, spatial feature $\mathbf{f}$, and roughness $\rho$ into the G-buffer.
In the second pass,
we then calculate the NDE $\mathbf{H}$, including a cubemap lookup for far-field feature $\mathbf{H}_f$ and the cone tracing of near-field feature $\mathbf{H}_n$,
and decode it to get the specular color $\mathbf{c}_s$.
The MLP evaluations are executed sequentially inside the pixel shader,
and we implement the early ray termination trick~\cite{yu2021plenoctrees,hedman2021baking} to stop the cone tracing if the accumulated transmittance is below 0.01.
Because small decoder MLPs tend to provide unstable geometry optimization,
we use the fixed SDF network weight from our NDE trained with 64 MLP width when training other variants that use smaller decoder MLPs (Sec.~5.3).

\subsection{Spatial mip-mapping strategies}
We introduce mip-mapping strategies of spatial encodings in Sec.~5.3 using either a triplane~\cite{chan2022efficient,hu2023tri} or a hash grid~\cite{muller2022instant}.
Let $\mathbf{T}_{xy},\mathbf{T}_{yz},\mathbf{T}_{zx}$ denote the three 2D planes of the triplane $\mathbf{T}$.
A mip-mapped query at location $\mathbf{x}\!=\!(x,y,z)$ of mip level $\lambda$ is given by:
\begin{equation}
\begin{gathered}
\begin{aligned}
        \text{mipmap}&(\mathbf{T}(\mathbf{x}),\lambda)=\\
        &\bigoplus_{\mathbf{u}\in U}\text{lerp}(\mathbf{T}^{\lfloor\lambda\rfloor}_\mathbf{u}(\mathbf{u}),\mathbf{T}^{\lceil\lambda\rceil}_\mathbf{u}(\mathbf{u}),\lambda\!-\!\lfloor\lambda\rfloor),
\end{aligned}\\
U\!=\!\{(x,y),\!(y,z),\!(z,x)\}, \mathbf{T}^k_\mathbf{u}\!=\!\text{downsample}(\mathbf{T}_\mathbf{u},k),
\end{gathered}
\end{equation}
where $\bigoplus$ is the concatenation operation.
For a hash grid feature $\mathbf{F}$ with $l^\text{th}$ level feature $\mathbf{F}_l$ (beginning from the finest resolution), its mip-mapping is given by:
\begin{equation}
    \text{mipmap}(\mathbf{F}(\mathbf{x}),\lambda)=\bigoplus_{l}
    \text{clamp}(l+1-\lambda,0,1)\mathbf{F}_l(\mathbf{x}).
\end{equation}

\begin{table}[t]
    \centering
    \setlength\tabcolsep{4 pt}
    \resizebox{0.7\linewidth}{!}{
        \begin{tabular}{l c c c}
        \toprule
        &ENVIDR & Ref-NeRF & \textbf{NDE (ours)}\\
        \midrule
        \textbf{PSNR}$\uparrow$&
        22.67 & \snd{23.46} & \fst{23.63}\\
        \bottomrule
        \end{tabular}
    }
    \caption{
        \textbf{PSNR on the Ref-NeRF Garden Spheres scene.}
    }
    \label{tab:ref-real}
\end{table}
\begin{figure}[t]
    \centering
    \setlength\tabcolsep{1.0pt}
    \resizebox{1.0\linewidth}{!}{
    \begin{tabular}{ccc}
        ENVIDR & \textbf{NDE (ours)} &Ground truth\\
        \includegraphics[width=0.34\linewidth]{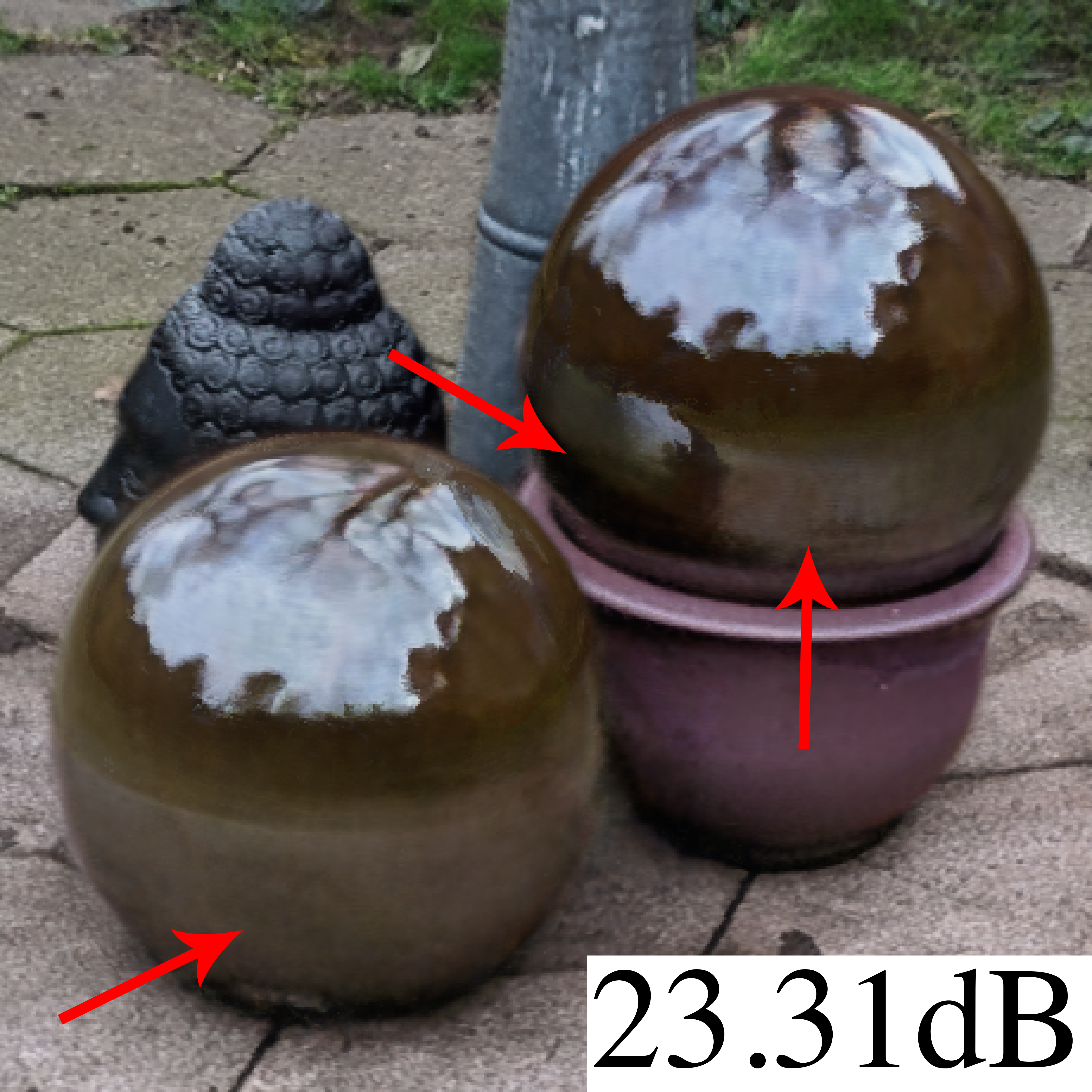}&
        \includegraphics[width=0.34\linewidth]{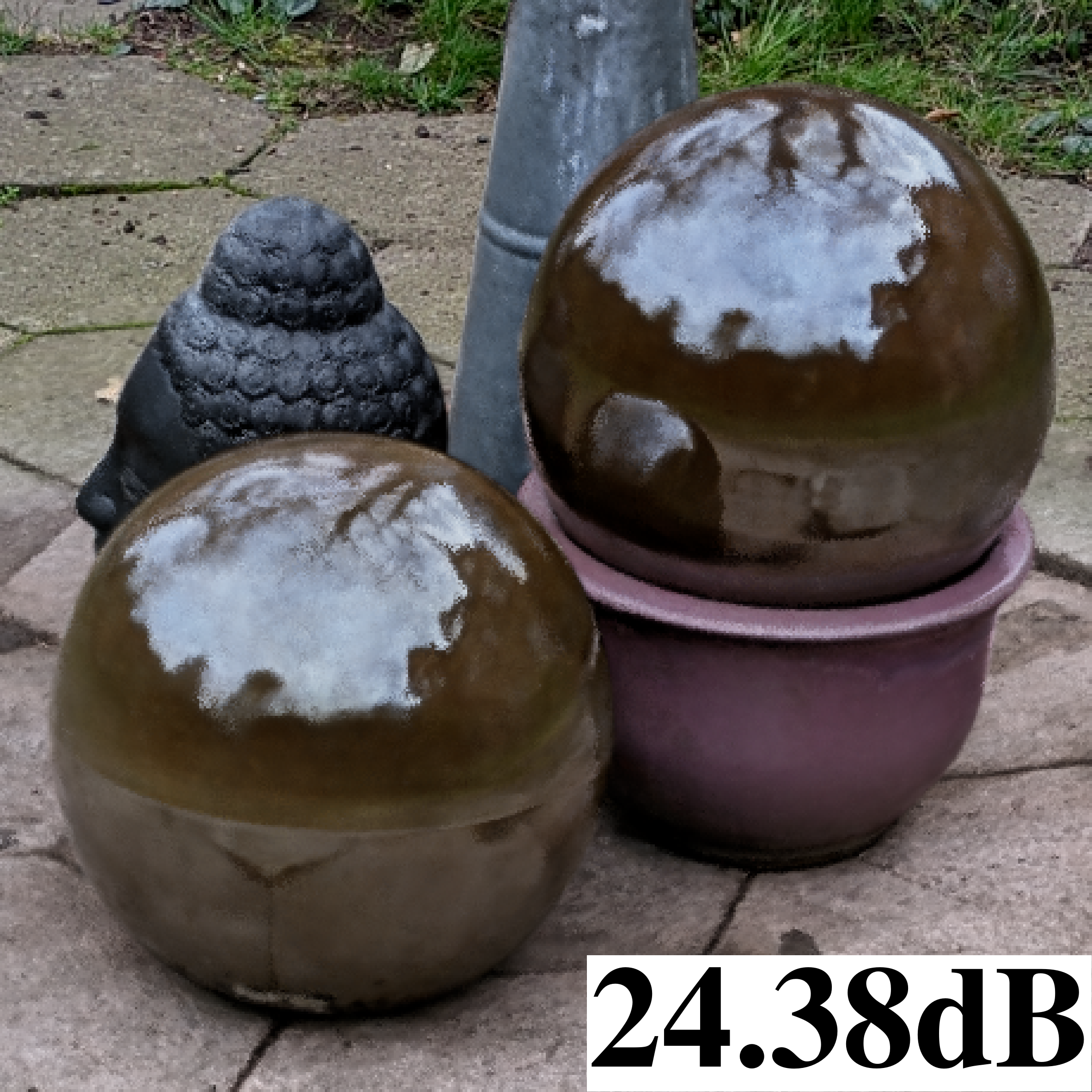}&
        \includegraphics[width=0.34\linewidth]{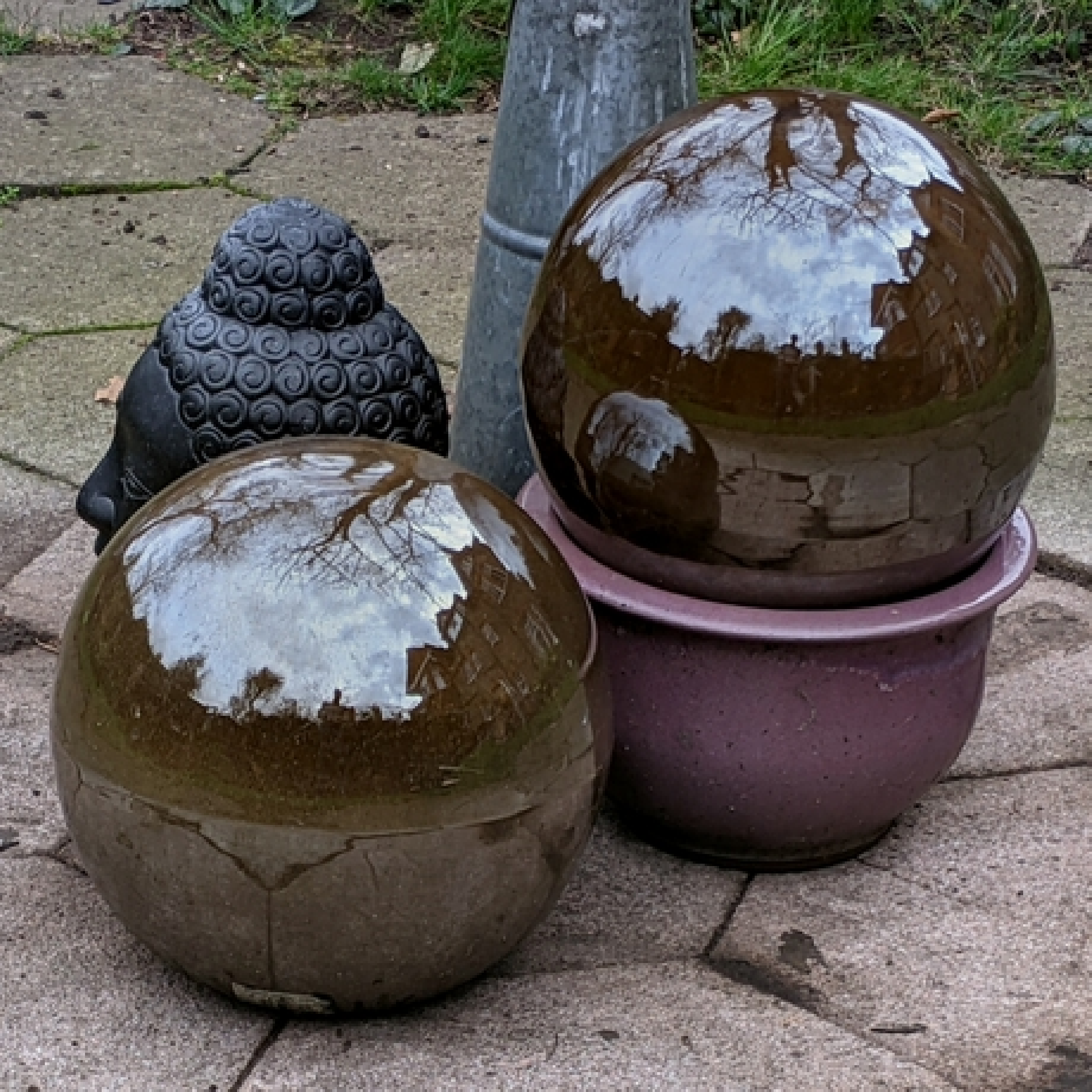}\\
    \end{tabular}
    }
    \caption{
        \textbf{Qualitative comparison on the Garden Spheres scene of Ref-NeRF real dataset.} Numbers shows the image PSNR; zoom in to see the difference.
    }
    \label{fig:ref-real}
\end{figure}
\section{Additional Results}
For the unbounded real scene evaluation,
we provide the results on the Garden Spheres scene of Ref-NeRF real dataset~\cite{verbin2022ref} in \cref{tab:ref-real} and \cref{fig:ref-real}. 
It can be seen that our method is able to recover more interreflection details in real-world compared to other baselines.
Considering perceptual measures are more reasonable for reflection quality comparison, we additionally show the FLIP~\cite{andersson2020flip} metric on synthetic scenes in \cref{tab:rebuttal-flip}.
Overall, our method still demonstrates higher rendering quality compared to other baselines.

\begin{table}[t]
    \centering
    \setlength\tabcolsep{2 pt}
    \resizebox{0.99\linewidth}{!}{
        \begin{tabular}{l c c c c c c c c}
        \toprule
        \textbf{Method} & \textbf{Mat.} & \textbf{Teapot} & \textbf{Toaster} & \textbf{Car} & \textbf{Ball} & \textbf{Coffee} & \textbf{Helmet} & \textbf{Mean} \\
        \midrule
        NeRO & 0.082 & 0.012 & \snd{0.097} & \snd{0.049} & 0.058 & 0.039 & 0.083 & 0.060 \\
        ENVIDR & 0.062 & 0.030 & 0.098 & 0.056 & \snd{0.037} & 0.046 & \snd{0.049} & 0.054 \\
        Ref-NeRF & \fst{0.023} & \snd{0.011} & 0.108 & 0.071 & 0.038 & \fst{0.030} & 0.072 & \snd{0.050} \\
        3DGS & 0.042 & \fst{0.007} & 0.153 & 0.051 & 0.104 & \snd{0.032} & 0.073 & 0.066\\
        \textbf{NDE} & \snd{0.039} & \fst{0.007} & \fst{0.065} & \fst{0.038} & \fst{0.027} & 0.035 & \fst{0.035} & \fst{0.035}\\
        \bottomrule
        \end{tabular}
    }
    \caption{
        \textbf{FLIP metric on synthetic scenes.}
    }
    \label{tab:rebuttal-flip}
\end{table}

\begin{table}[t]
    \centering
    \resizebox{0.99\linewidth}{!}{
    \begin{tabular}{r c c c c c}
    \toprule
    & NeRO & ENVIDR & Ref-NeRF & \textbf{NDE} & \textbf{NDE-RT}\\
    \midrule
    \textbf{PSNR} $\uparrow$ & 28.75 & 31.29 & 28.18 & \fst{34.08} & \snd{32.97}\\
    \textbf{SSIM} $\uparrow$ & 0.956 & 0.969 & 0.946 & \fst{0.985} & \snd{0.984}\\
    \textbf{LPIPS} $\downarrow$& 0.046 & 0.022 & 0.030 & \fst{0.008} & \snd{0.010}\\
\bottomrule
    \end{tabular}
    }
        \caption{\textbf{Quantitative results on the teaser scene.}
    }
    \label{tab:supp-teaser}
\end{table}
\begin{table}[t]
    \centering
    \setlength\tabcolsep{2 pt}
    \resizebox{0.99\linewidth}{!}{
        \begin{tabular}{l c c c c c c c c}
        \toprule
         & \textbf{Mat.} & \textbf{Teapot} & \textbf{Toaster} & \textbf{Car} & \textbf{Ball} & \textbf{Coffee} & \textbf{Helmet} & \textbf{Mean} \\
        \midrule
        \multicolumn{9}{c}{\textbf{PSNR} $\uparrow$}\\\midrule
        3DGS & 29.98 & 45.69 & 20.99 & 27.25 & 27.65 & 32.31 & 28.26 & 30.30\\
        \textbf{NDE-RT} &\fst{30.28} & \fst{47.02} & \fst{28.31} & \fst{28.91} & \fst{43.23} & \fst{34.21} & \fst{36.38} & \fst{35.48}\\
        \midrule
        \multicolumn{9}{c}{\textbf{SSIM} $\uparrow$}\\\midrule
        3DGS & 0.960 & 0.997 & 0.895 & 0.930 & 0.937 & \fst{0.972} & 0.951 & 0.949\\
        \textbf{NDE-RT} & \fst{0.967} & \fst{0.998} & \fst{0.954} & \fst{0.962} & \fst{0.994} & \fst{0.972} & \fst{0.987} & \fst{0.976}\\
        \midrule
        \multicolumn{9}{c}{\textbf{LPIPS} $\downarrow$}\\\midrule
        3DGS & 0.034 & 0.007 & 0.126 & 0.048 & 0.162 & 0.078 & 0.080 & 0.076\\
        \textbf{NDE-RT} & \fst{0.020} & \fst{0.003} & \fst{0.051} & \fst{0.030} & \fst{0.023} & \fst{0.041} & \fst{0.019} & \fst{0.027}\\
        %\textbf{FLIP}$\downarrow$ & 0.042 & 0.007 & 0.153 & 0.051 & 0.104 & 0.032 & 0.073 & 0.066\\
        \bottomrule
        \end{tabular}
    }
    \caption{
        \textbf{Per-scene comparison with 3DGS on synthetic scenes.}
    }
    \label{tab:3dgs}
\end{table}
\begin{table}[t]
    \centering
    \setlength\tabcolsep{2 pt}
    \resizebox{1\linewidth}{!}{
        \begin{tabular}{l c c c c c c c c}
        \toprule
        \textbf{Method} & \textbf{Mat.} & \textbf{Teapot} & \textbf{Toaster} & \textbf{Car} & \textbf{Ball} & \textbf{Coffee} & \textbf{Helmet} & \textbf{Mean} \\
        \midrule
        Hashgrid & 30.12 & 46.46 & 25.83 & 29.94 & 36.41 & 33.25 & 34.08 & 33.73\\
        Pos. enc. & \fst{31.53} & \fst{49.12} & \fst{30.32} & \fst{30.39} & \fst{44.66} & \fst{36.57} & \fst{37.77} & \fst{37.19}\\
        \bottomrule
        \end{tabular}
    }
    \caption{
        \textbf{Comparison of geometry encoding on synthetic scenes in PSNR.} ``Pos. enc.'' denotes positional encoding.
    }
    \label{tab:rebuttal-hashgrid}
\end{table}

\section{Experiment Details}
We provide the quantitative results on the teaser scene (Fig.~1 of the main paper) compared to the baselines in \cref{tab:supp-teaser} and the per-scene comparison of our real-time model (NDE-RT) with 3DGS~\cite{kerbl20233d} in \cref{tab:3dgs}.
Table~\ref{tab:rebuttal-hashgrid} shows the comparison of different SDF encodings (Fig.~12 of the main paper). 
Table~\ref{tab:supp-offline} and \ref{tab:supp-real-time} show the per-scene quantitative results of our real-time and offline model with different MLP width (Width) on the synthetic dataset.
In \cref{fig:supp}, we show the per-scene rendering results of both our offline (NDE) and real-time (NDE-RT) model on the synthetic dataset together with the reconstructed surface normals.
The normals are masked by the foreground mask to get rid of floaters with the background color.

\begin{table}[t]
    \centering
    \setlength\tabcolsep{2 pt}
    \resizebox{0.99\linewidth}{!}{
    \begin{tabular}{l c c c c c c c c}
    \toprule
    \textbf{Width} & \textbf{Mat.} & \textbf{Teapot} & \textbf{Toaster} & \textbf{Car} & \textbf{Ball} & \textbf{Coffee} & \textbf{Helmet} & \textbf{Mean} \\\midrule
    \multicolumn{9}{c}{\textbf{PSNR} $\uparrow$}\\\midrule
64  & 31.53 & 49.12 & 30.32 & 30.39 & 44.66 & 36.57 & 37.77 & 37.19 \\
32  & 30.89 & 48.88 & 29.33 & 29.51 & 44.34 & 36.24 & 37.63 & 36.69 \\
16  & 30.59 & 48.56 & 29.09 & 29.24 & 43.61 & 36.07 & 36.47 & 36.23 \\
\midrule

\multicolumn{9}{c}{\textbf{SSIM} $\uparrow$}\\\midrule
64  & 0.972 & 0.999 & 0.968 & 0.968 & 0.995 & 0.979 & 0.990 & 0.982 \\
32  & 0.968 & 0.999 & 0.961 & 0.962 & 0.994 & 0.977 & 0.989 & 0.979 \\
16  & 0.965 & 0.998 & 0.959 & 0.960 & 0.994 & 0.977 & 0.986 & 0.977 \\
\midrule

\multicolumn{9}{c}{\textbf{LPIPS} $\downarrow$}\\\midrule
64  & 0.017 & 0.002 & 0.039 & 0.024 & 0.022 & 0.033 & 0.014 & 0.022 \\
32  & 0.021 & 0.002 & 0.057 & 0.032 & 0.021 & 0.033 & 0.017 & 0.026 \\
16  & 0.023 & 0.002 & 0.058 & 0.034 & 0.021 & 0.034 & 0.022 & 0.028 \\
    \bottomrule
    \end{tabular}
    }
        \caption{\textbf{Per-scene results of our offline models on synthetic scenes.}
        The first column suggests the decoder MLP width.
    }
    \label{tab:supp-offline}
\end{table}
\begin{table}[t]
    \centering
    \setlength\tabcolsep{2 pt}
    \resizebox{0.99\linewidth}{!}{
    \begin{tabular}{l c c c c c c c c}
    \toprule
    \textbf{Width} & \textbf{Mat.} & \textbf{Teapot} & \textbf{Toaster} & \textbf{Car} & \textbf{Ball} & \textbf{Coffee} & \textbf{Helmet} & \textbf{Mean} \\\midrule
    \multicolumn{9}{c}{\textbf{PSNR} $\uparrow$}\\\midrule
64 & 30.28 & 47.02 & 28.31 & 28.91 & 43.23 & 34.21 & 36.38 & 35.48\\
32 & 28.84 & 43.75 & 27.28 & 27.76 & 41.00 & 33.92 & 35.26 & 33.97\\
16 & 28.33 & 43.14 & 26.95 & 27.47 & 41.48 & 33.81 & 34.81 & 33.71\\
\midrule

\multicolumn{9}{c}{\textbf{SSIM} $\uparrow$}\\\midrule
64 & 0.967 & 0.998 & 0.954 & 0.962 & 0.994 & 0.972 & 0.987 & 0.976\\
32 & 0.955 & 0.996 & 0.945 & 0.954 & 0.992 & 0.971 & 0.984 & 0.971\\
16 & 0.951 & 0.996 & 0.942 & 0.951 & 0.992 & 0.971 & 0.982 & 0.969\\
\midrule

\multicolumn{9}{c}{\textbf{LPIPS} $\downarrow$}\\\midrule
64 & 0.020 & 0.003 & 0.051 & 0.030 & 0.023 & 0.041 & 0.019 & 0.027\\
32 & 0.030 & 0.006 & 0.068 & 0.039 & 0.025 & 0.043 & 0.026 & 0.034\\
16 & 0.033 & 0.007 & 0.070 & 0.043 & 0.024 & 0.044 & 0.030 & 0.036\\
\midrule
\multicolumn{9}{c}{\textbf{FPS} $\uparrow$}\\\midrule
64 & 55 & 130 & 55 & 70 & 65 & 30 & 58 & 66\\
32 & 220 & 400 & 150 & 200 & 210 & 110 & 190 & 211\\
16 & 350 & 600 & 250 & 330 & 300 & 200 & 290 & 331\\
\bottomrule
    \end{tabular}
    }
        \caption{\textbf{Per-scene results of our real-time models on synthetic scenes.} The first column suggests the deocder MLP width.
    }
    \label{tab:supp-real-time}
\end{table}
\end{appendix}
\begin{figure*}[t]
    \centering
    \setlength\tabcolsep{0.5pt}
    \resizebox{0.97\linewidth}{!}{
    \begin{tabular}{ccc|cc}
NDE & NDE-RT & Ground truth & Estimated normal & Ground truth normal\\[-1mm]
\includegraphics[trim={0 50 0 50},clip,width=0.21\linewidth]{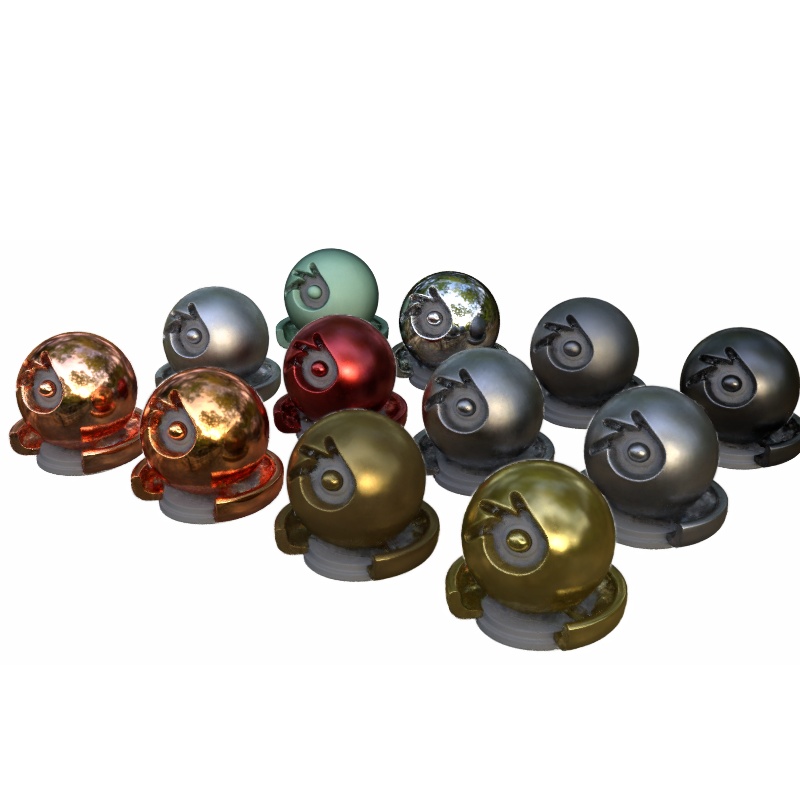} &
\includegraphics[trim={0 50 0 50},clip,width=0.21\linewidth]{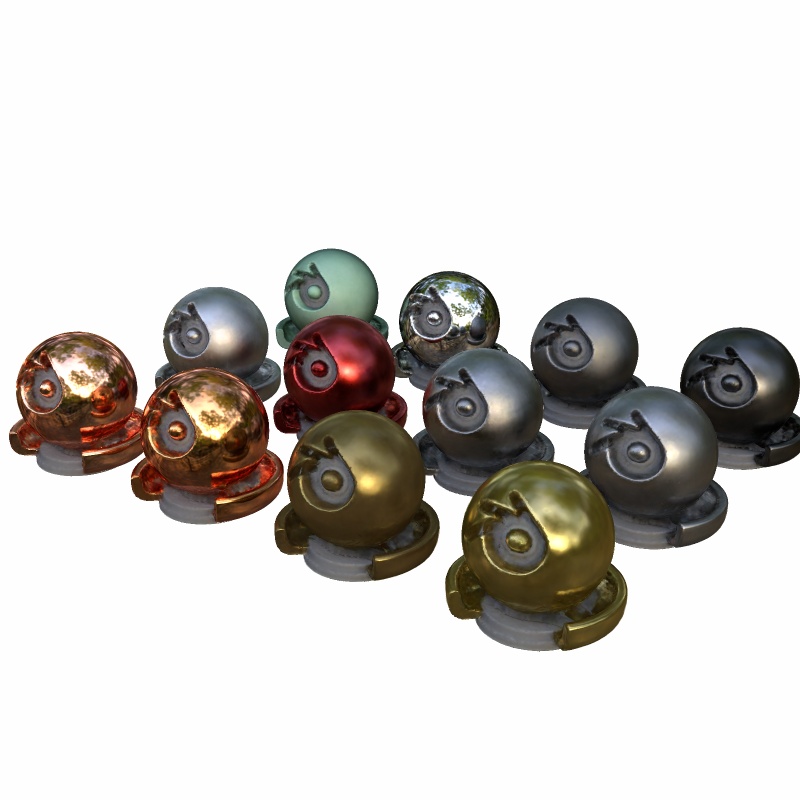} &
\includegraphics[trim={0 50 0 50},clip,width=0.21\linewidth]{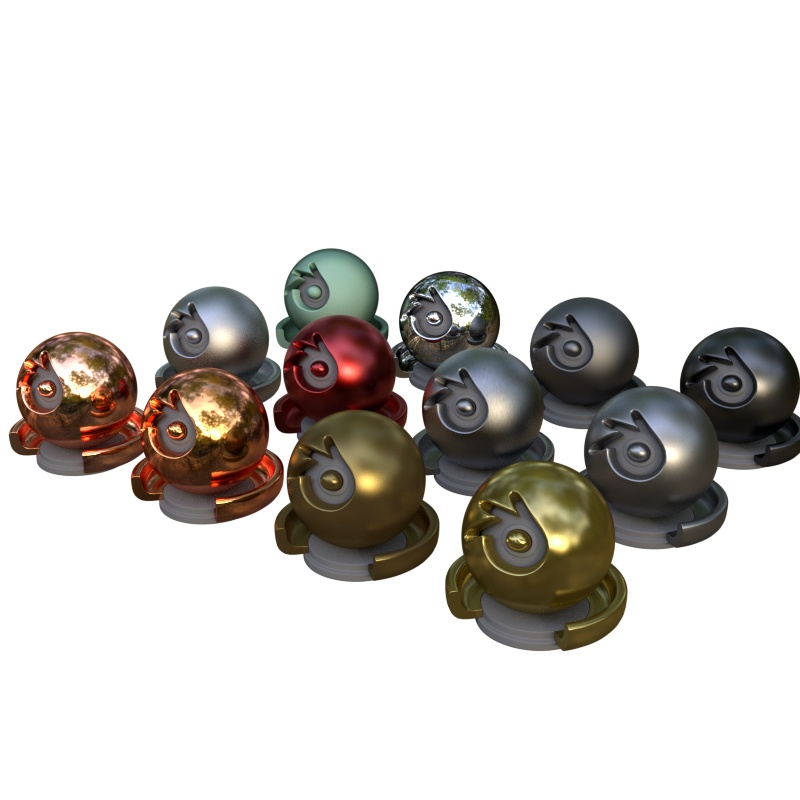} &
\includegraphics[trim={0 50 0 50},clip,width=0.21\linewidth]{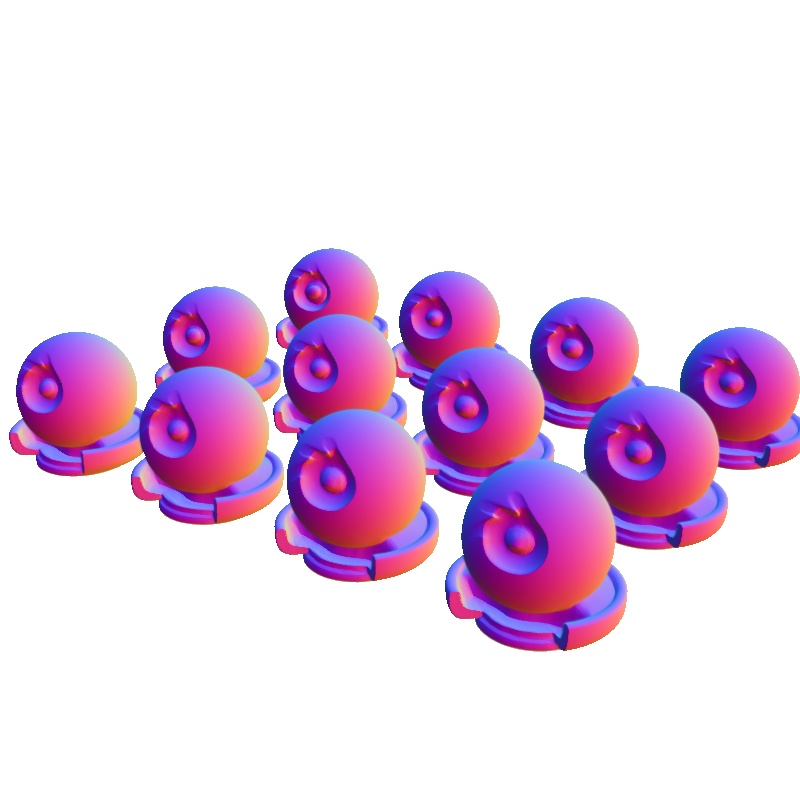} &
\includegraphics[trim={0 50 0 50},clip,width=0.21\linewidth]{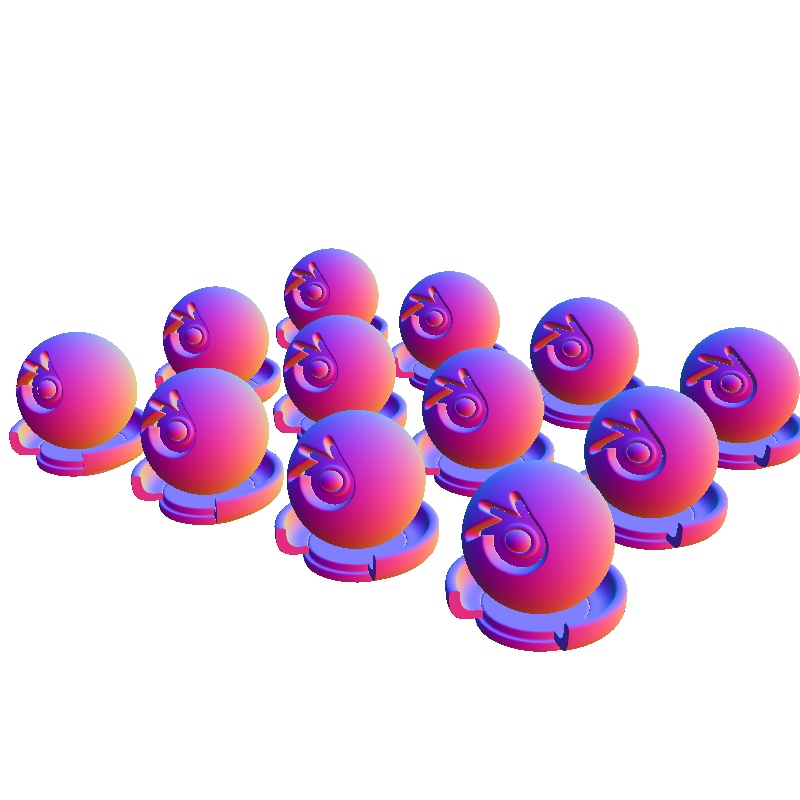} \\

\includegraphics[trim={0 50 0 50},clip,width=0.21\linewidth]{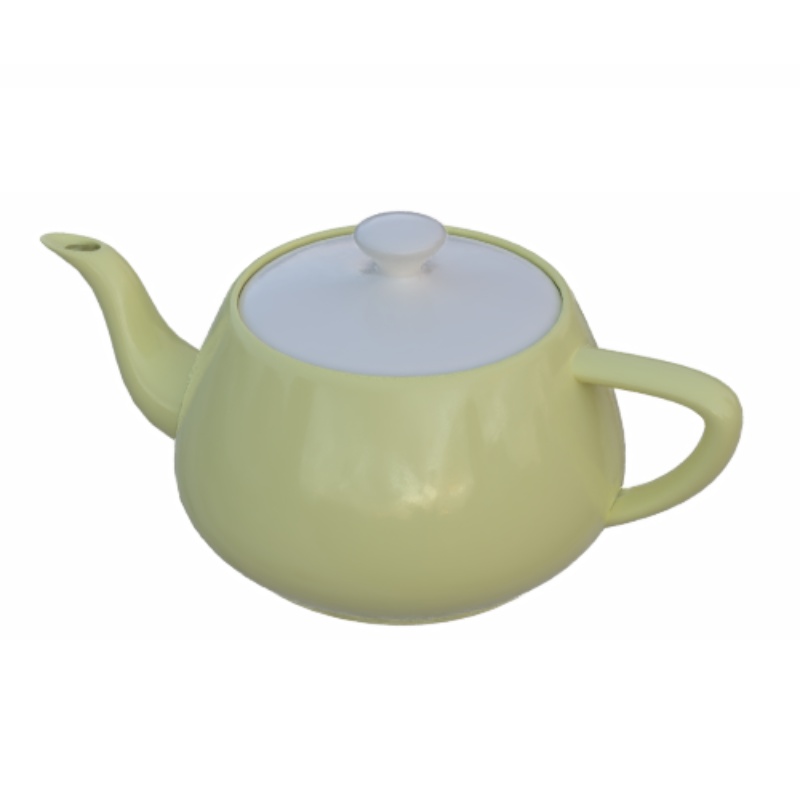} &
\includegraphics[trim={0 50 0 50},clip,width=0.21\linewidth]{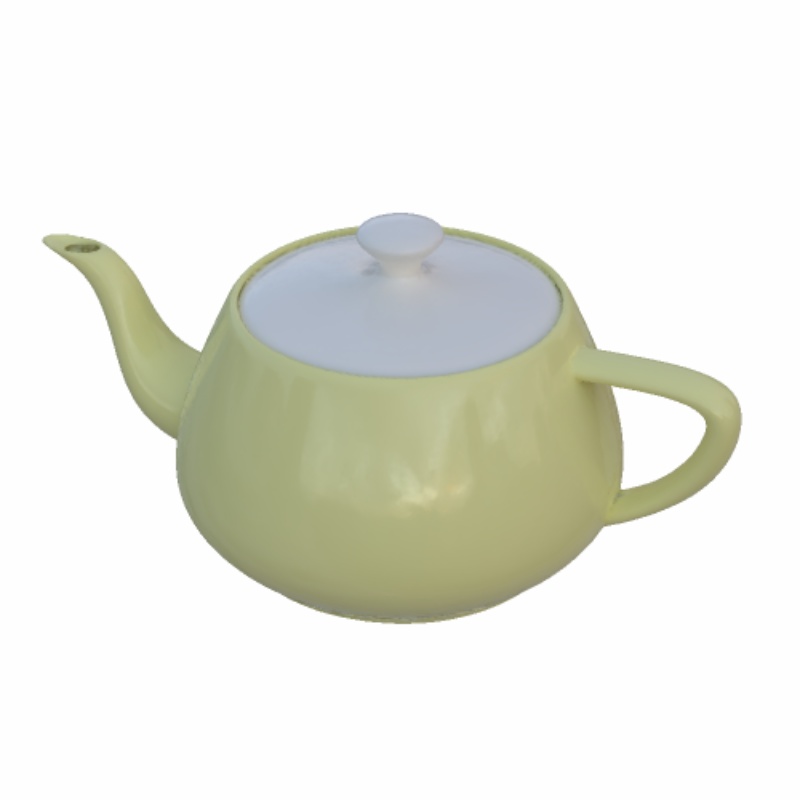} &
\includegraphics[trim={0 50 0 50},clip,width=0.21\linewidth]{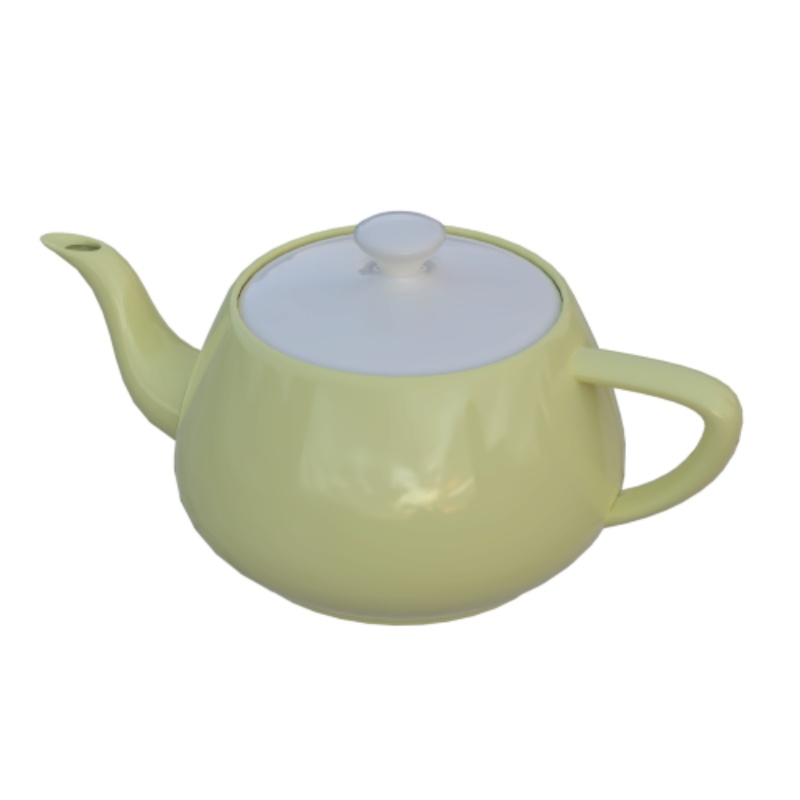} &
\includegraphics[trim={0 50 0 50},clip,width=0.21\linewidth]{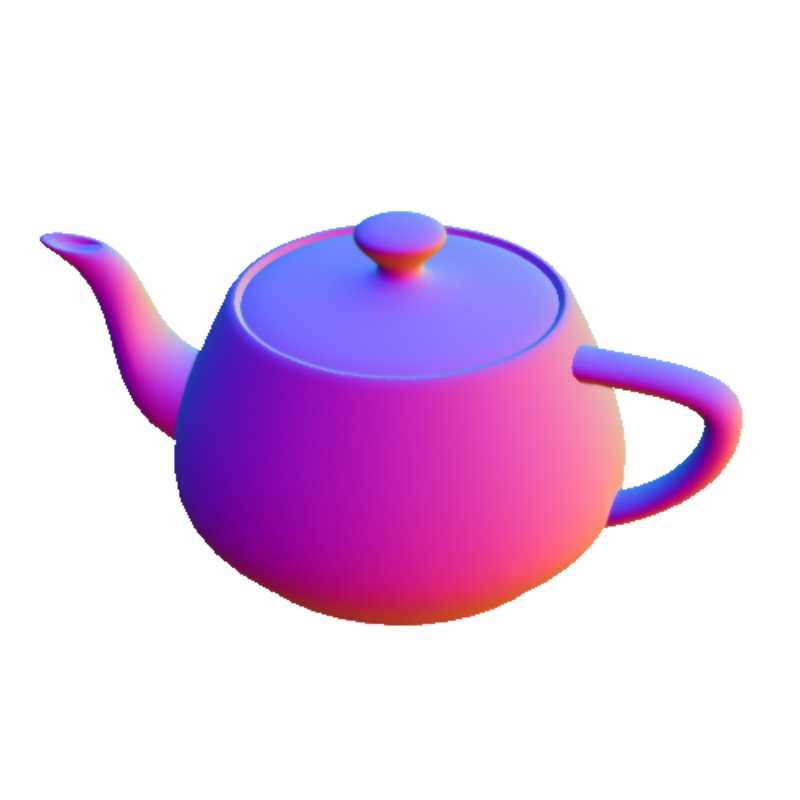} &
\includegraphics[trim={0 50 0 50},clip,width=0.21\linewidth]{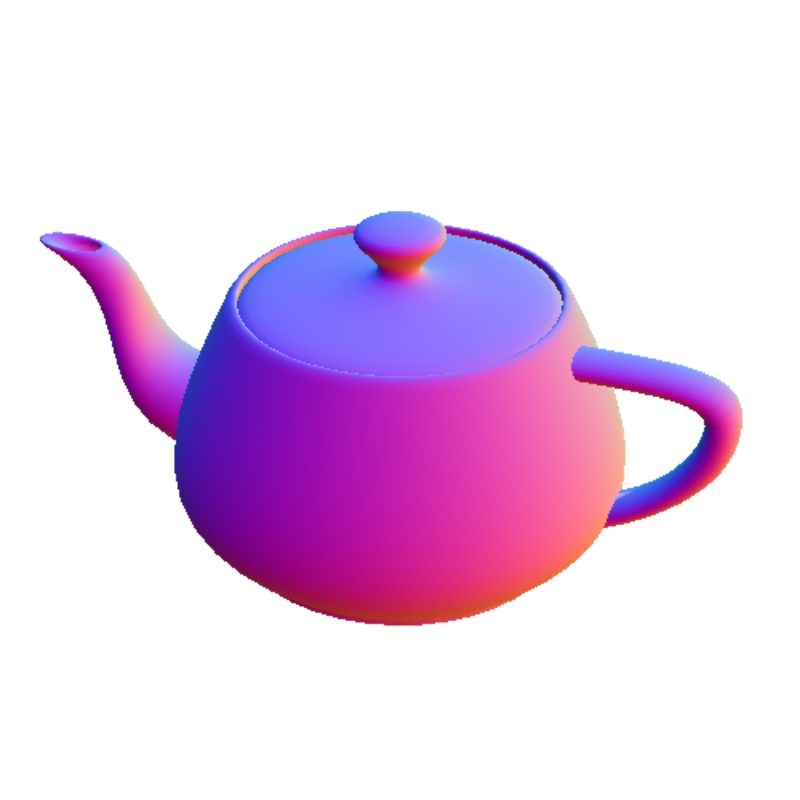} \\

\includegraphics[trim={0 40 0 60},clip,width=0.21\linewidth]{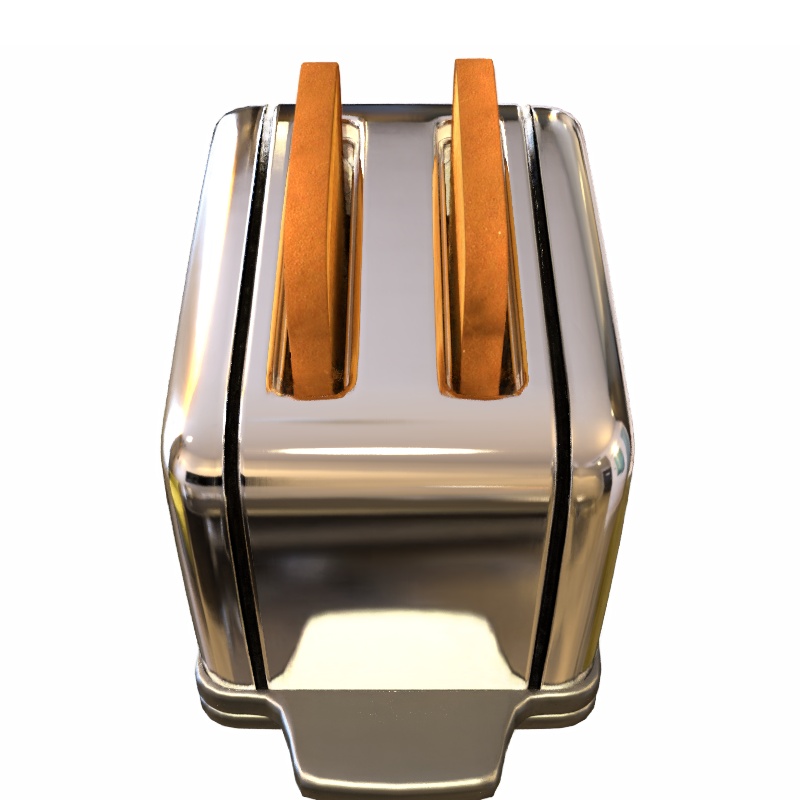} &
\includegraphics[trim={0 40 0 60},clip,width=0.21\linewidth]{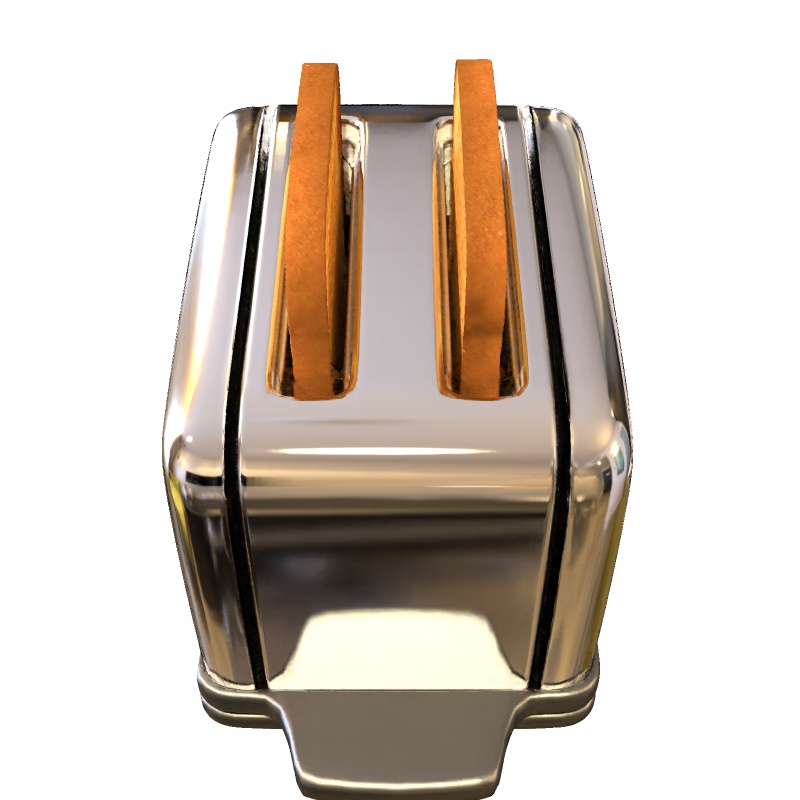} &
\includegraphics[trim={0 40 0 60},clip,width=0.21\linewidth]{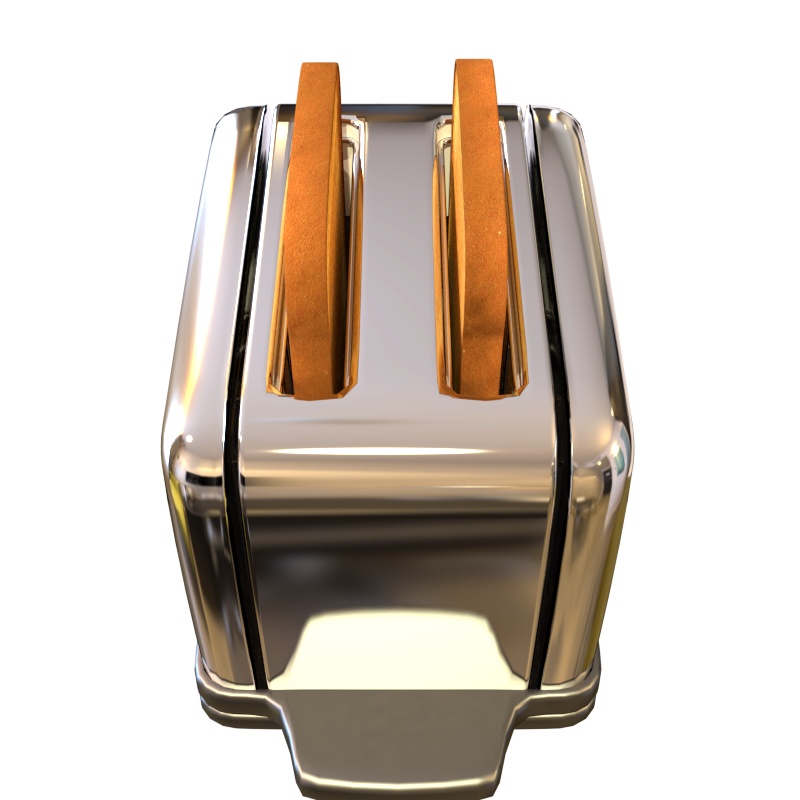} &
\includegraphics[trim={0 40 0 60},clip,width=0.21\linewidth]{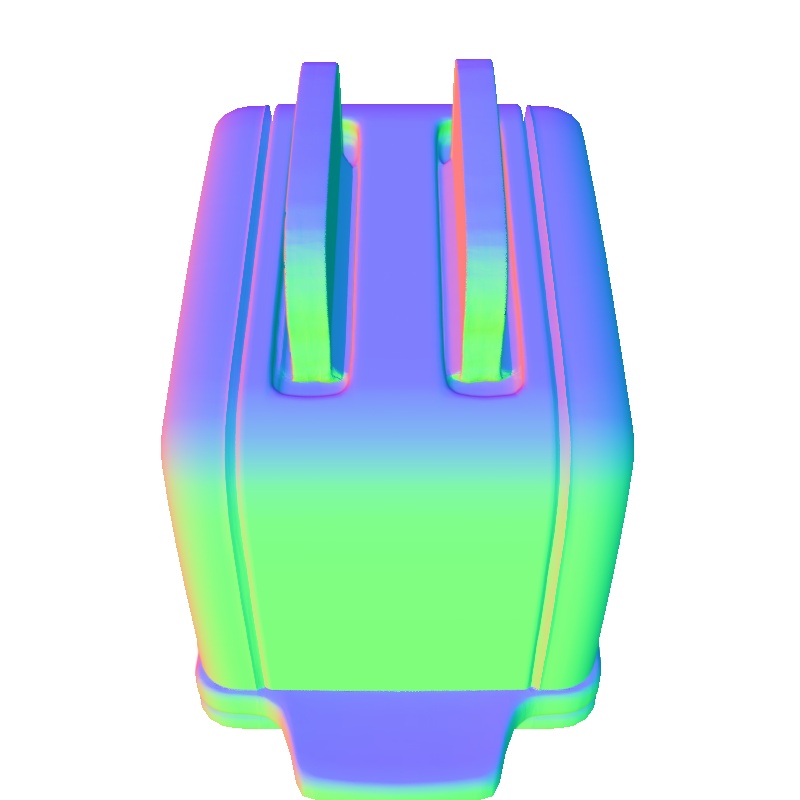} &
\includegraphics[trim={0 40 0 60},clip,width=0.21\linewidth]{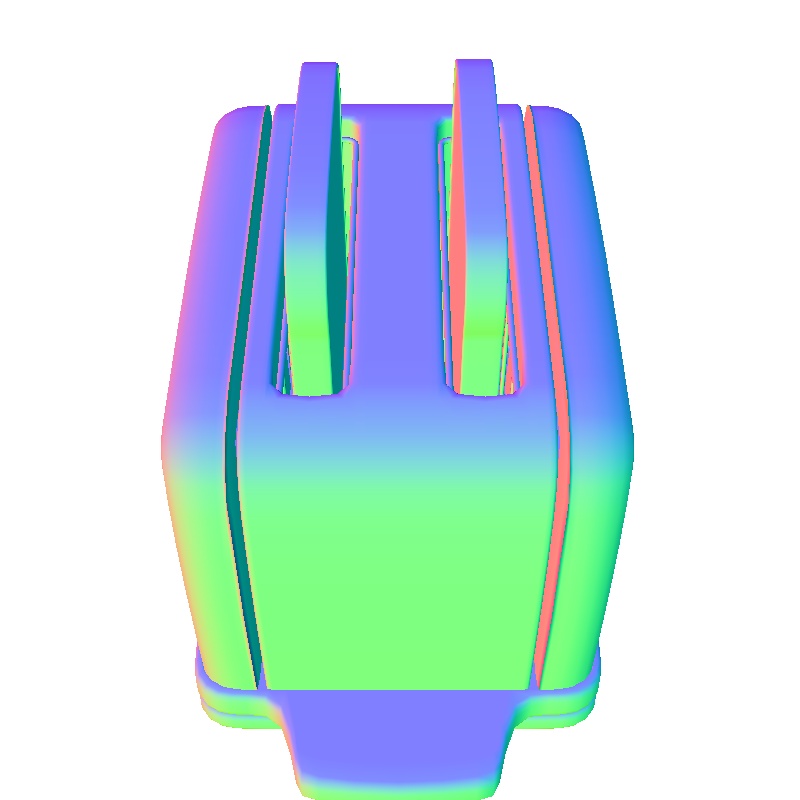} \\

\includegraphics[trim={0 50 0 50},clip,width=0.21\linewidth]{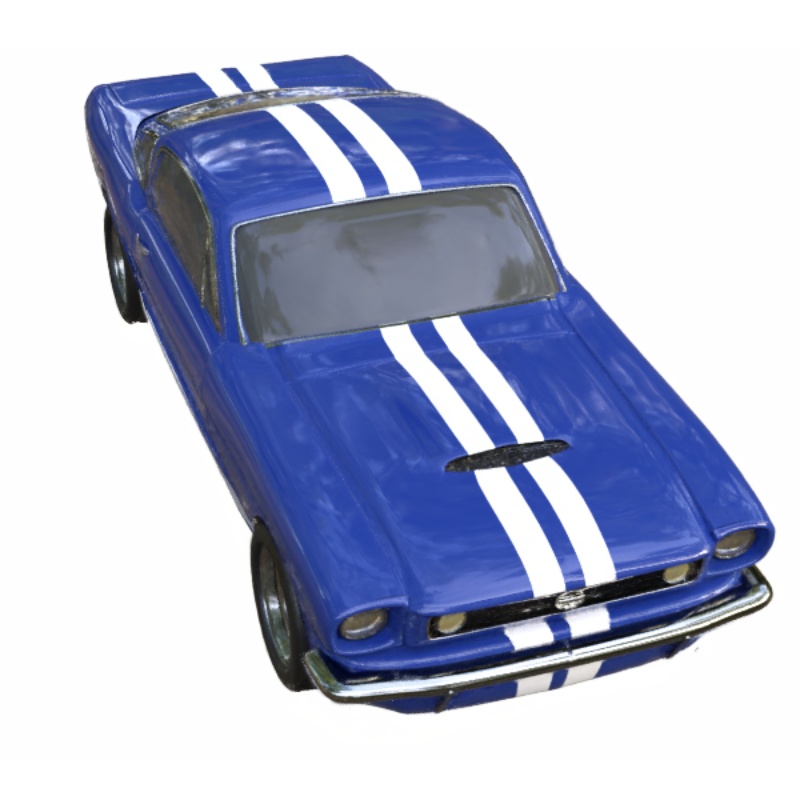} &
\includegraphics[trim={0 50 0 50},clip,width=0.21\linewidth]{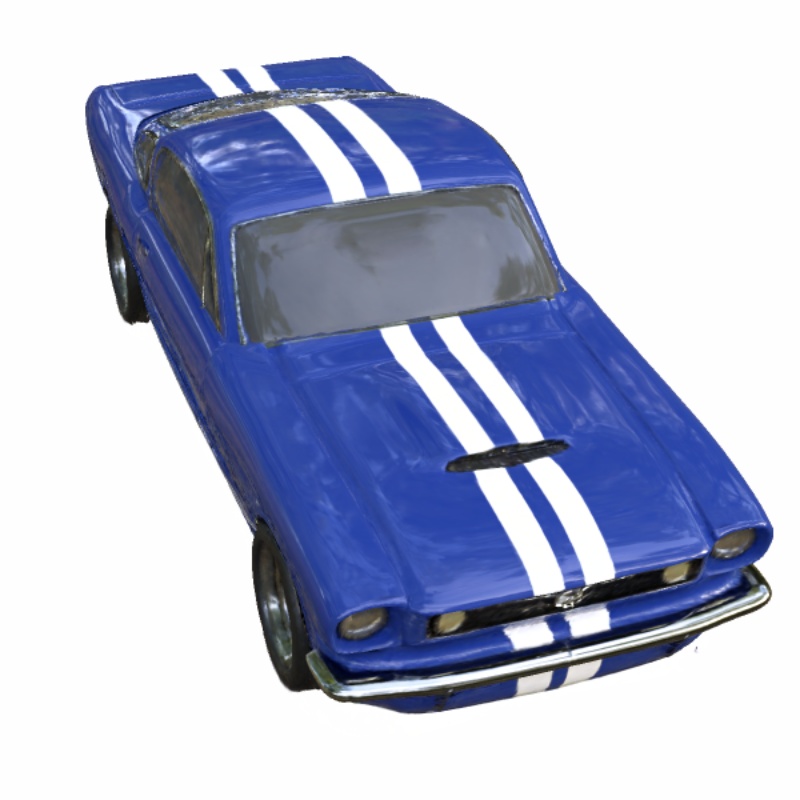} &
\includegraphics[trim={0 50 0 50},clip,width=0.21\linewidth]{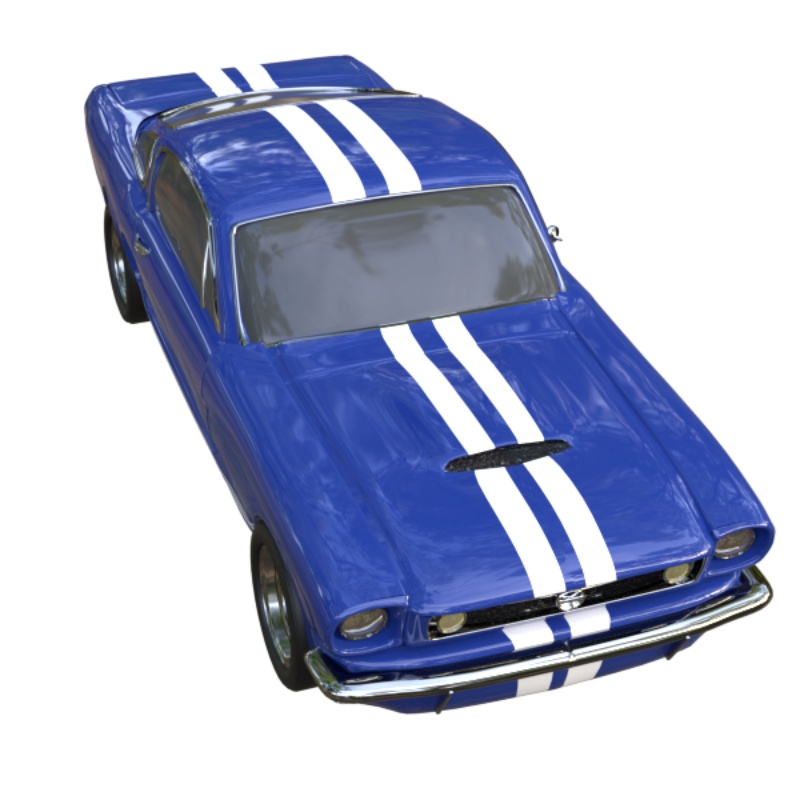} &
\includegraphics[trim={0 50 0 50},clip,width=0.21\linewidth]{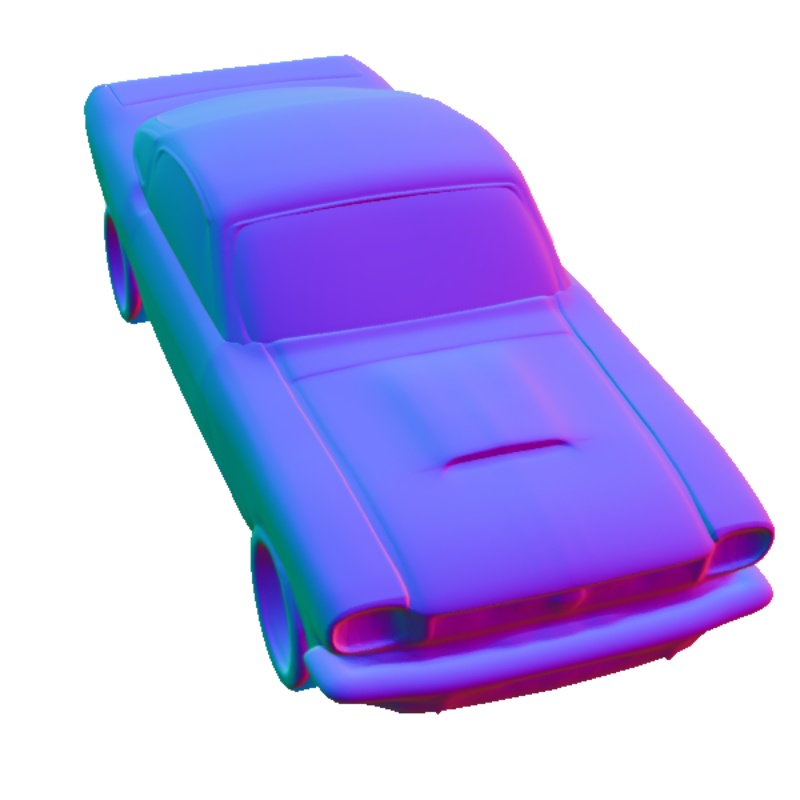} &
\includegraphics[trim={0 50 0 50},clip,width=0.21\linewidth]{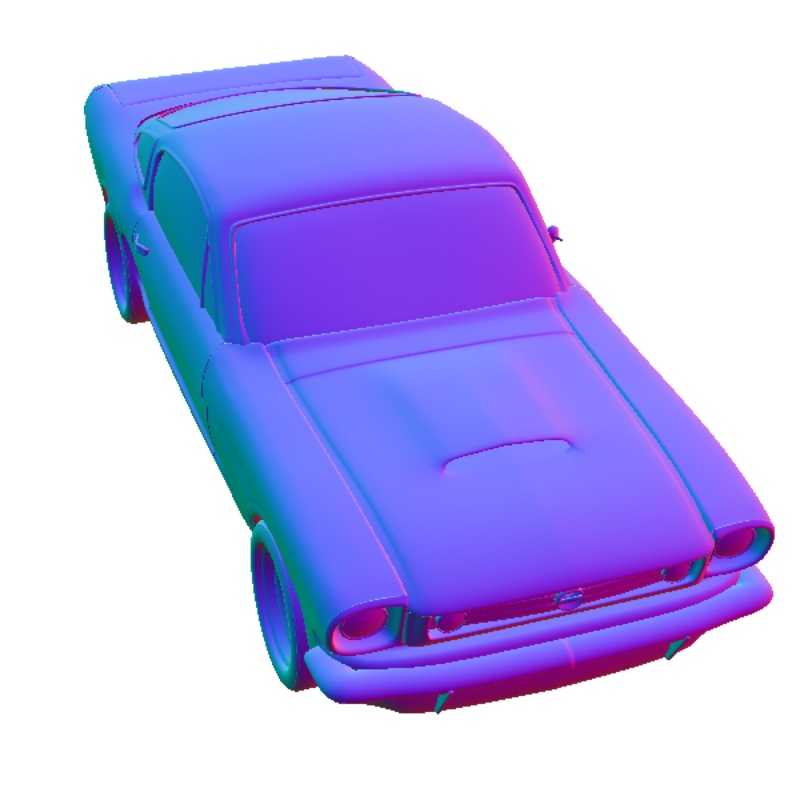} \\

\includegraphics[trim={0 50 0 50},clip,width=0.21\linewidth]{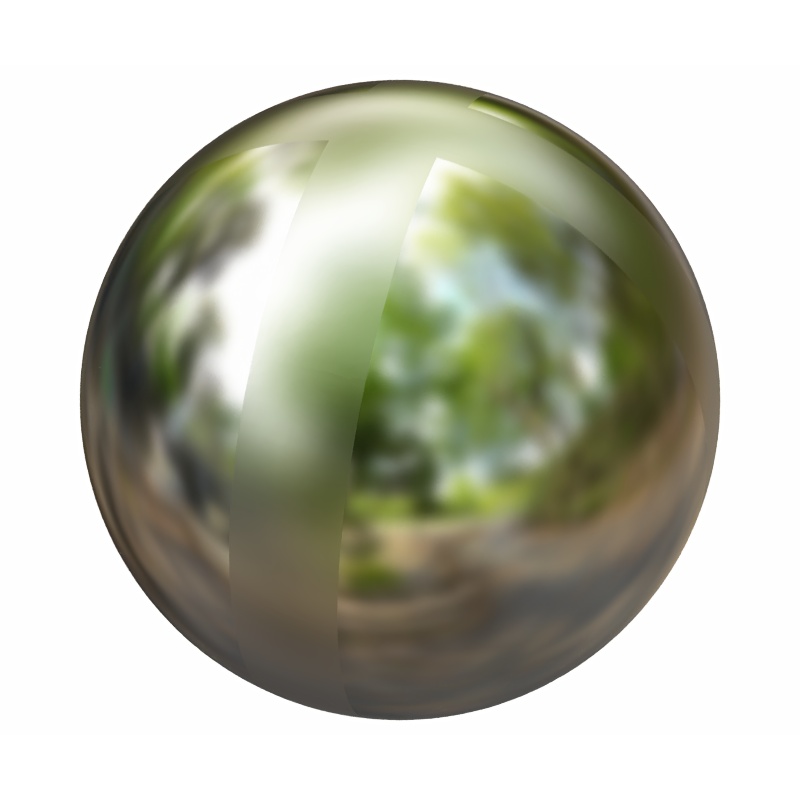} &
\includegraphics[trim={0 50 0 50},clip,width=0.21\linewidth]{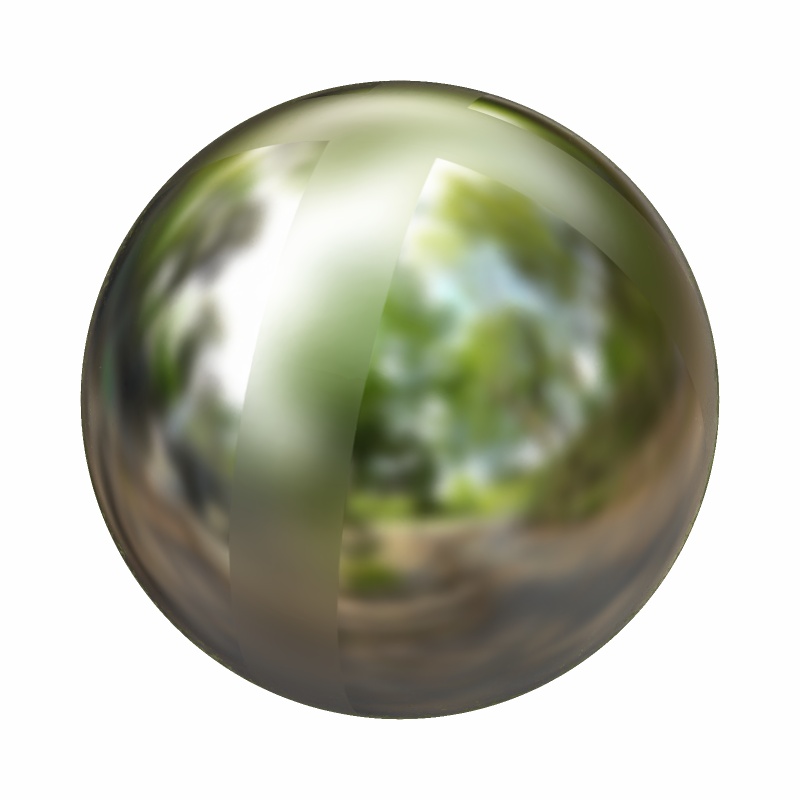} &
\includegraphics[trim={0 50 0 50},clip,width=0.21\linewidth]{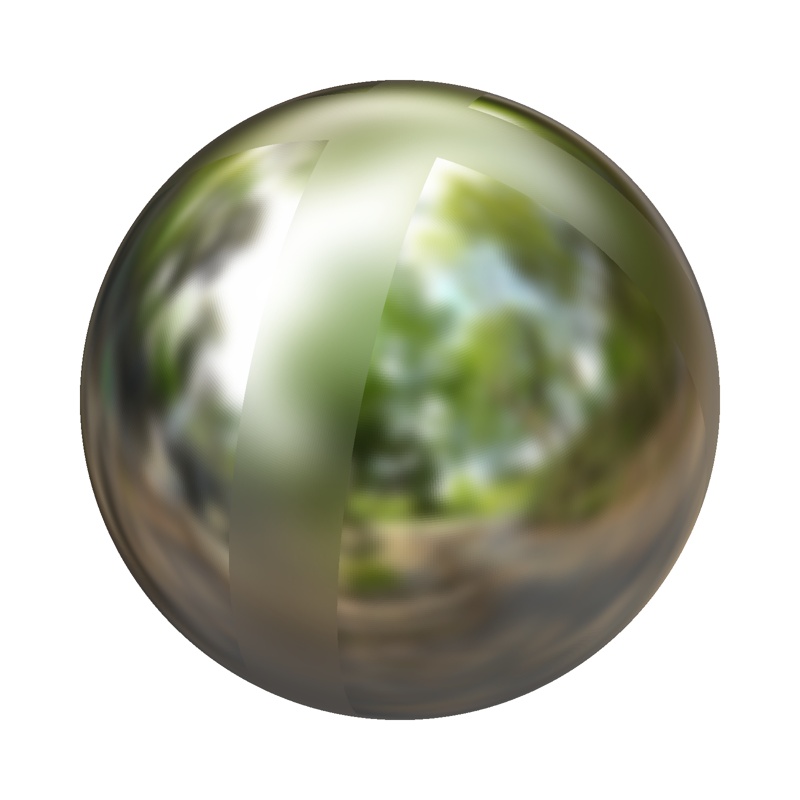} &
\includegraphics[trim={0 50 0 50},clip,width=0.21\linewidth]{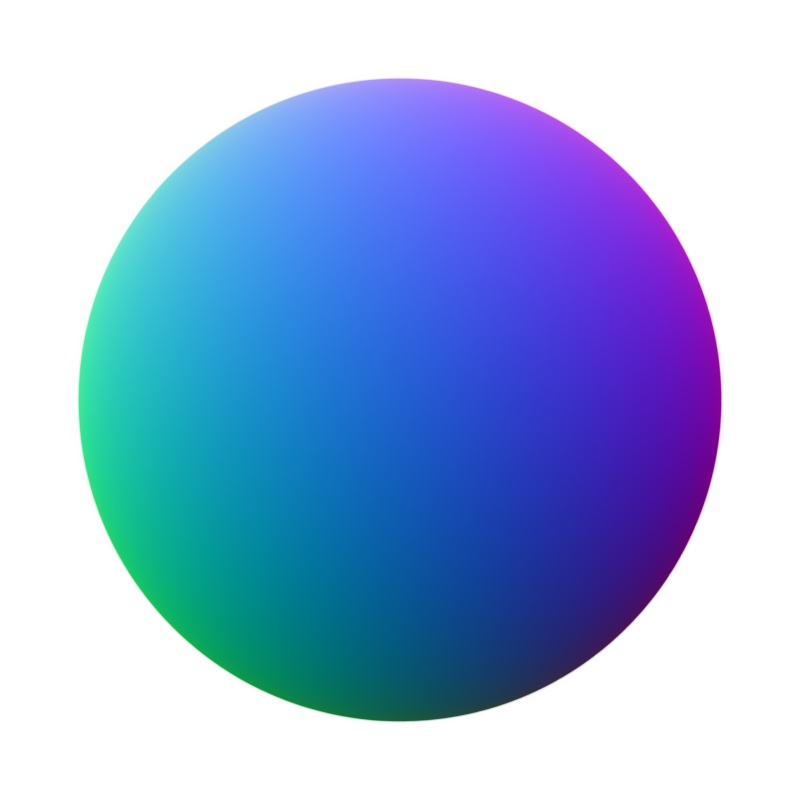} &
\includegraphics[trim={0 50 0 50},clip,width=0.21\linewidth]{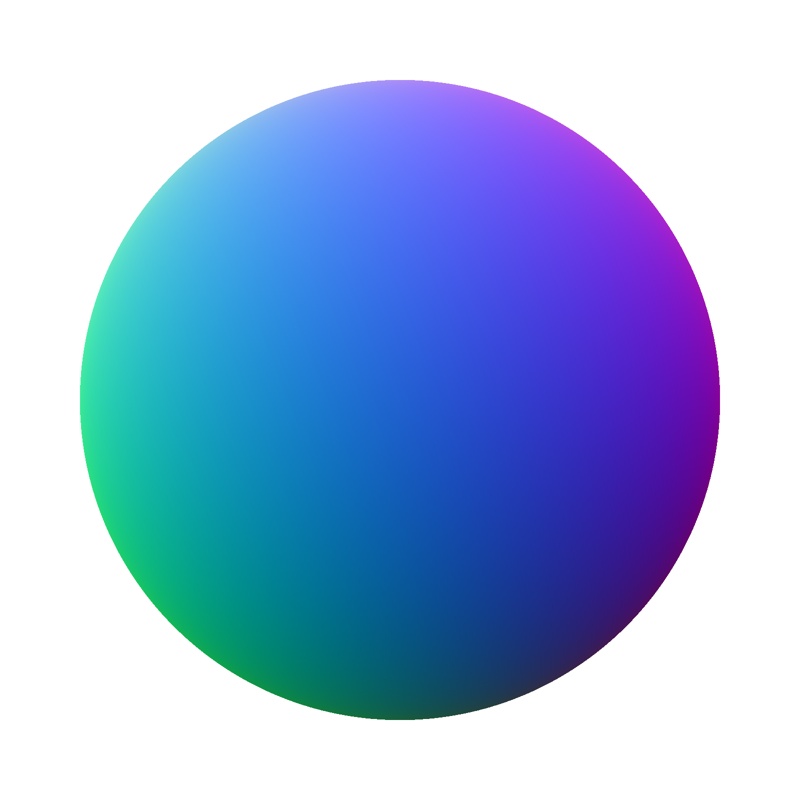} \\

\includegraphics[trim={0 25 0 75},clip,width=0.21\linewidth]{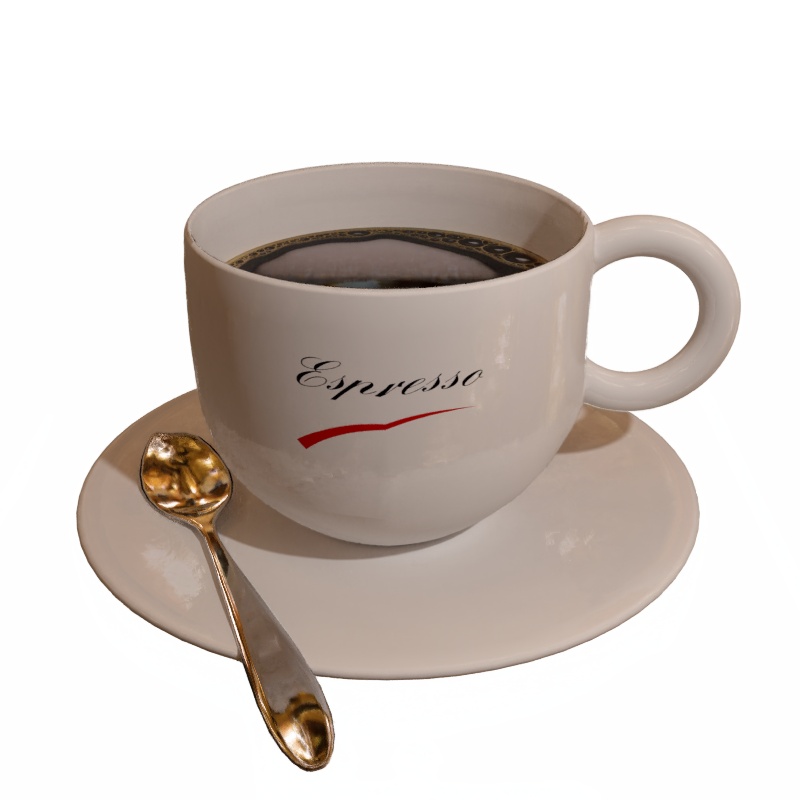} &
\includegraphics[trim={0 25 0 75},clip,width=0.21\linewidth]{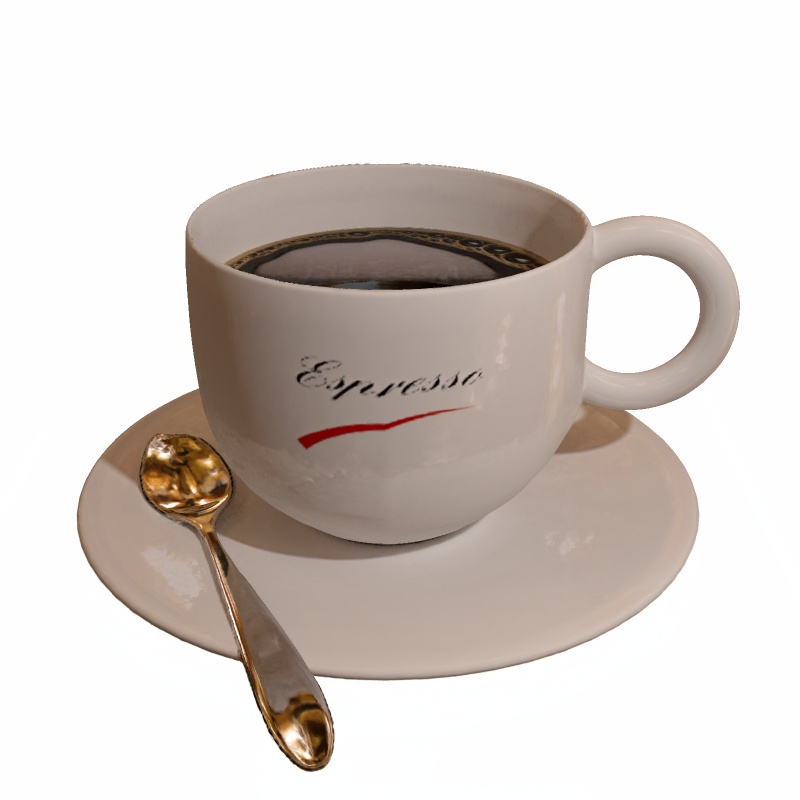} &
\includegraphics[trim={0 25 0 75},clip,width=0.21\linewidth]{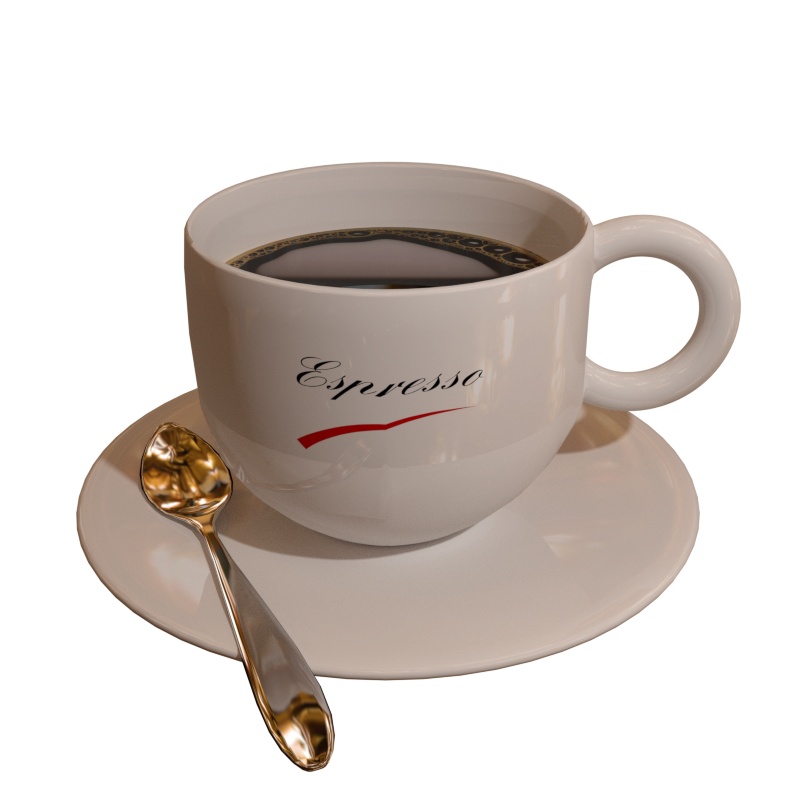} &
\includegraphics[trim={0 25 0 75},clip,width=0.21\linewidth]{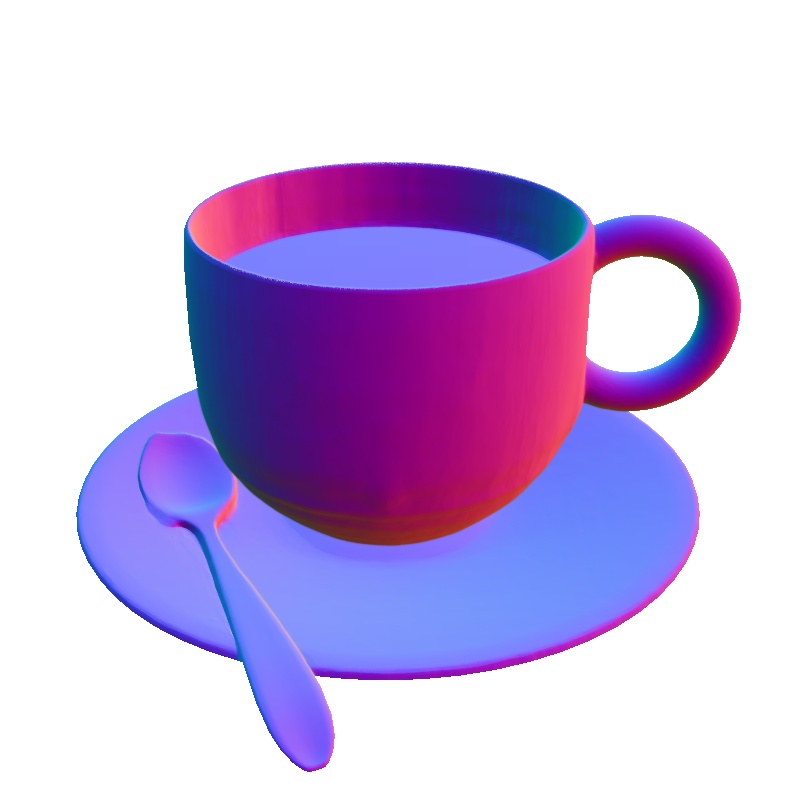} &
\includegraphics[trim={0 25 0 75},clip,width=0.21\linewidth]{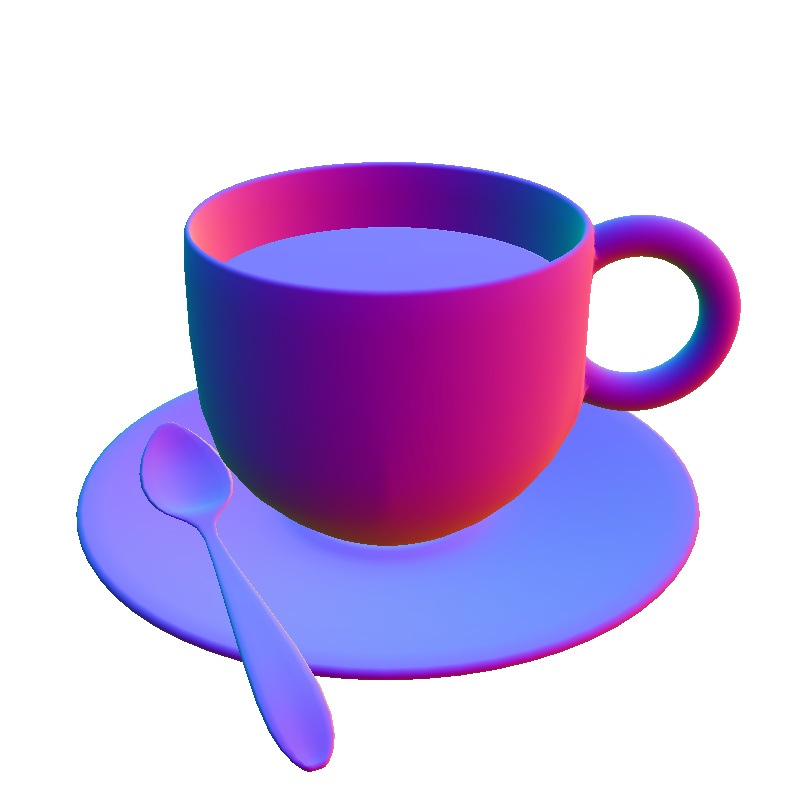} \\

\includegraphics[trim={0 50 0 50},clip,width=0.21\linewidth]{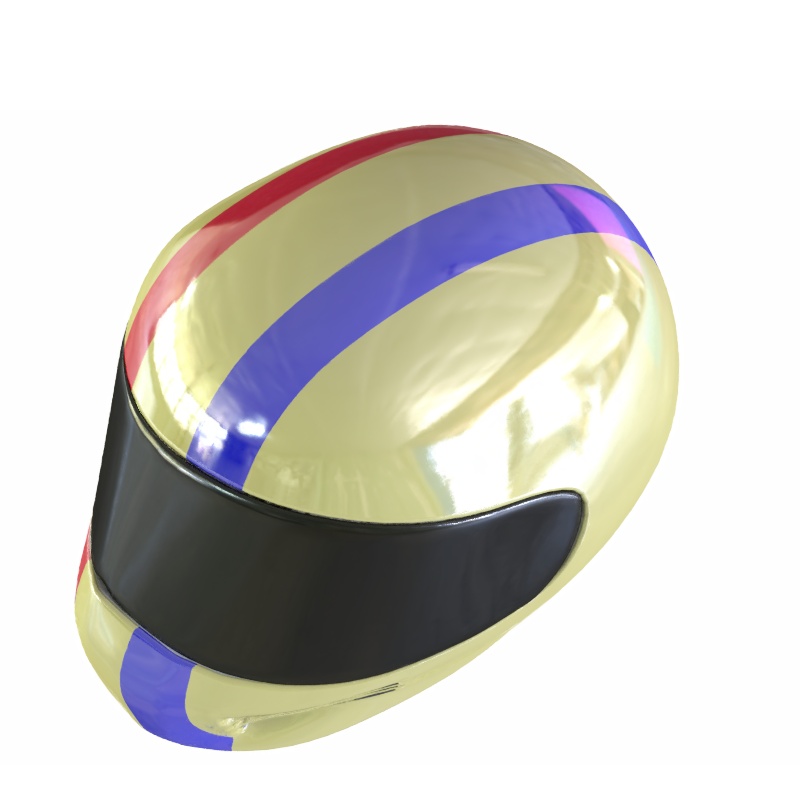} &
\includegraphics[trim={0 50 0 50},clip,width=0.21\linewidth]{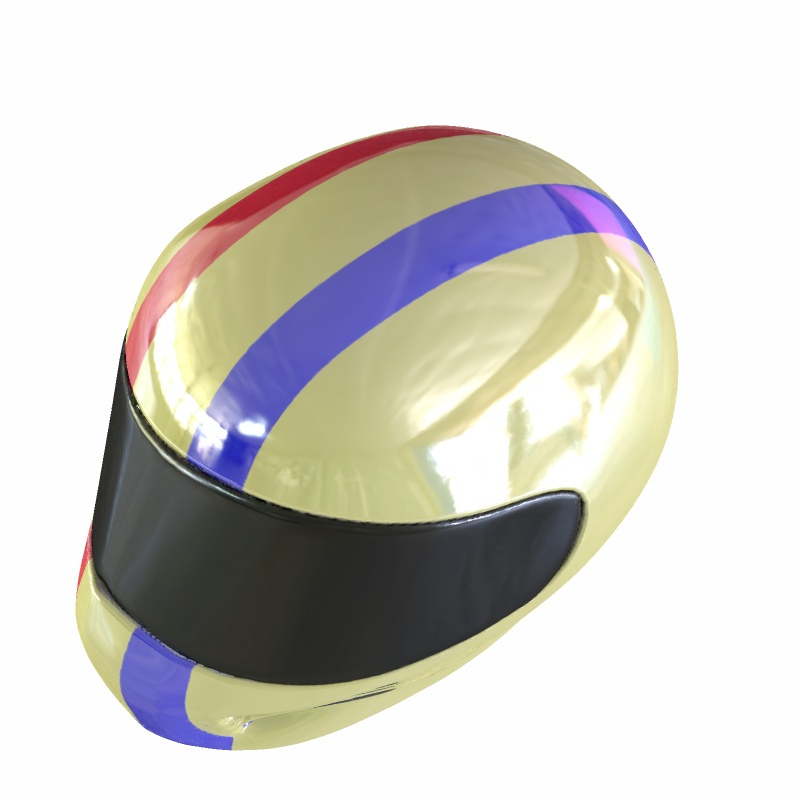} &
\includegraphics[trim={0 50 0 50},clip,width=0.21\linewidth]{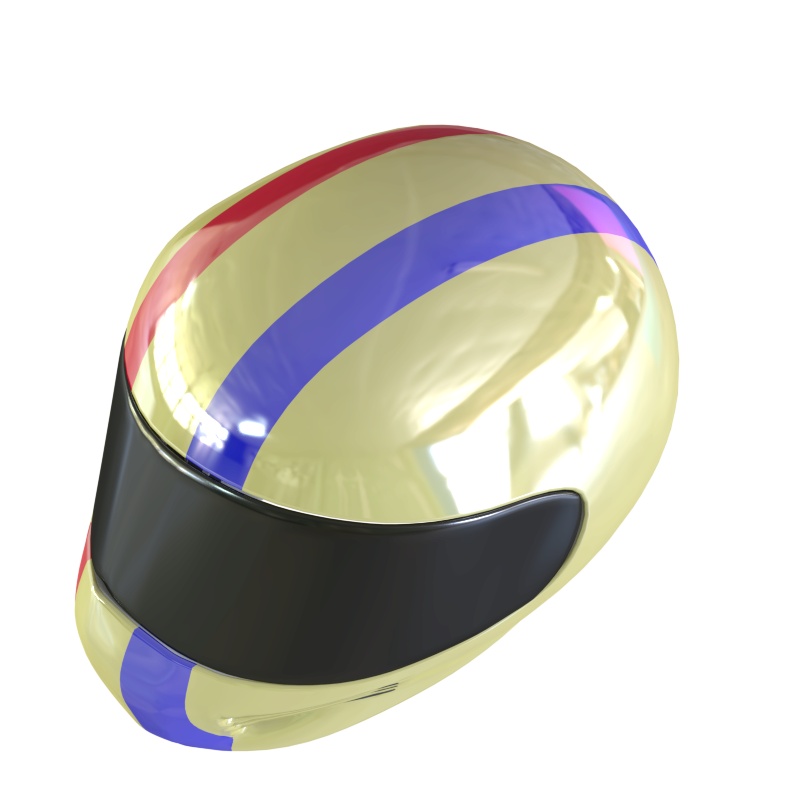} &
\includegraphics[trim={0 50 0 50},clip,width=0.21\linewidth]{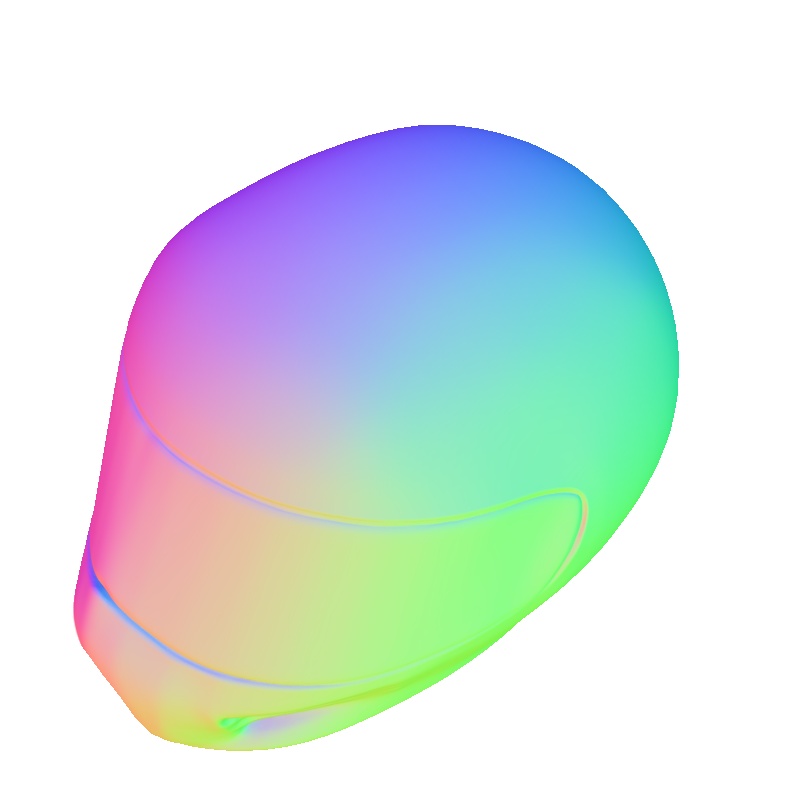} &
\includegraphics[trim={0 50 0 50},clip,width=0.21\linewidth]{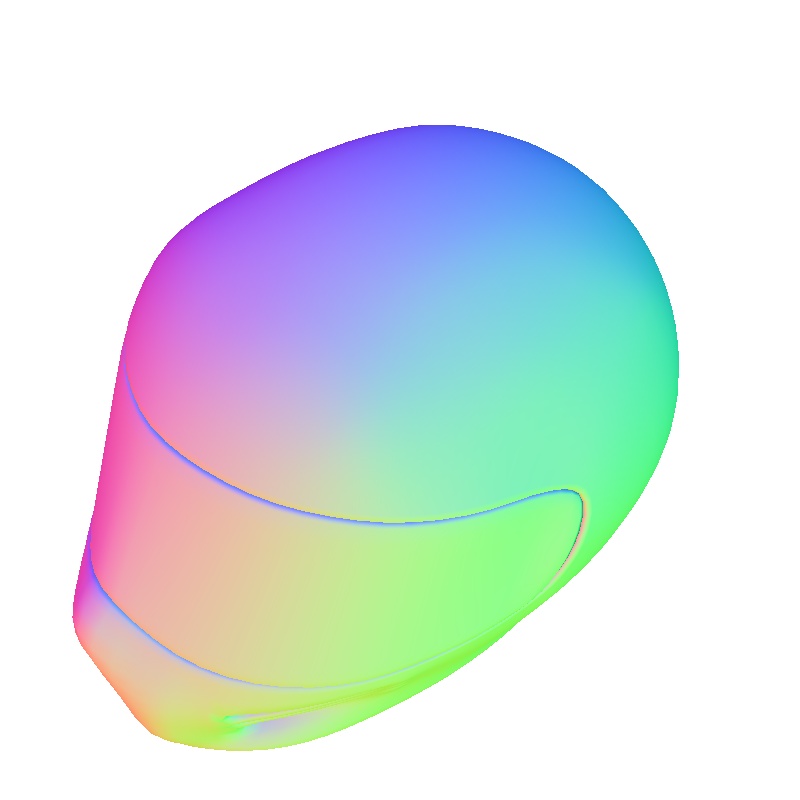} \\
    \end{tabular}
    }
    \caption{\textbf{Qualitative results on each synthetic scene} for our offline (NDE) and real-time (NDE-RT) methods.
    }
     \label{fig:supp}
\end{figure*}
\end{document}